\documentclass[12pt]{article}
\usepackage{amsmath}
\usepackage{graphicx,psfrag,epsf}
\usepackage{enumerate}
\usepackage{natbib}
\usepackage{bbm}
\usepackage{caption}
\usepackage{subcaption}
\usepackage{makecell}
\usepackage{amssymb}
\usepackage{multirow}
\usepackage{xcolor}
\usepackage[scr=boondox]{mathalpha}

\newcommand{\blind}{0}

\date{}

\begin{document}

\def\spacingset#1{\renewcommand{\baselinestretch}%
{#1}\small\normalsize} \spacingset{1}

\newcommand{\revision}{\color{black}}
\newcommand{\revisionii}{\color{black}}

\if0\blind
{
  \title{Mixture of Linear Models Co-supervised by Deep Neural Networks}
  \author{Beomseok Seo\hspace{.2cm}\\
    Department of Statistics, The Pennsylvania State University\\
     \\
    Lin Lin \\
    Department of Biostatistics and Bioinformatics, Duke University\\
     \\
    Jia Li \\
    Department of Statistics, The Pennsylvania State University}
  \maketitle
} \fi

\if1\blind
{
  \bigskip
  \bigskip
  \bigskip
  \begin{center}
    {\LARGE\bf Title}
\end{center}
  \medskip
} \fi

\maketitle

\bigskip
\begin{abstract}

Deep neural networks (DNN) have been demonstrated to achieve unparalleled prediction accuracy in a wide range of applications. Despite its strong performance, in certain areas, the usage of DNN has met resistance because of its black-box nature. In this paper, we propose a new method to estimate a mixture of linear models (MLM) for regression or classification that is relatively easy to interpret. We use DNN as a proxy of the optimal prediction function such that MLM can be effectively estimated. We propose visualization methods and quantitative approaches to interpret the predictor by MLM. Experiments show that the new method allows us to trade-off interpretability and accuracy. The MLM estimated under the guidance of a trained DNN fills the gap between a highly explainable linear statistical model and a highly accurate but difficult to interpret predictor.

\end{abstract}

\noindent%

\spacingset{1.5} 

\section{Introduction}
\label{sec:intro}

Deep neural network (DNN) models have achieved phenomenal success for applications in many domains, ranging from academic research in science and engineering to industry and business. The modeling power of DNN is believed to have come from the complexity and over-parameterization of the model, which on the other hand has been criticized for the lack of interpretation. Although certainly not true for every application, in some applications, especially in economics, social science, healthcare industry, and administrative decision making, scientists or practitioners are resistant to use predictions made by a black-box system for multiple reasons. One reason is that a major purpose of a study can be to make discoveries based upon the prediction function, {\it e.g.}, to reveal the relationships between measurements. Another reason can be that the training dataset is not large enough to make researchers feel completely sure about a purely data-driven result. Being able to examine and interpret the prediction function will enable researchers to connect the result with existing knowledge or  gain insights about new directions to explore. Although classic statistical models are much more explainable, their accuracy often falls considerably below DNN. In this paper, we propose an approach to fill the gap between relatively simple explainable models and DNN such that we can more flexibly tune the trade-off between interpretability and accuracy. Our main idea is a mixture of discriminative models that is trained with the guidance from a DNN. Although mixtures of discriminative models have been studied before, our way of generating the mixture is quite different.

\subsection{Related Work}
\label{sec:related}

Despite the fact that many attempts have been made to make DNN models more interpretable, a formal definition of ``human interpretability'' has remained elusive. The concept of interpretability is multifaceted and inevitably subjective. An in-depth discussion about the meaning of being interpretable is given by the review article of~\cite{doshi2017towards}, which confirms the richness of this concept and provides philosophical viewpoints. In the literature, different approaches have been proposed to  define ``interpretation''. A complete unified  taxonomy for all the existing approaches does not exist. Nevertheless, one way is to categorize recent works into two types: one that builds more interpretable models by reducing model complexity, and the other that attempts to interpret a complex model by examining certain aspects of it, {\it e.g.}, how decisions are made locally. For the first type of approaches, interpretability is aimed at during the model's training phase, and what constitutes good interpretability has been addressed in diverse ways. For example, attention-based methods \citep{bahdanau2014neural, vaswani2017attention} propose an attention score for neural machine translation; generalized additive models (GAM) \citep{lou2013accurate, agarwal2020neural, guo2020interpretable} interpret pair-wise interaction effects by imposing an additive structure; and feature selection methods impose cross-entropy error function~\citep{verikas2002feature} or $L_0$ penalization~\citep{tsang2018neural}. In contrast, the second type of approaches, the so-called model-agnostic methods \citep{ribeiro2016model}, are post-hoc in the sense that they analyze an already trained model using interpretable measurements. For example, sensitivity analysis \citep{lundberg2017unified}, local linear surrogate models \citep{ribeiro2016should} and Bayesian non-parametric mixture model \citep{NEURIPS2018_4b4edc26} search for important features for specific samples according to a given prediction model. The same aspect of being interpretable can be tackled by either type of the approaches.
For example, \cite{tsang2017detecting} searches for interaction effects of original features as captured in a trained neural network by computing a strength score based on the learned weights. In~\cite{tsang2018neural}, similar results can be obtained by 
training the neural network model with a penalty to learn the interaction effects.

Various kinds of interpretation have been suggested, including but not limited to visualizing the result, generating human explainable rules, selecting features, or constructing prototype cases. The pros and cons of existing methods are discussed in~\cite{molnar2020interpretable}. What kind of interpretation is useful also depends on the application field. For instance, feature selection for image analysis is usually not as meaningful as that for tabular data. Many proposed methods target particular applications, {\it e.g.}, image analysis \citep{zhang2018interpretable, chen2019looks}, text mining \citep{vaswani2017attention}, time series \citep{guo2019exploring}. 

\cite{ribeiro2016should} interprets neural networks by explaining how the decision is made in the vicinity of every instance. In particular, the trained neural network is used to generate pairs of input and output quantities, based on which a linear regression model is estimated. This linear model is used to explain the decision in the neighborhood of that instance. Although the local linear models help understand the prediction around every single point, they are far from providing global perspectives.
Our new method is inspired by~\cite{ribeiro2016should}, but we tackle two additional problems. First, our method will produce a stand-alone prediction model that is relatively easy to interpret. In another word, our method is not the model-agnostic type to explain an actual operating DNN. Second, our method aims at globally interpreting the decision.

Although motivated from rather different aspects, in terms of the formulation of the model, our method is related to locally weighted regression \citep{cleveland1988locally}. {\revision In other words, our method} splits the space of the original independent variables by exploiting a trained DNN and conducts linear regression inside each local region. In non-parametric regression, many other flexible regression models have been studied, for instance, kernel regression \citep{nadaraya1964estimating, watson1964smooth}, generalized additive models \citep{hastie1990generalized}, and classification and regression trees (CART) \citep{breiman1984classification}. Some of the methods such as CART are by construction highly interpretable, but some are not. More recently, \cite{guo2020interpretable} have used neural networks to efficiently learn the classic GAM \citep{hastie1990generalized} models. 

{\revision Our method is in spirit similar to the Mixture of Experts (MOE) model or a fuzzy system \citep{jacobs1991adaptive,takagi1985fuzzy} although profound differences exist in practically important aspects. MOE divides the feature space into regions and applies localized experts in each region. The study on MOE has focused on finding better ways to create the individual experts and to combine them into a single predictor. The approaches for creating the individual experts include stochastic partitioning \citep{ebrahimpour2008mixture}, negatively correlated experts \citep{masoudnia2012combining}, and clustering-based experts ({\it e.g.}, self-organizing maps (SOM))~\citep{tang2002input}. Weights to combine the experts have been computed by methods such as gating network \citep{kuncheva2014combining} and boosting \citep{kegl2013return}. However, the MOE method partitions the feature space using only the features. Our method is different because it uses a pre-trained DNN to partition the samples, in a supervised manner based on the prediction of the DNN. Our method is motivated by finding an easier interpretation while preserving the accuracy of a DNN. Experimental results are provided to compare our method with the mixture of experts approach.}

\subsection{Overview of Our Approach}
\label{sec:summary}

One natural idea to explain a complex model is to view the decision as a composite of decisions made by different functions in different regions of the input space. Explanation of the overall model consists of two parts: to explain how the partition of the input space is formed, and to explain the prediction in each region. This idea points to the construction of a mixture model in which every component is a linear model, which is called the {\em mixture of linear models} (MLM). Although a linear model is not necessarily always easy to explain, relatively speaking, it is explainable, and via LASSO-type sparsity penalty \citep{tibshirani1996regression} it can be made increasingly more explainable. 

The main technical hurdle is to find a suitable partition of the instances such that component-wise linear models can be estimated. The challenge faced here is in stark contrast to that for building generative mixture models, {\it e.g.}, Gaussian mixture models (GMM), where the proximity of the independent variables themselves roughly determines the grouping of instances into components. The same approach of forming components is a poor choice for building a mixture of discriminative models. The similarity between the instances is no longer measured by the proximity of the independent variables but by the proximity of the relationships between the dependent variable ($Y$) and the independent variables ($X$). {\revision As a related approach,
cluster-weighted modeling (CWM) \citep{gershenfeld1997nonlinear, ingrassia2012local} takes into account the joint distribution of $X$ and $Y$ when creating a generative mixture model.
However, in CWM, the grouping of $X$ and the estimation of local models are treated separately. We will compare the performance of CWM and MLM in the experiments.}

The dependence between $Y$ and $X$ cannot be adequately captured by the observation of one instance. In the case of moderate to high dimensions, it is also impractical to estimate the discriminative function, $\hat{Y}=f(X)$, based on nearby points. Our key idea is to use the DNN model as an approximation of the theoretically optimal prediction function, which is employed to guide the partition of the instances. We also propose to use visualization based on GMM and decision tree to interpret the partition of the instances, which is crucial for obtaining interpretation in a global sense. 

It is debatable whether DNN serves well as an approximation of the optimal prediction function. In many applications, we observe that DNN achieves the best accuracy among the state-of-the-art methods, {\it e.g.}, support vector machine (SVM) \citep{cortes1995support} and random forest (RF) \citep{breiman2001random}. It is thus reasonable to exploit DNN in this manner. However, we acknowledge that a weakness of our approach is that the accuracy is capped by DNN, and our approach is likely to suffer when DNN itself performs poorly.

The rest of the paper is organized as follows. In Section \ref{sec:prelim}, we introduce notations and summarize existing methods most relevant to our proposed work. 
We present MLM in Section \ref{sec:MLM} and describe the tools developed for interpretation based upon MLM in Section \ref{sec:interpret}. In Section \ref{sec:exp}, experimental results are reported for four real-world datasets including two clinical datasets for classification and two for regression. For prediction accuracy, comparisons have been made with multiple approaches including DNN. We also illustrate how the interpretation tools are used. Finally, conclusions are drawn in Section \ref{sec:discuss}.

\section{Preliminaries}
\label{sec:prelim}
Let $X\in\mathbb{R}^{p}$ be the independent variables (or covariates) and $Y\in\mathbb{R}$ be the dependent variable. Denote the sample space of $X$ by $\mathcal{X}$.
Let $\{y_i\}_{i=1}^n$ and $\{\mathbf{x}_i\}_{i=1}^n = \{(x_{i,1},\cdots,x_{i,p})^T\}_{i=1}^{n}$ be the $n$ observations of $Y$ and $X$. Denote the input data matrix by $\mathbf{X}\in\mathbb{R}^{n\times p}$. In regression analysis, we are interested in estimating the following regression function for any $\mathbf{x}\in \mathcal{X}$ (for classification, we simply substitute $Y$ by $g(Y)$ via a link function $g(\cdot)$).
\begin{eqnarray}
m(\mathbf{x}) = \mathbb{E}(Y|X=\mathbf{x}) \; .
\label{eq:regress}
\end{eqnarray}

In linear regression, it is assumed that $m(\mathbf{x})=\alpha+\mathbf{x}^T\mathbf{\beta}$. In non-linear regression, a linear expansion of basis functions is often used to estimate $m(\mathbf{x})$. For example, Nadaraya-Watson kernel regression \citep{nadaraya1964estimating, watson1964smooth} assumes $m(\mathbf{x})$ to be an additive form of kernel functions with a given bandwidth $h$:
\begin{eqnarray}
m(\mathbf{x})= \sum\limits_{i=1}^n\frac{G(\frac{\mathbf{x}-\mathbf{x}_i}{h})y_i}{\sum\limits_{i'=1}^n G(\frac{\mathbf{x}-\mathbf{x}_{i'}}{h})} \, ,
\label{eq:kernel}
\end{eqnarray}
where $G(\mathbf{x})=(2\pi)^{-k/2}e^{-\mathbf{x}^T\mathbf{x}/2}$. The locally weighted regression \citep{cleveland1988locally} extends Nadaraya-Watson's $m(\mathbf{x})$ by changing the constant prediction in each band to a linear function. Regression splines use piecewise polynomial basis functions between fixed points, known as knots, and ensure smoothness at the knots \citep{schoenberg1973cardinal}.
\cite{friedman1991multivariate} extended the spline method to handle higher dimensional data.
These kernel and spline-based approaches use the basic binning strategy to separate the covariate space into regions, suffering from curse of dimensionality even at moderate dimensions.

In recent years, DNN has become increasingly popular in high-dimensional applications because of its remarkable prediction accuracy. Consider a feed-forward neural network with $L$ hidden layers and one output layer. Its regression function, denoted by $\widetilde{m}(\mathbf{x})$, is defined by the composition of the affine transform and a non-linear activation function at each layer. Let $p_l$ be the number of hidden units at the $l$-th hidden layer, $l=0, ..., L$, and $p_0 = p$. Let $\mathbf{z}^{(l)} \in \mathbb{R}^{p_l}$ be the outputs of the $l$-th hidden layer. Set $\mathbf{z}^{(0)} = \mathbf{x}$. The mapping at the $l$-th hidden layer, $\mathbf{z}^{(l)}=h_l(\mathbf{z}^{(l-1)})$, is defined by
\begin{eqnarray}
	\mathbf{z}^{(l)} = h_l(\mathbf{z}^{(l-1)})
	 =  \sigma_l(\mathbf{W}^{(l)}\mathbf{z}^{(l-1)}+\mathbf{b}^{(l)}), \;\; l=1,\cdots,L,
\label{eqn:hiddenlayer}
\end{eqnarray}
where $\sigma_l(\cdot)$ is a non-linear element-wise activation function such as ReLU~\citep{nair2010rectified}, sigmoid or hyperbolic tangent, and $\mathbf{W}^{(l)}\in \mathbb{R}^{p_{l}\times p_{l-1}}$ and $\mathbf{b}^{(l)}\in\mathbb{R}^{p_{l}}$ are the model parameters. At the output layer, $g(\mathbf{z}^{(L)})$, is defined as either a linear or softmax function depending on whether the purpose is regression or classification. In summary, the neural network model, $\widetilde{m}(\mathbf{x})$, is given by
\begin{eqnarray}
\widetilde{m}(\mathbf{x}) = (g \circ h_L\circ h_{L-1} \circ \cdots \circ h_1)(\mathbf{x}),
\label{eqn:neuralnet}
\end{eqnarray}
and its parameters are trained by gradient descent algorithm so that the mean squared error loss for regression or the cross-entropy loss for classification is minimized. Specifically, the mean squared error loss is defined by
$\displaystyle \sum\limits_{i=1}^n \left[y_i-\widetilde{m}(\mathbf{x}_i)\right]^2/n$,
and the cross-entropy loss is
$\displaystyle -\sum\limits_{i=1}^n \left[y_i\log(\widetilde{m}(\mathbf{x}_i))+(1-y_i)\log(1-\widetilde{m}(\mathbf{x}_i))\right]/n$.
Because of the multiple layers and the large number of hidden units at every layer, it is difficult to interpret a DNN model, for example, to explain the effects of the input variables on the predicted variable.

\section{Methods}
\label{sec:methods}

\begin{figure}[htp]
\begin{center}
	\includegraphics[width=0.8\textwidth]{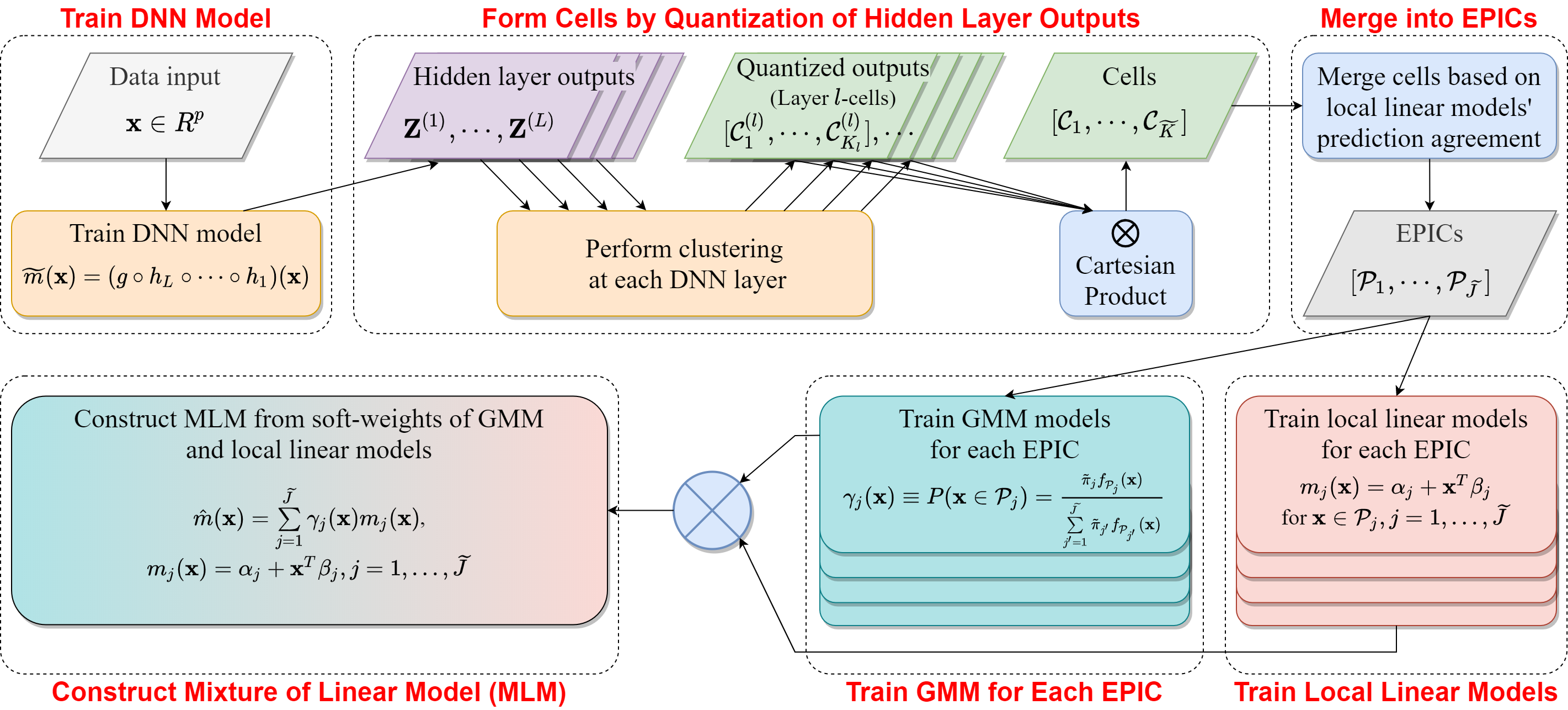}
	\caption{A schematic plot showing the steps of creating MLM.}
		\label{fig:mlm}
\end{center}
\end{figure}

Our core idea is to approximate the prediction function $\widetilde{m}(\mathbf{x})$ of a DNN model by a piecewise linear function $\hat{m}(\mathbf{x})$ called {\it Mixture of Linear Models} ({\it MLM}). We will present in the context of regression. Adaption to classification will be remarked upon later. Suppose the input sample space $\mathcal{X}$ is divided into $K$ mutually exclusive and collectively exhaustive sets, $\{\mathcal{P}_1,...,\mathcal{P}_{K}\}$, which form a {\it partition} of $\mathcal{X}$. Denote the partition by $\mathscr{P}=\{\mathcal{P}_1,...,\mathcal{P}_{K}\}$. The partition $\mathscr{P}$ induces a row-wise separation of the input data matrix $\mathbf{X}$ into $K$ sub-matrices: $\mathbf{X}^{(1)}$, ...,  $\mathbf{X}^{(K)}$.
Within each $\mathcal{P}_k$, $\widetilde{m}(\mathbf{x})$ is approximated by $m_k(\mathbf{x})$, which is referred to as the {\it local linear model}. To create $\mathscr{P}$, our criterion is not based on the proximity of $\mathbf{x}_i$'s but on the similarity between $\widetilde{m}(\mathbf{x})$ in the neighborhood of each $\mathbf{x}_i$. Because of this difference from the conventional mixture model, we are motivated to cluster $\mathbf{x}_i$'s by simulating from the prediction function of an estimated DNN. Without the guidance of the DNN, there will not be enough training points in the neighborhood of any $\mathbf{x}_i$ to permit a reasonable estimation of the prediction function unless the dimension is very low.

Let $I_{\mathcal{P}}(\mathbf{x})$ be the indicator function that equals $1$ when $\mathbf{x}\in\mathcal{P}$ and zero otherwise.
Given the partition $\mathscr{P}$, we express $\hat{m}(\mathbf{x})$ by
\begin{eqnarray}
\hat{m}(\mathbf{x})&=&
 I_{\mathcal{P}_{1}}(\mathbf{x})m_1(\mathbf{x})+\cdots +I_{\mathcal{P}_{K}}(\mathbf{x})m_{K}(\mathbf{x}),\nonumber\\
m_k(\mathbf{x}) &=& \alpha_k + \mathbf{x}^T\mathbf{\beta}_k, \qquad k=1,...,K.
\label{eq:mhat}
\end{eqnarray}

A schematic plot is provided in Figure~\ref{fig:mlm} to show the major steps in generating an MLM. The partition $\mathscr{P}$ is obtained by a two-stage process. First, clustering is performed based on the outputs at each layer of the DNN. This stage provides refined clusters, referred to as {\it cells}, of the data points and helps reduce computation at the next stage. Secondly, the cells are merged into larger clusters based on the similarity of $\widetilde{m}(\mathbf{x})$ computed via simulations. Clusters obtained at this stage become $\mathcal{P}_k$'s in $\mathscr{P}$.
We refer to $\mathcal{P}_k$ as an \textit{EPIC} (\textit{Explainable Prediction-induced Input Cluster}). 
A linear model is then fitted based on the data points in each EPIC as well as simulated data generated using the DNN. GMM models are then formed for each EPIC so that test data can be classified into the EPICs. Finally, a soft-weighted MLM modified from Eq. (\ref{eq:mhat}) is used as $\hat{m}(\mathbf{x})$.

The interpretability of MLM relies on both the local linear models $m_k(\mathbf{x})$ and the characterization of EPICs. We develop a visualization method and a decision-tree based method to help users explain EPICs and to open avenues for potential discovery. These methods are presented in Section~\ref{sec:interpret}. Next, we describe the steps to build an MLM.

\subsection{Construction of a Mixture of Linear Models}
\label{sec:MLM}

In Subsection~\ref{sec:linear_approx}, an approach is presented to approximate a neural network by a piecewise linear model and obtain cells. The approach to merge cells and obtain EPICs is described in  Subsection~\ref{sec:merging_EPIC}. Finally, we describe the soft-weighted form of MLM in Subsection \ref{sec:smooth_EPIC}.

\subsubsection{Piecewise Linear Approximation of DNN}
\label{sec:linear_approx}

The neural network model $\widetilde{m}(\mathbf{x})$ in Eq. (\ref{eqn:neuralnet}) is non-linear due to the non-linear activation functions $\sigma_l(\cdot)$ at each hidden layer. If the outputs at each layer are partitioned such that the non-linearity can be neglected within a cluster, the mapping of DNN from the original input to the prediction is approximately linear for data points that belong to the same cluster across all the layers. 
{\revision 
It is difficult to find such a partition directly from the original feature space $\mathcal{X}$ because points yielding similar outputs through the DNN layers are not necessarily close in $\mathcal{X}$. On the other hand, because the $l$-th hidden layer output $\mathbf{z}^{(l)}$ in DNN is formed by the composite of monotone functions and linear functions of the input, clusters generated based on output $\mathbf{z}^{(l)}$ correspond to polytopes in the input space. Within each polytope, the mapping from the input to the output is approximately linear. After we conduct clustering of $\mathbf{z}^{(l)}$ at each layer, the final cluster is determined by the sequence of clusters through the layers. Within each final cluster, the DNN mapping from the input to the output at every layer 
is approximately linear, and thus the overall mapping is linear.}

Let $\{\mathcal{C}^{(l)}_1,\cdots,\mathcal{C}^{(l)}_{K_l}\}$ be a partition of the $l$-th layer outputs of the DNN, where $K_l$ is the number of clusters. We call this partition \textit{layer $l$-cells}. Then, with layer $l$-cells, we can approximate the $l$-th hidden layer map $h_l(\mathbf{z}^{(l-1)})$ as a piecewise linear function:
\begin{eqnarray}
\hat{h}_l(\mathbf{z}^{(l-1)}) = I_{\mathcal{C}^{(l)}_1}(\mathbf{z}^{(l-1)}) (\mathbf{W}^{(1)}_1\mathbf{z}^{(l-1)}+\mathbf{b}^{(1)}_1)+\cdots + I_{\mathcal{C}^{(l)}_{K_l}}(\mathbf{z}^{(l-1)}) (\mathbf{W}^{(l)}_{K_l}\mathbf{z}^{(l-1)}+\mathbf{b}^{(l)}_{K_l}),
\label{eqn:linear_layer}
\end{eqnarray}
where $\mathbf{W}^{(l)}_k \in \mathbb{R}^{p_l\times p_{l-1}}$ and $\mathbf{b}^{(l)}_k \in \mathbb{R}^{p_l}$ are model parameters, and $I_{\mathcal{C}_k^{(l)}}(\mathbf{z}^{(l-1)})$ is the indicator function.

The layer $l$-cells are obtained by clustering the observed $l$-th hidden layer outputs. Denote by $\{\mathbf{z}^{(l)}_i\}_{i=1}^n$ the observed $l$-th hidden layer outputs corresponding to $\{\mathbf{x}_i\}_{i=1}^n$ respectively, that is, $\mathbf{x}_i\rightarrow \mathbf{z}_i^{(l)}$. We apply GMM-based clustering on $\{\mathbf{z}^{(l)}_i\}_{i=1}^n$ with $K_l$ clusters, and obtain the following GMM model. Denote the density function by $f(\cdot)$, and the prior, mean,  and covariance matrix of each Gaussian component by $\pi^{(l)}_k$, $\mu_{k}^{(l)}$, and $\Sigma_{k}^{(l)}$. Then
\begin{eqnarray}
f(\mathbf{z}^{(l)}) = \sum\limits_{k=1}^{K_l} \pi^{(l)}_k \phi(\mathbf{z}^{(l)}|\mu_{k}^{(l)}, \Sigma_{k}^{(l)}).
\label{eqn:gmm_layer}
\end{eqnarray}

As usual, the maximum a posteriori (MAP) criterion is applied to cluster $\mathbf{z}^{(l)}_i$'s based on the GMM.
We apply GMM clustering on each hidden layer separately to compute layer $l$-cells. Denote the set of cluster labels for the layer $l$-cells by $\mathcal{K}_l = \{1,\cdots,K_l\}$. The set of all possible sequences of the cluster labels across the $L$ layers is the Cartesian product $\widetilde{\mathcal{K}}=\mathcal{K}_1\times\cdots\times\mathcal{K}_L$. Let the cardinality of $\widetilde{\mathcal{K}}$ be 
$\widetilde{K}_0=\prod\limits_{l=1}^L K_l$. For each sequence in $\widetilde{\mathcal{K}}$, $(k_1, ..., k_L)$, $k_l\in \mathcal{K}_l$,  assign it a label $k$, $1\leq k\leq \widetilde{K}_0$, according to the lexicographic order. Denote the mapping by $(k_1, ..., k_L)\rightarrow k$. Then, the original input $\mathbf{x}_i$ belongs to cluster $\mathcal{C}_k$ if $(k_1, ..., k_L)\rightarrow k$ and $\mathbf{z}_i^{(l)}\in \mathcal{C}_{k_l}^{(l)}$ for $l=1, ..., L$. It can often occur that no $\mathbf{x}_i$ corresponds to a particular sequence in $\widetilde{\mathcal{K}}$. Thus the actual number of clusters formed, denoted by $\widetilde{K}$, is smaller than $\widetilde{K}_0$.

We call the clusters $\mathcal{C}_k$ \textit{cells}. After obtaining the cells, we estimate a linear model for each cell. As in Eq.(\ref{eq:mhat}), the linear model for the $k$th cell is $m_k(\mathbf{x})$. To train each local linear model $m_k(\mathbf{x})$, we use both the original data points that belong to a cell $k$ and simulated data points perturbed from the original points with responses generated by the DNN model $\widetilde{m}(\mathbf{x})$. A major reason for adding the simulated sample is that a cell usually does not contain sufficiently many points for estimating $m_k(\mathbf{x})$, the very difficulty of high dimensions. By the same rationale, we attempt to mimic the decision of the DNN, which is an empirically strong prediction model trained using the entire data. If the DNN can be well approximated, we expect MLM to perform strongly as well. Estimating a linear model using simulated data generated by the DNN is a data-driven approach to approximate the DNN locally. We thus aptly call this practice {\it co-supervision} by DNN.

Let $n_k$ be the number of observations in $\{\mathbf{x}_i\}_{i=1}^n$ that belong to cell $\mathcal{C}_k$. We have $\sum\limits_{k=1}^{\widetilde{K}} n_k = n$. Without loss of generality, let $\{\mathbf{x}^\prime_{k,i}\}_{i=1}^{n_k}$ be the set of points that belong to cell $\mathcal{C}_k$: $\{\mathbf{x}^\prime_{k,i}\}_{i=1}^{n_k} = \{\mathbf{x}_j|\mathbf{x}_j\in \mathcal{C}_k \text{ for } j=1,\cdots,n\}$ with an arbitrary ordering of $\mathbf{x}^\prime_{k,1},\cdots,\mathbf{x}^\prime_{k,{n_k}}$, and let $y^\prime_{k,i}$ for $i=1,\cdots,n_k$ be the corresponding dependent variable of $\mathbf{x}^\prime_{k,i}$ for $i=1,\cdots,n_k$. We denote by $\bar{\mathbf{x}}^\prime_k$ the sample mean of $\mathbf{x}^\prime_{k,1},\cdots,\mathbf{x}^\prime_{k,{n_k}}$: $\bar{\mathbf{x}}^\prime_k = \frac{\sum_{i=1}^{n_k} \mathbf{x}^\prime_{k,i}}{n_k}$. Then we generate $m$ perturbed sample points $\mathbf{v}_{k,1},\cdots,\mathbf{v}_{k,m}$ by adding Gaussian noise to the mean $\bar{\mathbf{x}}^\prime_k$ with a pre-specified variance parameter $\epsilon$ for each cell $k=1,\cdots,\widetilde{K}$. That is,
\begin{eqnarray}
\mathbf{v}_{k,i} \sim \mathcal{N}(\bar{\mathbf{x}}^\prime_k, \epsilon{\revision I_p}),\text{ for } i=1,\cdots,n_k.
\label{eqn:perturbed_samp}
\end{eqnarray}
For the perturbed sample points $\{\mathbf{v}_{k,i}\}_{i=1}^{m}$, the predicted dependent variable $\{w_{k,i}\}_{i=1}^{m}$ is computed using the DNN model $\widetilde{m}(\cdot)$: 
\begin{eqnarray}
w_{k,i} = \widetilde{m}(\mathbf{v}_{k,i}), \;\; i=1,\cdots,m.
\label{eqn:perturbed_target}
\end{eqnarray}
In the case of classification, the prediction of DNN is the probabilities of the classes. We then generate a class label by taking the maximum. 
We combine the original sample points and the simulated ones,
\[
\{(\mathbf{x}^\prime_{k,i}, y^\prime_{k,i})\}_{i=1}^{n_k}\bigcup 
\{(\mathbf{v}_{k,i}, w_{k,i})\}_{i=1}^{m} \]
to train $m_k(\mathbf{x})$ for $k=1, ...,\widetilde{K}$.
For each local linear model, if the dimension $p$ of $X$ is large, we can apply a penalized method such as LASSO to select the variables. For the brevity of presentation below, we introduce unified notations for the original data and the simulated data within each cell: $\mathbf{v}^\prime_{k,i}=\mathbf{x}^\prime_{k,i}$, $i=1, ..., n_k$, $\mathbf{v}^\prime_{k,n_k+i}=\mathbf{v}_{k,i}$, $i=1, ..., m$, $w^\prime_{k,i}=y^\prime_{k,i}$, $i=1, ..., n_k$, $w^\prime_{k,n_k+i}=w_{k,i}$, $i=1, ..., m$.

We need to pre-choose the hyperparameters $K_l$, $l=1,\cdots,L$, $m$, and $\epsilon$. In our experiments, we set $m=100$ and $\epsilon\leq 0.1$, the particular value of which is selected {\revision to minimize the cross-validation error for the training data. In general, when the data dimension is high, we tend to use a larger $m$ to prevent multicollinearity when estimating the local linear models.} 
 For simplicity, we set $K_1=K_2=\cdots K_L$ and choose $K_1$ using cross-validation. 

{\revision This method of constructing a piecewise linear function from DNN can be viewed as model
regularization. The DNN prediction function is piecewise linear when ReLU is used, usually the number of pieces being enormous. Roughly speaking, the clustering algorithm combines these linear pieces into a smaller
number of regions, and consequently, the model based on the combined regions reduces the complexity of the original DNN.}


\subsubsection{MLM based on EPICs}
\label{sec:merging_EPIC}

The complexity of MLM in Eq. (\ref{eq:mhat}) depends primarily on the number of local linear models. If we fit a local linear model for each cell, that is, to treat a cell directly as a EPIC, we usually end up with too many local models. To interpret the model $\hat{m}(\mathbf{x})$, we must interpret the EPICs. The task is easier if there are a smaller number of EPICs. Therefore, we further merge the cells by hierarchical clustering to generate the EPICs. The similarity between two cells is defined by the similarity of their corresponding local linear models. 

For classification, we define $d_{s,t}$ using the inverse of F1-score \citep{Rijsbergen1979information}:
$\displaystyle d_{s,t} = \frac{fp+fn}{2\cdot tp}$,
where 
\begin{eqnarray*}
tp = \left |\{\mathbf{v}^\prime_{k,i}|m_s(\mathbf{v}^\prime_{k,i})=1 \text{ and } m_t(\mathbf{v}^\prime_{k,i})=1, \text{ for }i=1,\cdots,n_k+m, \, k=s,t\}\right |\, ,
\\
fp = \left |\{\mathbf{v}^\prime_{k,i}|m_s(\mathbf{v}^\prime_{k,i})=1 \text{ and } m_t(\mathbf{v}^\prime_{k,i})=0, \text{ for }i=1,\cdots,n_k+m, \, k=s,t\}\right |\, ,
\\
fn = \left |\{\mathbf{v}^\prime_{k,i}|m_s(\mathbf{v}^\prime_{k,i})=0 \text{ and } m_t(\mathbf{v}^\prime_{k,i})=1, \text{ for }i=1,\cdots,n_k+m, \, k=s,t\} \right |\, ,
\end{eqnarray*}
and $|\cdot |$ denotes the cardinality of a set. 

Treating $d_{s,t}$ as a distance between the $s$th and $t$th cells, $s, t\in \{1,\cdots,\widetilde{K}\}$, we apply hierarchical clustering, specifically, Ward's linkage~\citep{ward1963hierarchical}, to merge the cells $\mathcal{C}_1,\cdots,\mathcal{C}_{\widetilde{K}}$ into $\widetilde{J}$ clusters. {\revision Ward's linkage generates a dendrogram by recursively computing the distance between a newly merged cluster $\nu$ of $s$ and $t$, and another cluster $\upsilon$ using the following equation:}
\begin{eqnarray*}
\revision
d_{\nu,\upsilon} = \sqrt{\frac{|\upsilon|+|s|}{T} d_{\upsilon,s}^2+\frac{|\upsilon|+|t|}{T} d_{\upsilon,t}^2-\frac{|\upsilon|}{T} d_{s,t}^2} \; ,
\end{eqnarray*}
{\revision where $T = |\upsilon|+|s|+|t|$.} {\revisionii Ward's linkage is known to perform well for clusters with spherical multivariate normal distributions \citep{kaufman2009finding}.} {\revision We provide users the option of different linkage schemes in our python package.} We assume $\widetilde{J}$ is user specified.
Let $\mathcal{J}_j$ be the set of the indices of cells that are merged into the $j$th cluster, $j=1, ..., \widetilde{J}$. For example, if cell $\mathcal{C}_1$ and $\mathcal{C}_2$ are merged into the first cluster, then $\mathcal{J}_1=\{1,2\}$. Clearly
 $\bigcup\limits_{j=1}^{\widetilde{J}} \mathcal{J}_j = \{1,\cdots,\widetilde{K}\}$. We define the merged cells as an EPIC. For $j=1, ..., \widetilde{J}$, 
\begin{eqnarray}
\mathcal{P}_j = \bigcup\limits_{k\in \mathcal{J}_j} \mathcal{C}_k \, .
\label{eqn:EPIC}
\end{eqnarray}
The original data and the simulated data contained in $\mathcal{P}_j$ form the set
$\displaystyle
\bigcup_{k\in \mathcal{J}_j} \{(\mathbf{v}^\prime_{k,i},w^\prime_{k,i})\}_{i=1}^{n_k+m}$,
based on which we refit a local linear model for the EPIC. With a slight abuse of notation, we still denote each local linear model by $m(\mathbf{x})$:
\begin{eqnarray}
\hat{m}(\mathbf{x}) &=& I_{\mathcal{P}_{1}}(\mathbf{x})m_1(\mathbf{x})+\cdots +I_{\mathcal{P}_{\widetilde{J}}}(\mathbf{x})m_{\widetilde{J}}(\mathbf{x}),\nonumber\\
m_j(\mathbf{x}) &=& \alpha_j + \mathbf{x}^T\mathbf{\beta}_j,\text{ for } j=1,...,\widetilde{J}\, .
\label{eqn:piecewise_linear_EPIC}
\end{eqnarray}


\subsubsection{Soft-weighted MLM based on EPICs}
\label{sec:smooth_EPIC}


Although Eq. (\ref{eqn:piecewise_linear_EPIC}) is used to fit the training data, it is not directly applicable to new test data because which $\mathcal{P}_j$ a test point belongs to is unknown. We need a classifier for the EPICs: $\mathcal{P}_1$, ..., $\mathcal{P}_{\widetilde{J}}$. Given how the EPICs are generated in training, one seemingly obvious choice is to compute the DNN inner layer outputs for the test data and associate these outputs to the trained cells and subsequently EPICs. However, this approach hinders us from interpreting the overall MLM because categorization into the EPICs requires complicated mappings of a DNN model, even though the local linear models in MLM are relatively easy to interpret. We thus opt for an easy to interpret classifier for EPICs. Instead, we will build a classifier for the EPICs using the original independent variables. In Subsection~\ref{sec:interpret}, we will also develop ways for visualization and rule-based descriptions of EPICs.

From now on, we treat the partition of the training data $\mathbf{x}_i$, $i=1, ..., n$, into the EPICs $\mathcal{P}_1$, ..., $\mathcal{P}_{\widetilde{J}}$  as the ``ground truth'' labels when discussing classification of EPICs based on the original variables. Denote the labels by $\zeta_i$. We have $\mathbf{x}_i\in\mathcal{P}_{\zeta_i}$. We first construct a GMM directly from the cells $\mathcal{C}_1$, ..., $\mathcal{C}_{\widetilde{K}}$ by fitting a single Gaussian density for each cell. Denote the estimated prior, mean, and covariance matrix of $\mathcal{C}_k$ by $\hat{\pi}_k$, $\hat{\mu}_{k}$, and $\hat{\Sigma}_{k}$, $k=1, ..., \widetilde{K}$. The estimated prior $\hat{\pi}_k$ is simply the empirical frequency, and $\hat{\mu}_{k}$ is given by the sample mean. The covariance $\hat{\Sigma}_{k}$ can be estimated with different types of structural constraints, {\it e.g.}, diagonal, spherical, or pooled covariance across components. Recall that $\mathcal{P}_j$ contains cells $\mathcal{C}_k$ with $k\in\mathcal{J}_j$. Let $\phi(\cdot)$ be the Gaussian density. Let the estimated prior of $\mathcal{P}_j$ be $\tilde{\pi}_j=\sum_{k\in\mathcal{J}_j}\hat{\pi}_k$.
The density of $X$ given that $X\in\mathcal{P}_j$ is
\begin{eqnarray}
f_{\mathcal{P}_j}(\mathbf{x})=
\sum_{k\in\mathcal{J}_j}\frac{\hat{\pi}_k}{\tilde{\pi}_j}\cdot \phi(\mathbf{x}\mid \mu_{k},\Sigma_k) \, , \quad j=1, ..., \widetilde{J}.
\label{eq:pjdensity}
\end{eqnarray}
The posterior $P(X\in\mathcal{P}_j\mid X=\mathbf{x})\propto \tilde{\pi}_j f_{\mathcal{P}_j}(\mathbf{x})$ is used as the weight for the local linear models in MLM. Let the posterior for $\mathcal{P}_j$ be $\displaystyle \gamma_j(\mathbf{x})=\frac{\tilde{\pi}_j f_{\mathcal{P}_j}(\mathbf{x})}{\sum_{j'=1}^{\widetilde{J}}\tilde{\pi}_{j'} f_{\mathcal{P}_{j'}}(\mathbf{x})}$. Then the soft-weighted MLM is
\begin{eqnarray}
\hat{m}(\mathbf{x}) &=& \sum_{j=1}^{\widetilde{J}} \gamma_j(\mathbf{x})
m_j(\mathbf{x}) \nonumber\\
m_j(\mathbf{x}) &=& \alpha_j + \mathbf{x}^T\mathbf{\beta}_j, \quad j=1,...,\widetilde{J}.
\label{eqn:smooth_linear}
\end{eqnarray}
The above soft-weighted MLM yields smoother transitions between the EPICs. As explained in Subsection~\ref{sec:merging_EPIC}, the local linear models $m_j(\mathbf{x})$ are fitted by least square regression using the original and simulated data in $\mathcal{P}_j$.
We note that the weights $\gamma_j(\mathbf{x})$ play a similar role as the kernel functions in Eq.(\ref{eq:kernel}). Instead of using  Gaussian density functions centered at each data point, we use $f_{\mathcal{P}_j}(\mathbf{x})$, the densities of the EPICs. Another important difference is that we use local linear models trained under the co-supervision of a DNN.


\subsection{Interpretation}
\label{sec:interpret}
To interpret MLM, we assume that the local linear model within each EPIC can be interpreted, or useful insight can be gained for each EPIC based on its local linear model. Although this assumption may not always hold depending on the dataset, our focus here is to interpret the EPICs. If we can understand the EPICs, we can better understand the heterogeneity across the sample space in terms of the relationship between the dependent and independent variables. 
Our experiments show that the heterogeneity across EPICs can be large, and MLM can achieve considerably higher accuracy than a single linear model.

To help understand EPICs, we develop two approaches, one based on visualization and the other based on descriptive rules. For the first approach, we aim at selecting a small number of variables based on which a EPIC can be well separated from the others. If such a small subset of variables exist, we can visualize the EPIC in low dimensions. The second approach aims at identifying easy-to-describe regions in the sample space that are dominated by one EPIC. We call the first approach the {\it Low Dimensional Subspace} ({\it LDS}) method and the second the {\it Prominent Region} ({\it PR}) method.
  
\subsubsection{Interpretation of EPIC by LDS}
\label{sec:LDS}

Recall that we model the density of $X$ in each EPIC, denoted by $f_{\mathcal{P}_j}(\mathbf{x})$, $j=1, ..., \widetilde{J}$, by a GMM in Eq.(\ref{eq:pjdensity}). The marginal density of $f_{\mathcal{P}_j}(\mathbf{x})$ on any subset of variables of $X$ can be readily obtained because the marginal density of any Gaussian component in the mixture is Gaussian. Denote a subset of variable indices by $\mathbf{s}$, $\mathbf{s}\subseteq \{1, ..., p\}$. Denote the subvector of $X$ specified by $\mathbf{s}$ by $X_{[\mathbf{s}]}$ and correspondingly the subvector of a realization $\mathbf{x}$ by $\mathbf{x}_{[\mathbf{s}]}$. Denote the marginal density of  $X_{[\mathbf{s}]}$ by $f_{\mathcal{P}_j, \mathbf{s}}(\mathbf{x}_{[\mathbf{s}]})$. 
Using the marginal densities $f_{\mathcal{P}_{j'}, \mathbf{s}}(\mathbf{x}_{[\mathbf{s}]})$, $j'=1, ..., \widetilde{J}$, and applying MAP, we can classify whether a sample point $\mathbf{x}_i$, $i=1, ..., n$, belongs to $\mathcal{P}_j$. Let $\hat{\mathbf{q}}_j=(\hat{q}_{j,1}, ..., \hat{q}_{j,n})$ be the indicator vector for $\mathbf{x}_i$ being classified to $\mathcal{P}_j$
based on the marginal densities of $X_{[\mathbf{s}]}$:
\[
\hat{q}_{j,1}=\left\{
\begin{array}{ll}
1 & \tilde{\pi}_j f_{\mathcal{P}_j, \mathbf{s}}(\mathbf{x}_{[\mathbf{s}]}) \geq
\sum_{j': j'\neq j}\tilde{\pi}_{j'} f_{\mathcal{P}_{j'}, \mathbf{s}}(\mathbf{x}_{[\mathbf{s}]})\\
0 & \mbox{otherwise} \, .
\end{array}
\right .
\]
Let $\mathbf{q}_j=(q_{j,1}, ..., q_{j,n})$ be the indicator for EPIC $\mathcal{P}_j$ based on the EPIC labels $\zeta_i$, that is, $q_{j,i}=1$ if $\zeta_i=j$, zero otherwise. 

\begin{table}[htp]
\centering
\begin{tabular}{lll}
\hline
\multicolumn{3}{l}{Algorithm to find explainable dimensions $\mathbf{s}_j^*$ for EPIC $\mathcal{P}_j$}\\
\hline
1 & \multicolumn{2}{l}{Set hyper-parameter $0<\xi<1$}\\
2 & \multicolumn{2}{l}{$\mathbf{s}_j^\dagger=\emptyset$; $r_j^\dagger=0$}\\
3 & \multicolumn{2}{l}{\textbf{Do while} $r_j^\dagger\leq \xi$ and $|\mathbf{s}_j^\dagger|< p$}\\
4 && Form a collection of sets $\mathcal{S} =\{\mathbf{s}|\mathbf{s} = \mathbf{s}_j^\dagger\bigcup s \text{ for any }s\in \{1,\cdots,p\}\setminus \mathbf{s}_j^\dagger\}$\\
5 && Compute $r_{\mathbf{s}} = F_1(\hat{\mathbf{q}}_j,\mathbf{q}_j)$ for all $\mathbf{s}\in\mathcal{S}$\\
6 && Update $\mathbf{s}_j^\dagger$: $\arg\max\limits_{\mathbf{s}\in \mathcal{S}}r_{\mathbf{s}}\rightarrow \mathbf{s}_j^\dagger$\\
7 && Update $r^\dagger_j$: $r_{\mathbf{s}_j^\dagger}\rightarrow r^\dagger_j$ \\
8 & \multicolumn{2}{l}{\text{Return} $\mathbf{s}_j^\dagger$ and $r^{\dagger}_j$ as $(\mathbf{s}_j^*,r_{\mathbf{s}^*_j})$ if $r^{\dagger}_j>\xi$. Otherwise, declare failure to find a valid $\mathbf{s}_j^*$.}\\
\hline
\end{tabular}
\caption{The algorithm to find explainable dimensions and the corresponding explainable rate for each EPIC.}
\label{tab:alg1}
\end{table}

We seek for a subset $\mathbf{s}^{*}_{j}$ such that the cardinality $|\mathbf{s}^{*}_{j}|$ is small and $\hat{\mathbf{q}}_j$ is close to $\mathbf{q}_j$, the disparity between them measured by the F1-score between binary classification results. Denote the F1-score by $F_1(\hat{\mathbf{q}}_j, \mathbf{q}_j)$, which is larger for better agreement between the binary vectors. In the algorithm presented in Table~\ref{tab:alg1}, we find $\mathbf{s}^{*}_{j}$ by step-wise greedy search. In a nutshell, the algorithm adds variables one by one to a set until $F_1(\hat{\mathbf{q}}_j, \mathbf{q}_j)>\xi$, where $0<\xi<1$ is a pre-chosen hyper-parameter. It is possible that the search does not yield any valid $\mathbf{s}^{*}_{j}$, which indicates that the EPICs cannot be accurately classified and thus easily interpreted.
We call variables in $\mathbf{s}^*_j$ \textit{explainable dimensions} for EPIC $\mathcal{P}_j$ and the F1-score obtained by $\mathbf{s}^*_j$ {\it explainable rate}, which is denoted by $r_{\mathbf{s}^{*}_j}$.

\subsubsection{Interpretation of EPIC by PR}
	\label{sec:PR}
	
In this subsection, we explore a more explicit way to characterize EPICs. Using a decision tree, we seek prominent regions that are dominated by points from a single EPIC and can be characterized by a few conditions on individual variables. Such descriptions of EPICs are more direct interpretation than visualization in low dimensions. However, the drawback is that prominent regions are not guaranteed to exist. Given a prominent region, researchers can pose a hypothesis specific to this region rather than for the whole population.

Let $\mathcal{D}$ be a decision tree trained for binary classification by CART~\citep{breiman1984classification}. Let the training class labels be $\mathbf{q}=\{q_i|q_i\in\{0,1\}, \; i=1,\cdots,n\}$ and input data matrix $\mathbf{X}\in \mathbb{R}^{n\times p}$. CART recursively divides the data by thresholding one variable at each split. A \textit{leaf node}, also called {\it terminal node}, is a node that is not split; a \textit{pure node} is a node that contains sample points from a single class; and a \textit{fully grown tree} is a tree whose leaf nodes either are pure nodes or contain a single point or multiple identical points. It is assumed that when a node becomes pure, it is not further split. Consider a 
node denoted by $\mathscr{e}$. Define the \textit{depth} of $\mathscr{e}$, $d(\mathscr{e})$, as the number of splits to traverse from the root node to $\mathscr{e}$. Define the \textit{size} of $\mathscr{e}$, $s(\mathscr{e})$, as the number of sample points contained in the node. Define the \textit{sample index set} $u(\mathscr{e})$ as the set of indices of sample points contained in $\mathscr{e}$, and the \textit{decision path} of $\mathscr{e}$, denoted by $\mathcal{H}$, as the sequence of split conditions to traverse from the root node to $\mathscr{e}$. $\mathcal{H}$ consists of $d(\mathscr{e})$ number of conditions each expressed by thresholding one variable, {\it e.g.}, $x_{\cdot, j}>a$ or $x_{\cdot, j}\leq a$.

Again let $\mathbf{q}_{j} = (q_{j,1},\cdots,q_{j,n})$ be the indicator vector for any sample point belonging to EPIC $\mathcal{P}_j$. We fit a fully grown decision tree $\mathcal{D}_j$ based on $\mathbf{X}$ and the class labels $\mathbf{q}_{j}$ (class 1 means belonging to $\mathcal{P}_j$). Let $\psi$ be a pre-chosen threshold, $0< \psi\leq 1$. We prune off the descendant nodes of $\mathscr{e}$ if its proportion of points in class 1 reaches $\psi$:
\begin{eqnarray}
\frac{\sum\limits_{i=1}^n I_{u(\mathscr{e})}(i) q_{j,i}}{s(\mathscr{e})} \geq \psi \, .
\label{eqn:purity}
\end{eqnarray}

We collect all the leaf nodes satisfying Eq.(\ref{eqn:purity}) 
with sizes above a pre-chosen minimum size threshold $\eta$. Since leaf nodes are not further split, they are ensured to contain non-overlapping points. Suppose there are $\kappa_j$ leaf nodes $\hat{\mathscr{e}}_\tau$, $\tau=1, ..., \kappa_j$, such that $s(\hat{\mathscr{e}}_{\tau})>\eta$ and $\hat{\mathscr{e}}_{\tau}$ satisfies Eq.(\ref{eqn:purity}). 
Denote the decision path of $\hat{\mathscr{e}}_{\tau}$ by $\mathcal{H}_{\tau}$, which contains $d(\hat{\mathscr{e}}_{\tau})$ conditions. In $\mathcal{H}_{\tau}$, one variable may be subject to multiple conditions (that is, this variable is chosen multiple times to split the data). We will identify the intersection region of multiple conditions applied to the same variable, which is in general a finite union of intervals. At the end, a decision path $\mathcal{H}_{\tau}$ specifies a region of the sample space $\mathcal{X}$ by the Cartesian product $\mathcal{R}_1\times \mathcal{R}_2\times\cdots \times \mathcal{R}_p$. If $\mathcal{R}_{j'}=\mathbb{R}$, then variable $X_{j'}$ does not appear in any condition of $\mathcal{H}_{\tau}$. Suppose $\mathcal{R}_{j'_1}$, ..., $\mathcal{R}_{j'_{p'}}$ are proper subsets of $\mathbb{R}$. The region given by $\mathcal{H}_{\tau}$ can be described by $\{X_{j'_1}\in \mathcal{R}_{j'_1}\}\cap \{X_{j'_2}\in \mathcal{R}_{j'_2}\}\cap \cdots \cap \{X_{j'_{p'}}\in \mathcal{R}_{j'_{p'}}\}$. We call this form of the region decided by $\mathcal{H}_{\tau}$ an {\it explainable condition}. 

When the depth of $\mathcal{H}_{\tau}$ is small, the number of conditions $p'$ in $\mathcal{H}_{\tau}$ will be small, and thus the explainable condition is simpler. For the sake of interpretation, simpler explainable conditions are preferred. It is also desirable to have large $s(\hat{\mathscr{e}}_{\tau})$, which means that many data points from $\mathcal{P}_j$ are covered by this explainable condition. Whether we can find simple explainable conditions that also have high coverage of points depends strongly on the data being analyzed. Hence, the search for explainable conditions can only be viewed as a tool to assist interpretation, but it cannot guarantee that simple interpretations will be generated. 
In practice, it may suffice to require that class 1 accounts for a sufficiently high percentage of points in $\hat{\mathscr{e}}_{\tau}$ by setting the purity parameter $\psi<1$. With this relaxation, we can find $\hat{\mathscr{e}}_{\tau}$'s with smaller depth.

\section{Experiments}
\label{sec:exp}

In this section, we present experimental results of the proposed methods. We demonstrate prediction accuracy of MLM based on cells (MLM-cell) and EPICs (MLM-EPIC), and interpret the formed EPICs by the LDS and PR methods. {\revisionii We use one synthesized dataset and five real datasets of different fields}: two {\revision The-Cancer-Genome-Atlas (TCGA)} datasets (Section \ref{sec:tcga}), one Parkinson's disease (PD)  clinical dataset (Section \ref{sec:pd}) for classification, the bike sharing demand data (Section \ref{sec:bike}), and California housing price data (Section \ref{sec:calhousing}) for regression.

We compare the prediction accuracy of MLM-cell and MLM-EPIC with the following methods: {\revision nearest class mean (NCM) (k-means is applied to the feature vectors to generate $100$ classes)~\citep{webb2003statistical}},
random forest (RF) \citep{breiman2001random}, support vector machine (SVM) \citep{cortes1995support}, multilayer perceptron (MLP), linear regression (LR) {\revision (specifically, logistic regression for KIRC, SKCM and PD data, and Poisson regression for Bike Sharing data)}, generalized additive model (GAM) \citep{hastie1990generalized}, Spatial Autoregressive (SAR) \citep{pysal2007} (for California housing price data that include spatial variables), multivariate adaptive regression splines (MARS) (only for regression tasks, i.e., bike sharing and California housing datasets) \citep{friedman1991multivariate}, {\revision mixture of expert neural networks (MOE) \citep{jacobs1991adaptive}}, {\revision cluster-weighted modeling(CWM) \citep{gershenfeld1997nonlinear} (for KIRC, SKCM, and PD data, CWM is not applied because the estimated covariance matrix is singular).} The following python packages are used: \textbf{pygam} for GAM, \textbf{pyearth} for MARS, \textbf{pysal} for SAR, and \textbf{scikit-learn} {\revision for LR, RF, SVM. CWM is provided by the \textbf{flexCWM} R package.}

{\revisionii
\subsection{EPIC Clusters with Synthesized Data}

\begin{figure}[h]
    \centering
    \begin{subfigure}[b]{0.24\textwidth}
         \centering
         \includegraphics[width=\textwidth]{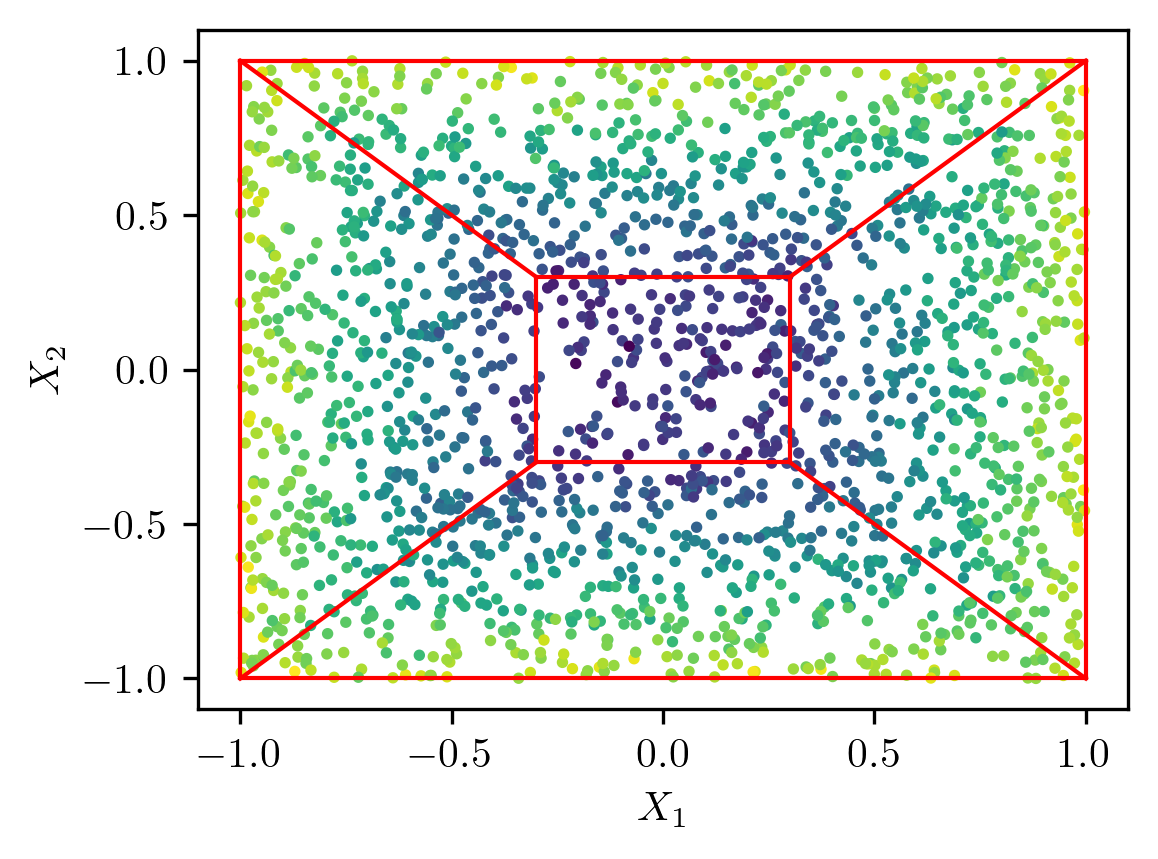}
         \caption{Ground Truth}
    \end{subfigure}
	\begin{subfigure}[b]{0.24\textwidth}
         \centering
         \includegraphics[width=\textwidth]{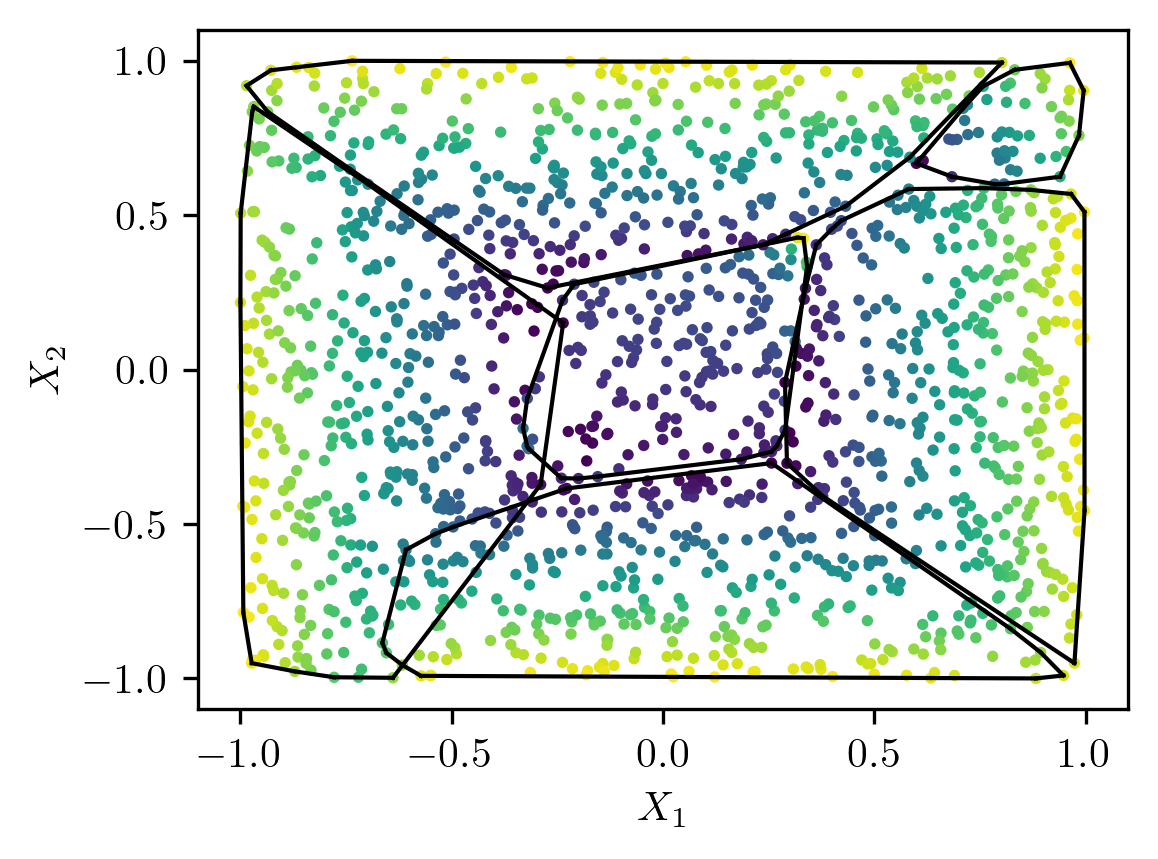}
         \caption{MLM}
    \end{subfigure}
	\begin{subfigure}[b]{0.24\textwidth}
         \centering
         \includegraphics[width=\textwidth]{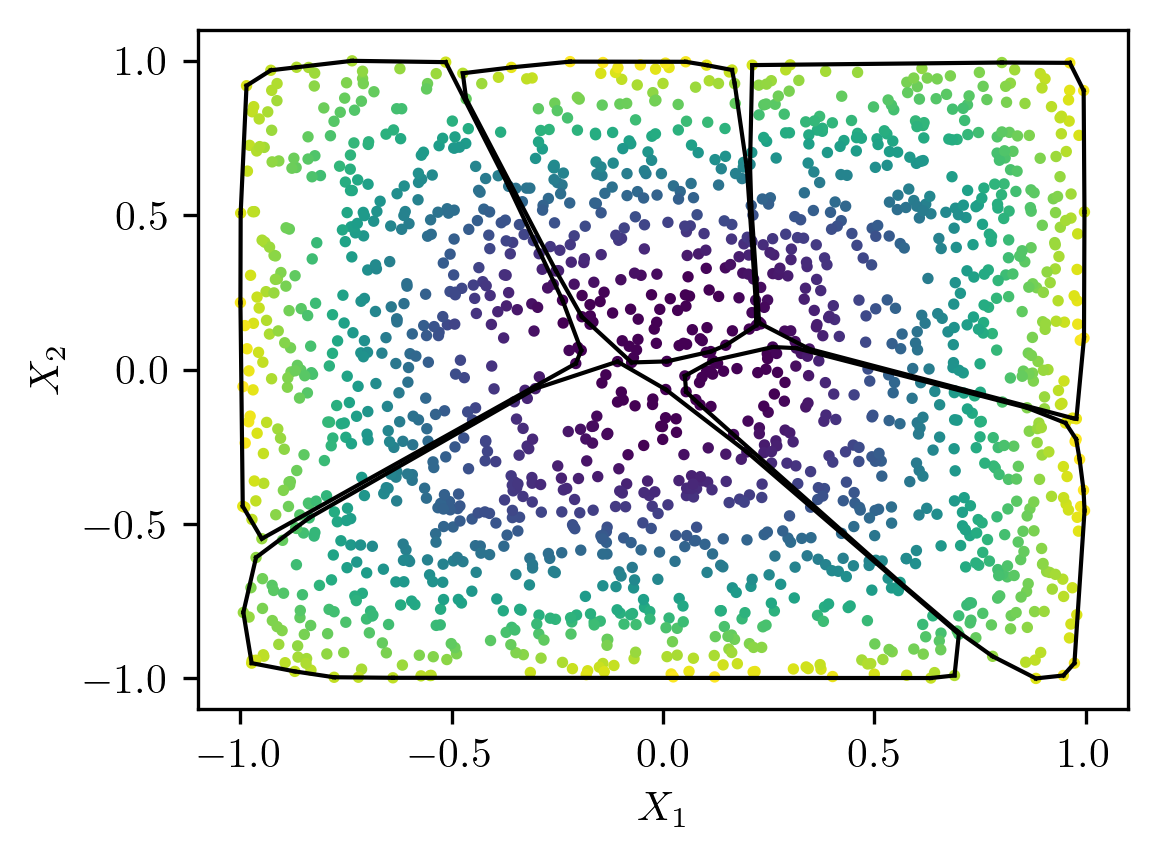}
         \caption{MOE}
    \end{subfigure}\\ 
	\centering
    \begin{subfigure}[b]{0.24\textwidth}
         \centering
         \includegraphics[width=\textwidth]{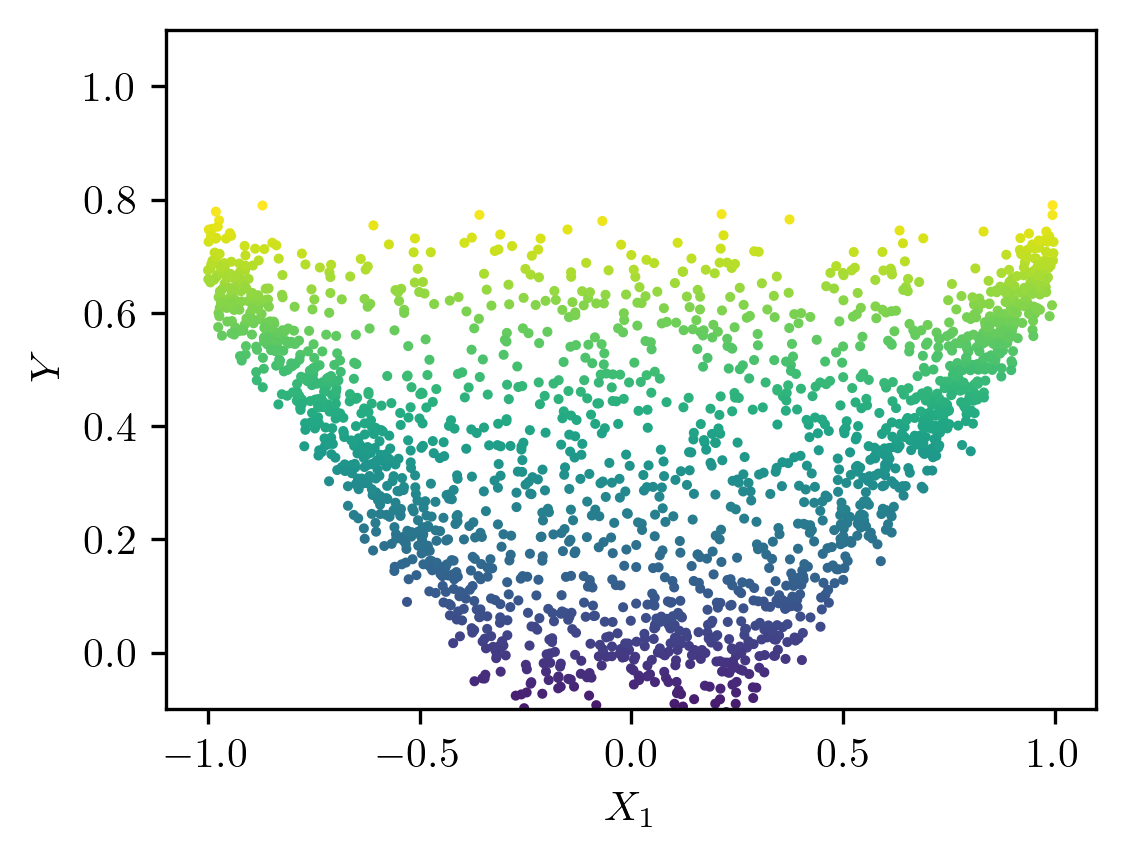}
         \caption{Ground Truth}
    \end{subfigure}
	\begin{subfigure}[b]{0.24\textwidth}
         \centering
         \includegraphics[width=\textwidth]{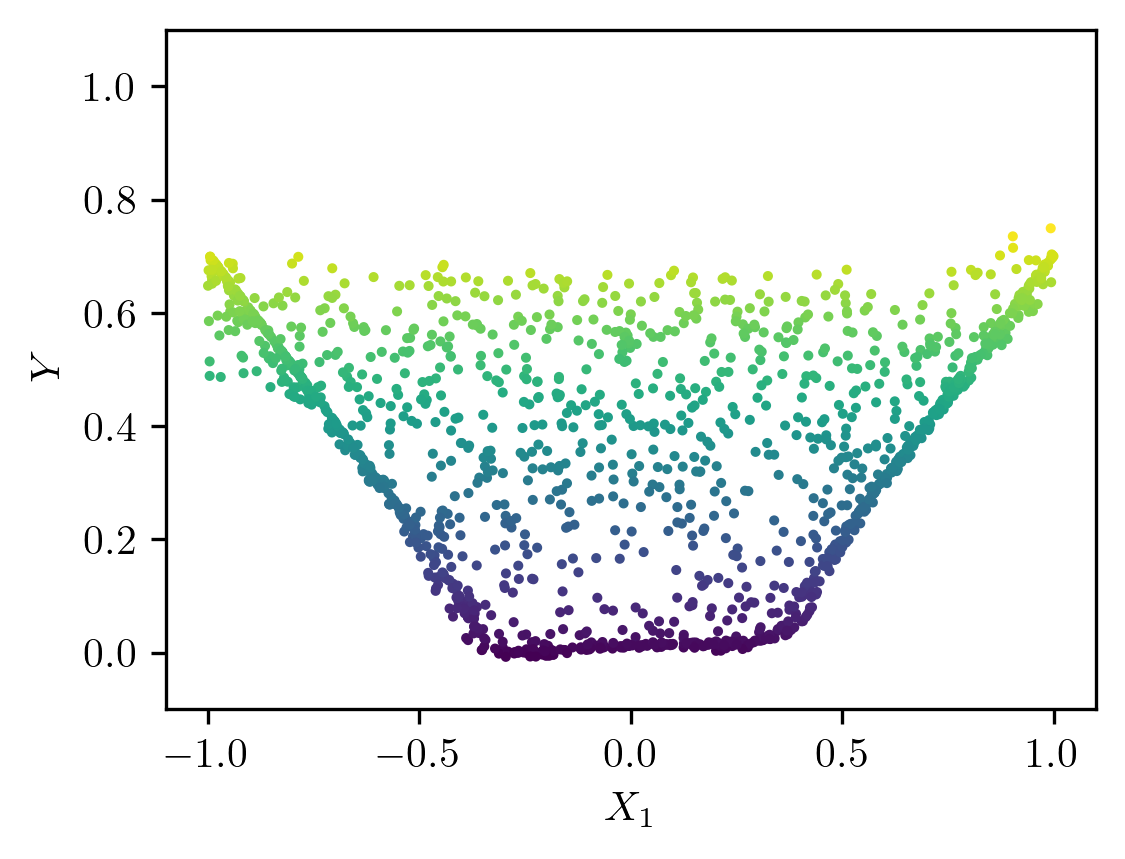}
         \caption{MLM}
    \end{subfigure}
	\begin{subfigure}[b]{0.24\textwidth}
         \centering
         \includegraphics[width=\textwidth]{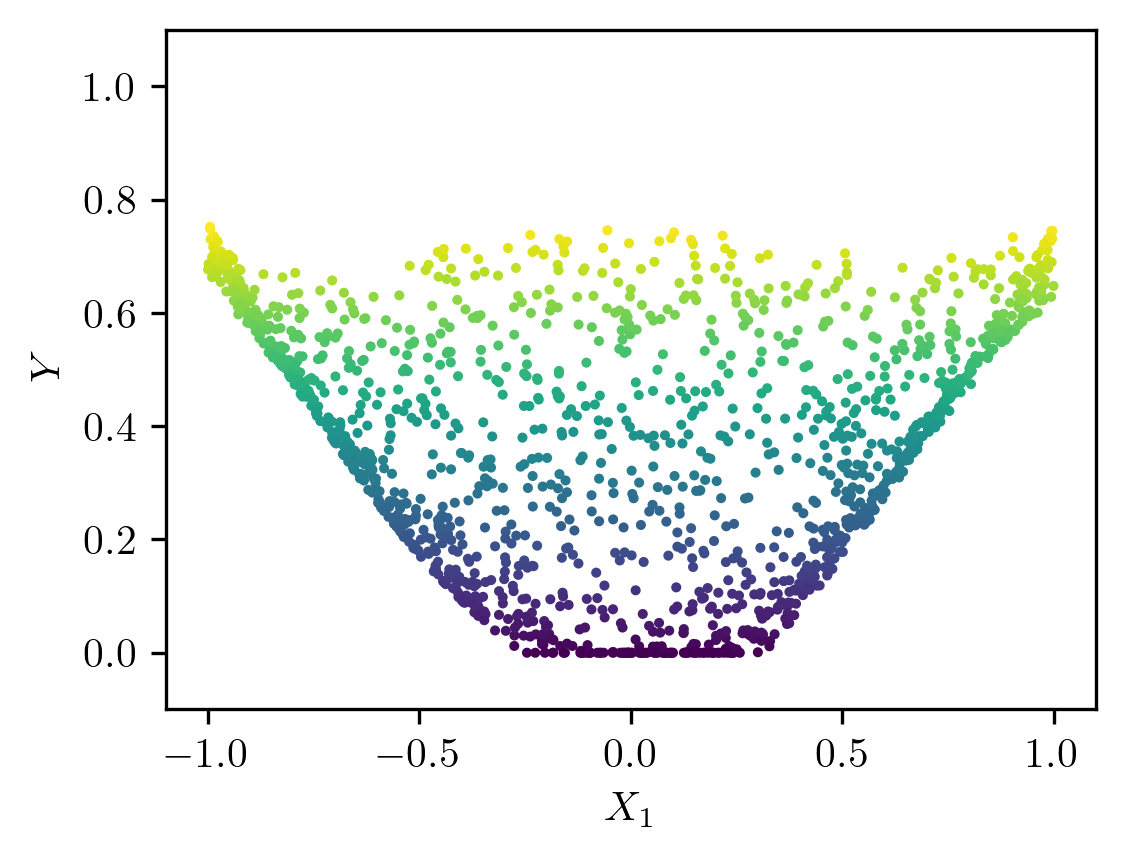}  \caption{MOE}
    \end{subfigure}
	\begin{subfigure}[b]{0.24\textwidth}
         \centering
         \includegraphics[width=\textwidth]{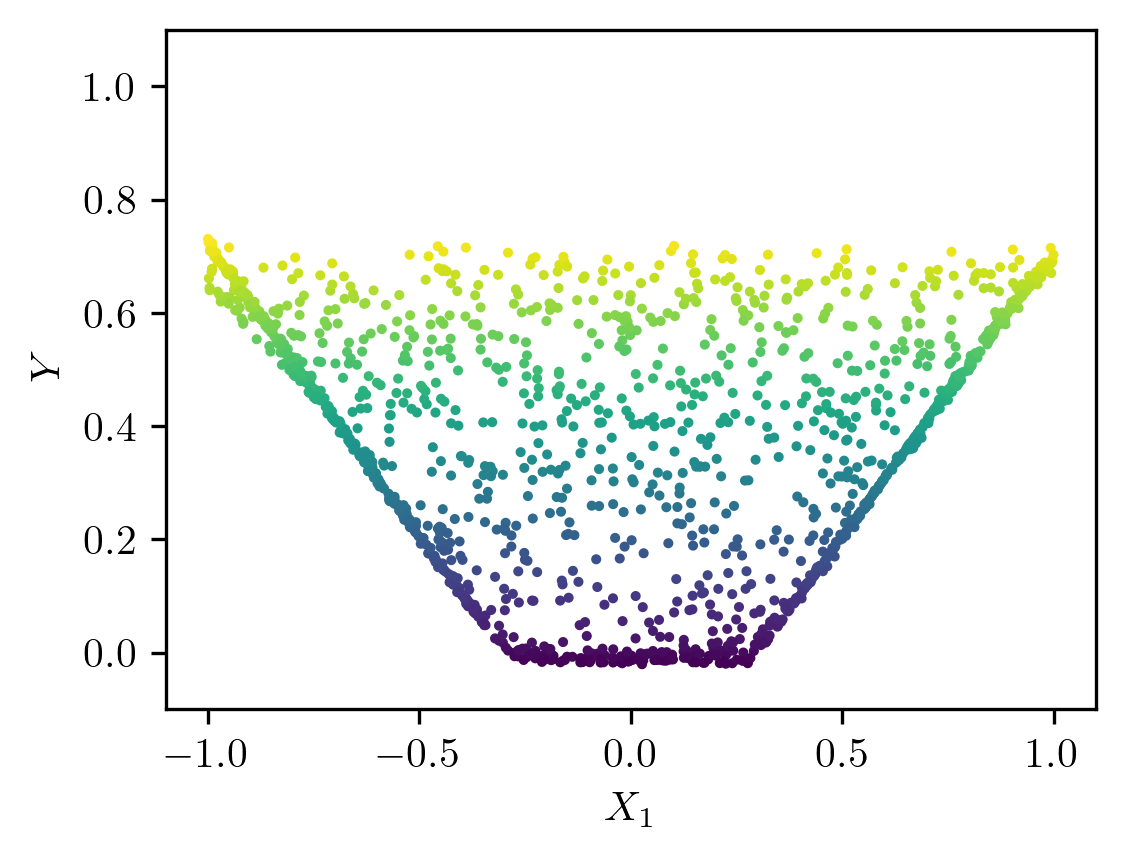}  \caption{DNN}
    \end{subfigure}\\
	\centering
    \begin{subfigure}[b]{0.24\textwidth}
         \centering
         \includegraphics[width=\textwidth]{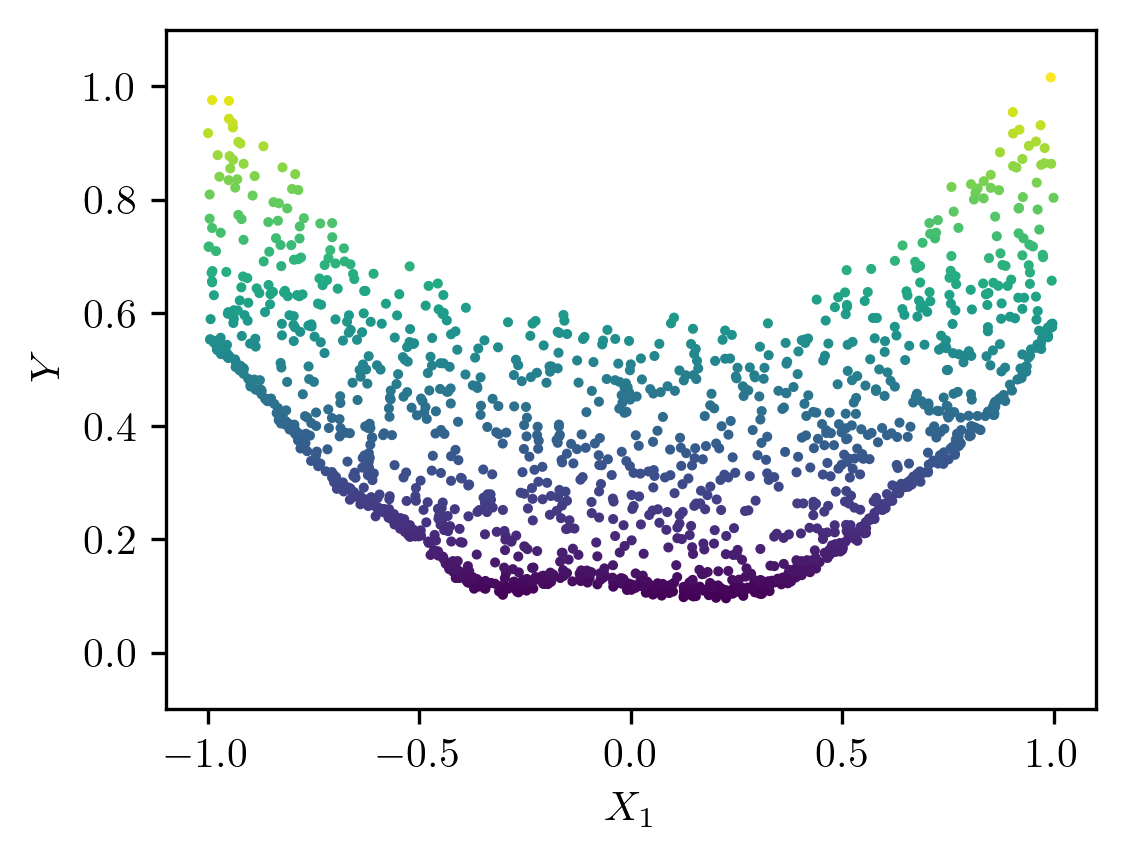}
         \caption{GAM}
    \end{subfigure}
	\begin{subfigure}[b]{0.24\textwidth}
         \centering
         \includegraphics[width=\textwidth]{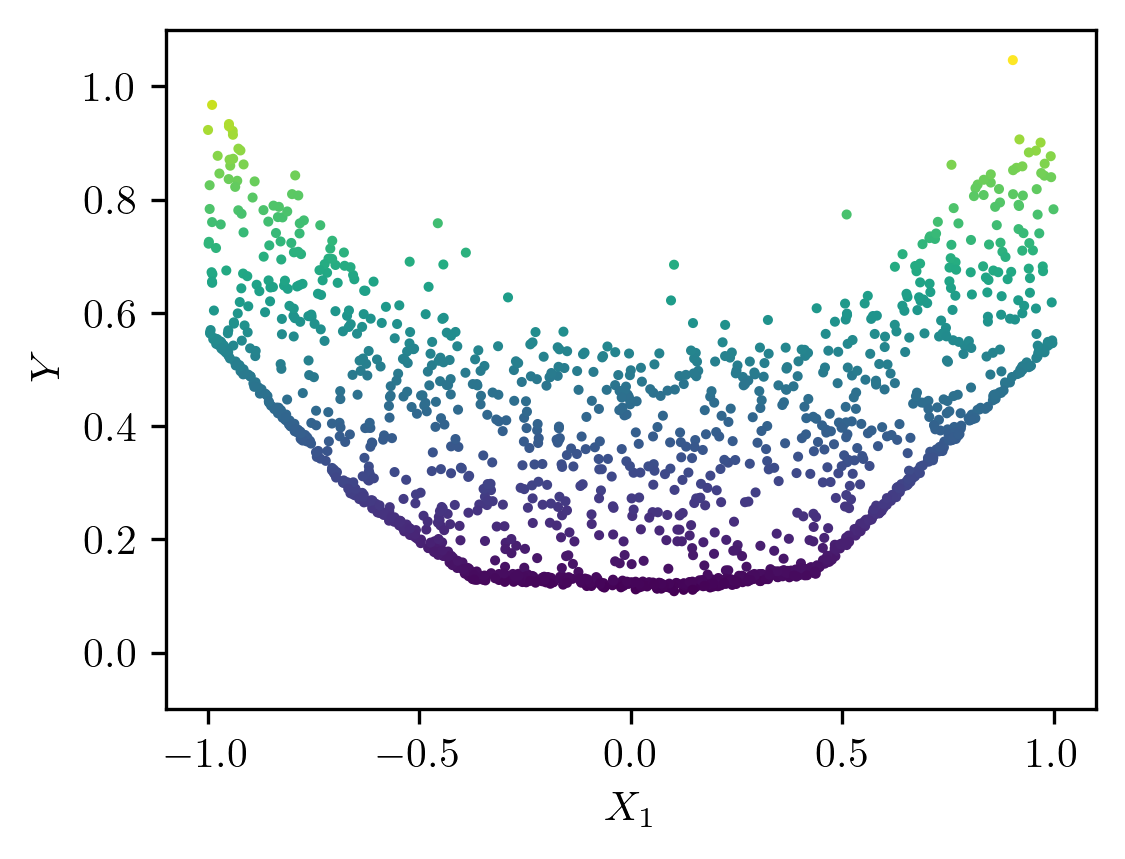}
         \caption{MARS}
    \end{subfigure}
	\begin{subfigure}[b]{0.24\textwidth}
         \centering
         \includegraphics[width=\textwidth]{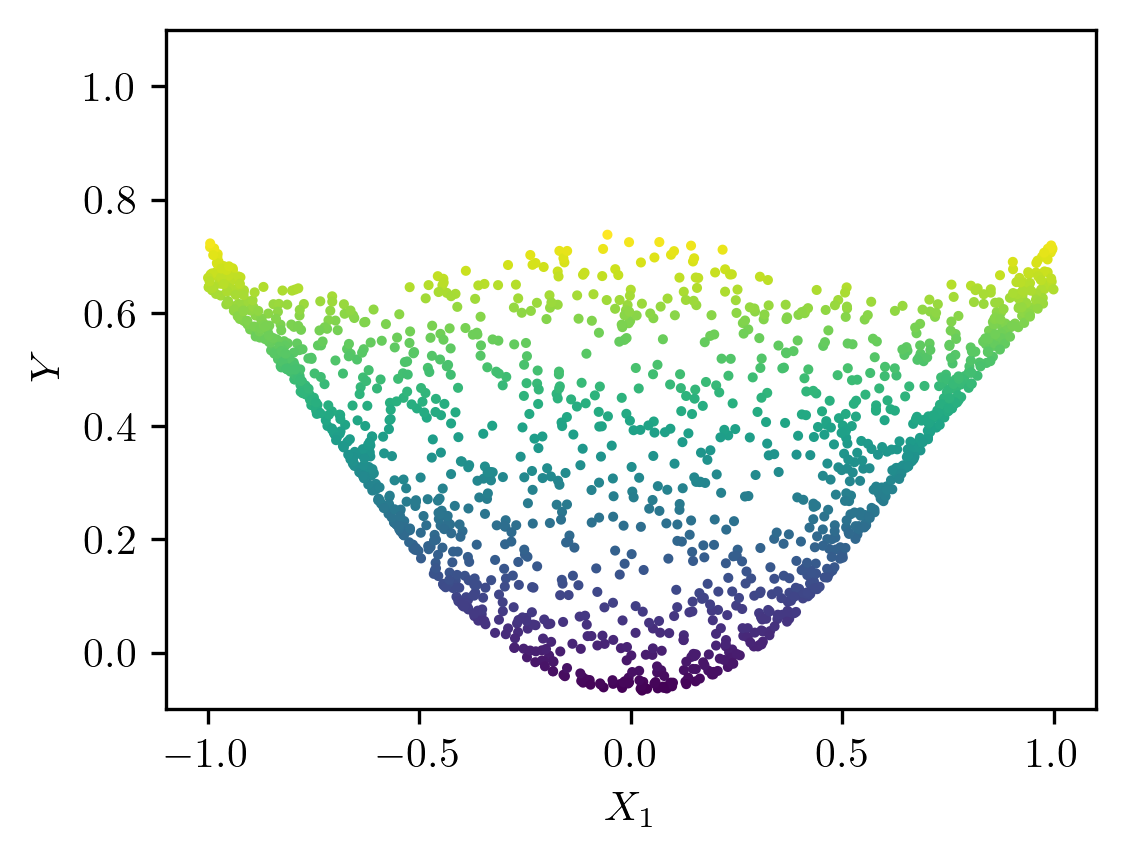}
         \caption{SVR(RBF kernel)}
    \end{subfigure}
	\begin{subfigure}[b]{0.24\textwidth}
         \centering
         \includegraphics[width=\textwidth]{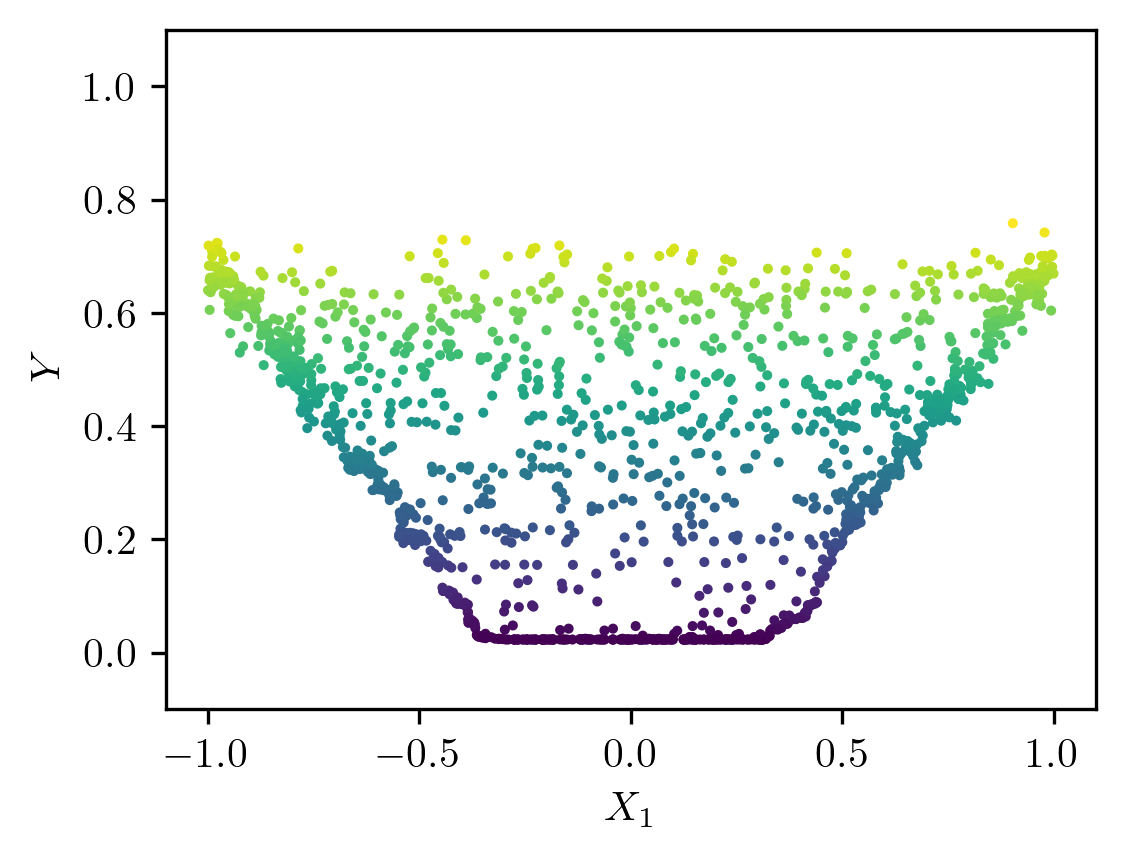}
         \caption{RF}
    \end{subfigure}
	\caption{\revision (a-c) Scatter plots on the two explaining variables, $X_1$ and $X_2$, with ground truth groups or convex hulls of estimated clusters for each model. (d-k) Scatter plots on the predicted values, $\hat{Y}$, and an explaining variable, $X_1$, for each model. }
	\label{fig:syndat1}
\end{figure}

We study a synthesized dataset with given clusters and examine whether MLM can identify those clusters via EPICs.
This dataset is generated by a uniform distribution on the square $[-1, 1]\times [-1,1]$ (two independent variables $X_1$ and $X_2$). Two thousand sample points are generated. The dependent variable $Y$ is computed from the following formula, bearing the shape of a trapezoid bowl plus random noise.
\begin{eqnarray}
y=\begin{cases}
X_2-0.3+e \text{ if } X_2-X_1>0 \text{ and } X_1+X_2>0\\
X_1-0.3+e \text{ if } X_2-X_1<0 \text{ and } X_1+X_2>0\\
-X_2-0.3+e \text{ if } X_2-X_1<0 \text{ and } X_1+X_2<0\\
-X_1-0.3+e \text{ if } X_2-X_1>0 \text{ and } X_1+X_2<0\\
0+e \text{ otherwise, }
\end{cases} 
\end{eqnarray}
where $e$ is a Gaussian noise with standard deviation $0.05$. 

Clearly the best partition of splitting the sample points into local regions, which of each is modeled by a linear regression, is to follow the edges of the trapezoid bowl. Figure \ref{fig:syndat1}(b) shows that MLM generates EPICs in better agreement with the geometry of the trapezoid bowl than the local regions of MOE shown in (c). Figures \ref{fig:syndat1}(d-k) show how the nonlinear methods predict $Y$. Only the neural-network-based predictors, i.e., MLM, MOE, DNN, and the tree-based predictor RF can fit the non-smooth edges of the regression function well, while the other methods generate over-smoothed prediction functions.
}

{\revisionii
\subsection{MLM for Real-world Problems}
}

{\revisionii 
For the real-world problems, we use five datasets described below.} For all the datasets, we use dummy encoding for nominal variables. That is, we generate $c-1$ binary vectors for a nominal variable with $c$-classes. The numbers of samples and features in all the datasets are shown in Table \ref{tab:description}.

\begin{table}[htp]
\centering
\begin{tabular}{l|c|c|c|c|c}
	\Xhline{2\arrayrulewidth}
	\hline\hline
	Data & KIRC & SKCM & PD & \shortstack{Bike\\Sharing} & \shortstack{Cal\\Housing} \\
	\hline
	Number of samples & 430 & 388 & 756 & 17379 & 20640 \\
	Number of original features & 24 & 30 & 753 & 12 & 8 \\
	\hspace{1em} Continuous / Ordinal & 20 & 25 & 753 & 10 & 8\\	
	\hspace{1em} Nominal & 4 & 5 & 0 & 2 & 0\\
	\hspace{1em} After dummy encoding & 45 & 73 & 753 & 16 & 8\\
	\hline\hline
	\Xhline{2\arrayrulewidth}
\end{tabular}
	\caption{Data descriptions.}
	\label{tab:description}
\end{table}

\begin{itemize}
\item \textit{KIRC} and \textit{SKCM} are respectively the Cancer Genome Atlas Kidney Renal Clear Cell Carcinoma (TCGA-KIRC)  \citep{akin2016radiology} and {The Cancer Genome Atlas Skin Cutaneous Melanoma (TCGA-SKCM)} clinical data. Both are part of a large collection made for studying the connection between cancer phenotypes and genotypes. The original datasets include tissue images, clinical, biomedical, and genomics data. We only use clinical data in this experiment. For both datasets, the target variable is overall survival (OS) status coded in binary ($1$ indicating living and $0$ deceased). The covariates consist of demographic and clinical variables ($24$ for KIRC and $30$ for SKCM) such as age, gender, race, tumor status, dimension of specimen, and so on. After dummy encoding of categorical variables, KIRC has $45$ features and SKCM has $73$ features. Both datasets are available in R package \texttt{TCGAretriever}.

\item \textit{Bike Sharing} \citep{fanaee2014event} is an hourly time series data for bike rentals in the Capital Bikeshare system between years $2011$ and $2012$. The count of total rental bikes is the target variable, and $12$ covariates consist of Year$\,\in\{2011,2012\}$, Month$\,\in\{1,\cdots,12\}$, Hour$\,\in\{0,\cdots,23\}$, Holiday$\,\in\{0,1\}$, Weekday$\,\in\{0,\cdots,6\}$, Working day$\,\in\{0,1\}$, Temperature$\,\in(0,1)$ that is normalized from the original range of $(-8,39)$, Feeling temperature$\,\in(0,1)$ that is normalized from $(-16,50)$, Humidity$\,\in(0,1)$, Wind speed$\,\in(0,1)$, Season$\,\in\{0$(winter), $1$(spring), $2$(summer), $3$(fall)$\}$, Weather$\,\in\{0$(clear), $1$(cloudy), $2$(light rain or snow), $3$(heavy rain or snow)$\}$. After dummy encoding of Season and Weather, we get $16$ covariates.

\item \textit{Cal Housing} \citep{pace1997sparse} is a dataset for the median house values in California districts, which are published by the $1990$ U.S. Census. In this dataset, a geographical block group is an instance/unit. The original covariates consist of $8$ features including latitude and longitude which are spatial variables, median income, median house age (house age), average number of rooms, average number of bedrooms (bedrooms), block population (population), average house occupancy (occupancy), and latitude and longitude.  Given the average number of rooms and the average number of bedrooms, we compute the average number of non-bedroom rooms as a new covariate to replace the former. 

\item \textit{Parkinson's Disease (PD)} dataset \citep{sakar2019comparative} is collected from 188 patients with PD and 64 healthy individuals to study PD detection from the vocal impairments in sustained vowel phonations of patients. The target variable is  binary, 1 being PD patients and 0 otherwise. The covariates consist of various clinical information generated by processing the sustained phonation of the vowel 'a' from each subject. A total of 753 features are generated by various speech signal processing algorithms, {\it e.g.}, Time Frequency Features, Mel Frequency Cepstral Coefficients (MFCCs).
\end{itemize}

\begin{table}[htp]
	\centering
	\begin{tabular}{ll|cc|cc|cc|cc|cc}
	\Xhline{2\arrayrulewidth}
	\hline\hline
	\multicolumn{2}{r|}{Data} &
	\multicolumn{2}{c|}{KIRC} & \multicolumn{2}{c|}{SKCM} & \multicolumn{2}{c|}{PD} & \multicolumn{2}{c|}{Bike Sharing} & \multicolumn{2}{c}{Cal Housing}   \\
	\multicolumn{2}{r|}{} & \multicolumn{2}{c|}{(AUC)} & \multicolumn{2}{c|}{(AUC)} & \multicolumn{2}{c|}{(AUC)} & \multicolumn{2}{c|}{(RMSE)} & \multicolumn{2}{c}{(RMSE)} \\ 	
	\multicolumn{2}{l|}{Model} & Train & Test & Train & Test & Train & Test & Train & Test & Train & Test\\ 
	\hline
	\hline
	& {\revision NCM} & .834& .723& .774& .650& .794& .750& 122.1 & 127.7 & .795 & .815 \\
	& {\revision MOE} & {\revision.888}& {\revision.873}& {\revision.973}& {\revision.756}& {\revision.891}& {\revision.815}& {\revision 47.6}& {\revision 52.8}& {\revision.684}& {\revision.687} \\
	& {\revision CWM} & {\revision -}& {\revision -}& {\revision -}& {\revision -}& {\revision -}& {\revision -}& {\revision 38.5}& {\revision 109.2}& {\revision.369}& {\revision.623} \\
	
	& RF 	& .983& .869& .997& .688& .997& .794& 43.8 & 52.2 & .421 & .554 	\\
	& SVM 	& .843& .856& .764& .667& .756& .728& 145.6 & 148.1	& .671 & .676  \\
	& MARS & -&-& -&-& -&-& 141.1 & 140.1 & .629& .640 \\
	& MLP 	& .974& \textbf{.907}& .987& \textbf{.777}& .975& \textbf{.833}& 40.1 	& \textbf{47.6}& .503 & \textbf{.517}\\
	\hline
	& LR 	& .853& .888& .826& .702& .880& .789& 141.1 & 140.3	& .723 & .727 \\
	& SAR & -&-& -&-& -&-& -&-& .655& .652\\
	& GA${}^1$M & .838& .847& -&-& -&-& 99.4& 101.1 & .613 & .640\\
	& MLM-cell 	& .904& \textbf{.891}& .948& .728& 1.000& \textbf{.860}& 52.7 & \textbf{60.9} & .560 & \textbf{.570}\\
	& MLM-EPIC & .902& \textbf{.891}& .861& \textbf{.742}& .975& .851& 62.8 & 66.7 & .569 & .584\\
	\hline\hline
	\Xhline{2\arrayrulewidth}
		
	\end{tabular}
	\caption{Prediction accuracy of regression and classification methods. We partition these methods into two categories: complex (top panel) and interpretable models (bottom panel). For each method, the model parameters are fine tuned. If a method is not applicable to a dataset, the corresponding entry in the table is not listed. {\revision The best test accuracy achieved for each dataset and by either category of methods is highlighted in bold font. Higher values of AUC or lower values of RMSE correspond to better performance.} }
	\label{tab:real}
\end{table}

\subsubsection{TCGA Clinical Data}
\label{sec:tcga}

For both KIRC and SKCM data, MLP is constructed using $2$ hidden layers that have respectively $128$ and $32$ units for KIRC and $256$ and $32$ units for SKCM, and trained for $10$ epochs. Random forest is trained with maximum depth equal to $10$. For both MLM-cell and MLM-EPIC, hyper-parameters $K_l$ is set based on $10$-fold cross validation (CV) within training data, and assumed to be equal across the $2$ hidden layers. We choose another hyper-parameter $\widetilde{J}$ by comparing the training accuracy obtained at several values. 
For KIRC, the number of layer $l$-cells is set to $K_l=4$ according to CV for $l=1,2$, and $12$ cells are formed at the end. These $12$ cells are merged into $\widetilde{J} = 10$ EPICs. For SKCM, the number of layer $l$-cells is set to $K_l = 13$ according to CV for $l=1,2$, and $77$ cells are formed at the end. The cells are merged into $\widetilde{J}=10$ EPICs.

The first two columns of Table \ref{tab:real}  show the prediction accuracy for KIRC and SKCM. The area under curve (AUC) score is used as the efficacy metric. For both datasets, MLP achieves the best prediction for the overall survival (OS) status. RF has a higher training AUC, but not for the test data. Among the interpretable models, GAM does not converge for SKCM. LR for KIRC and SKCM, and GAM for KIRC have higher test AUC scores compared to other complex models although their training AUC scores are not better than RF. On the other hand, MLM-cell and MLM-EPIC achieve relatively high AUC scores in both training and testing accuracy, and outperform LR.

\begin{figure}[h]
	\centering
    \begin{subfigure}[b]{0.41\textwidth}
         \centering
         \includegraphics[width=\textwidth]{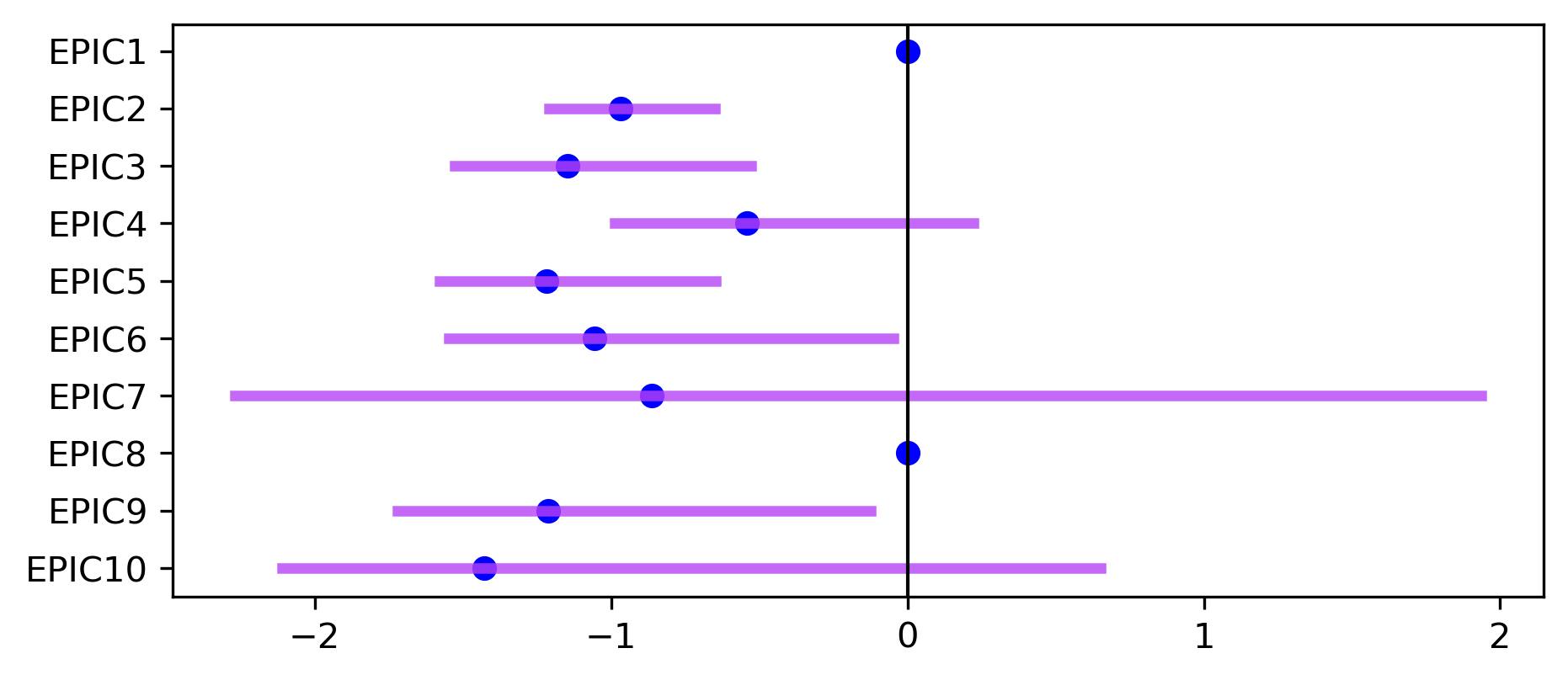}
         \caption{Age}
    \end{subfigure}
	\begin{subfigure}[b]{0.41\textwidth}
         \centering
         \includegraphics[width=\textwidth]{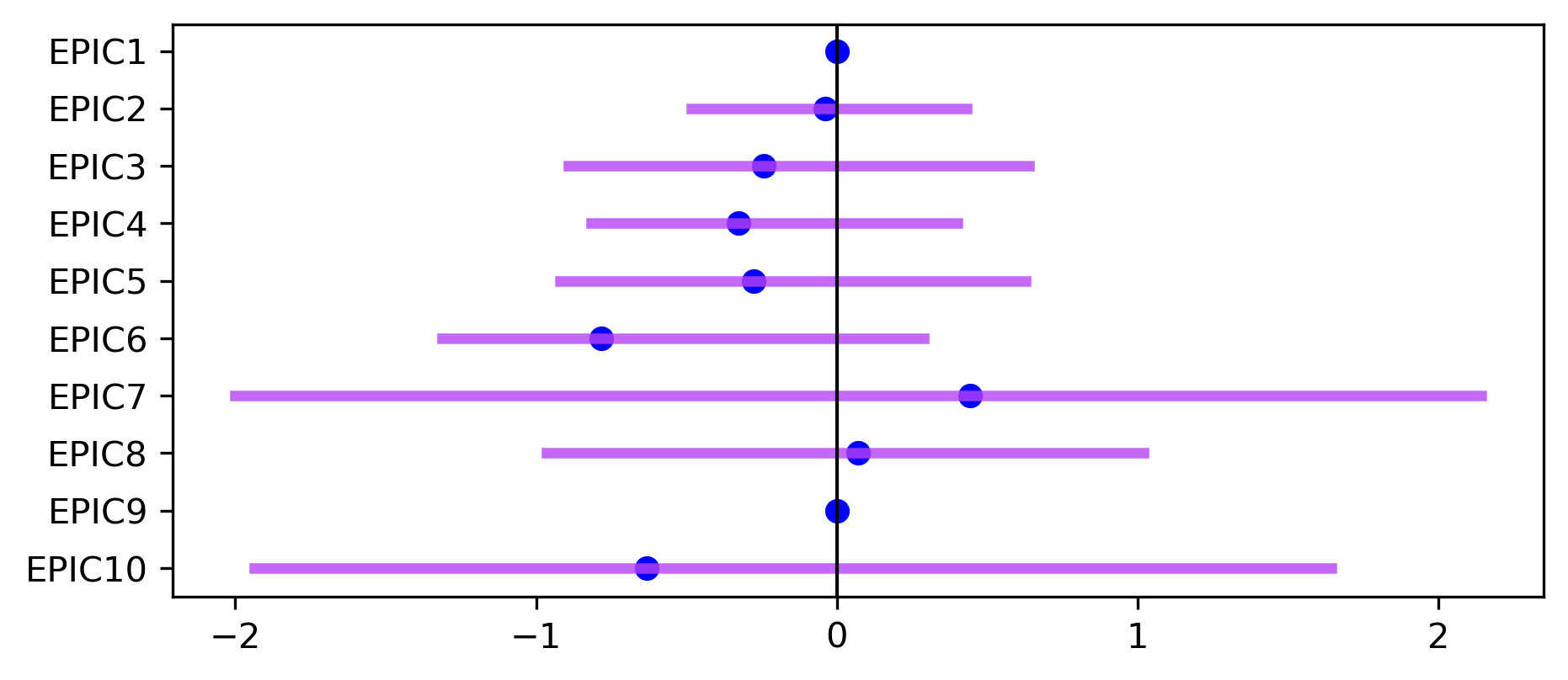}                  
         \caption{Sex}
    \end{subfigure}\\
    \begin{subfigure}[b]{0.41\textwidth}
         \centering
         \includegraphics[width=\textwidth]{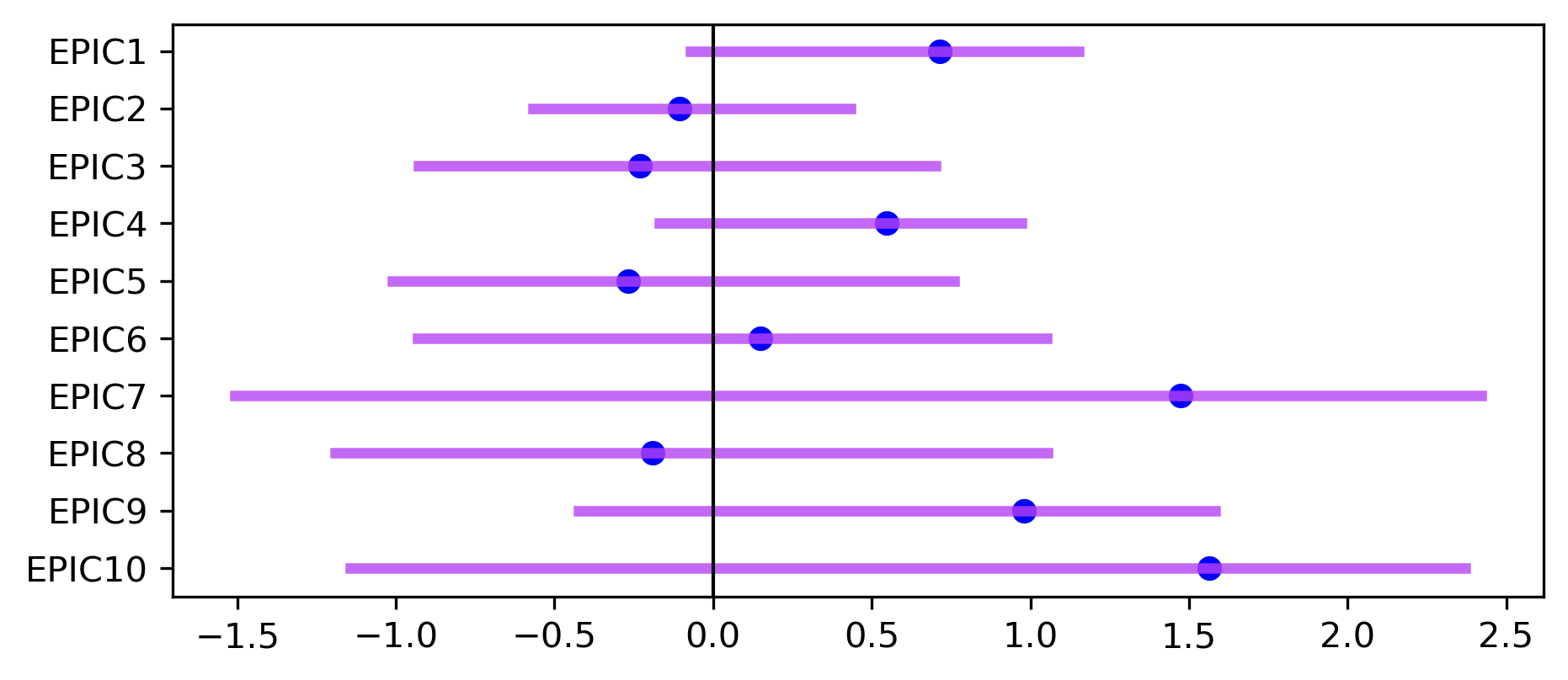}
         \caption{Radiation treatment adjuvant}
    \end{subfigure}
	\begin{subfigure}[b]{0.41\textwidth}
         \centering
         \includegraphics[width=\textwidth]{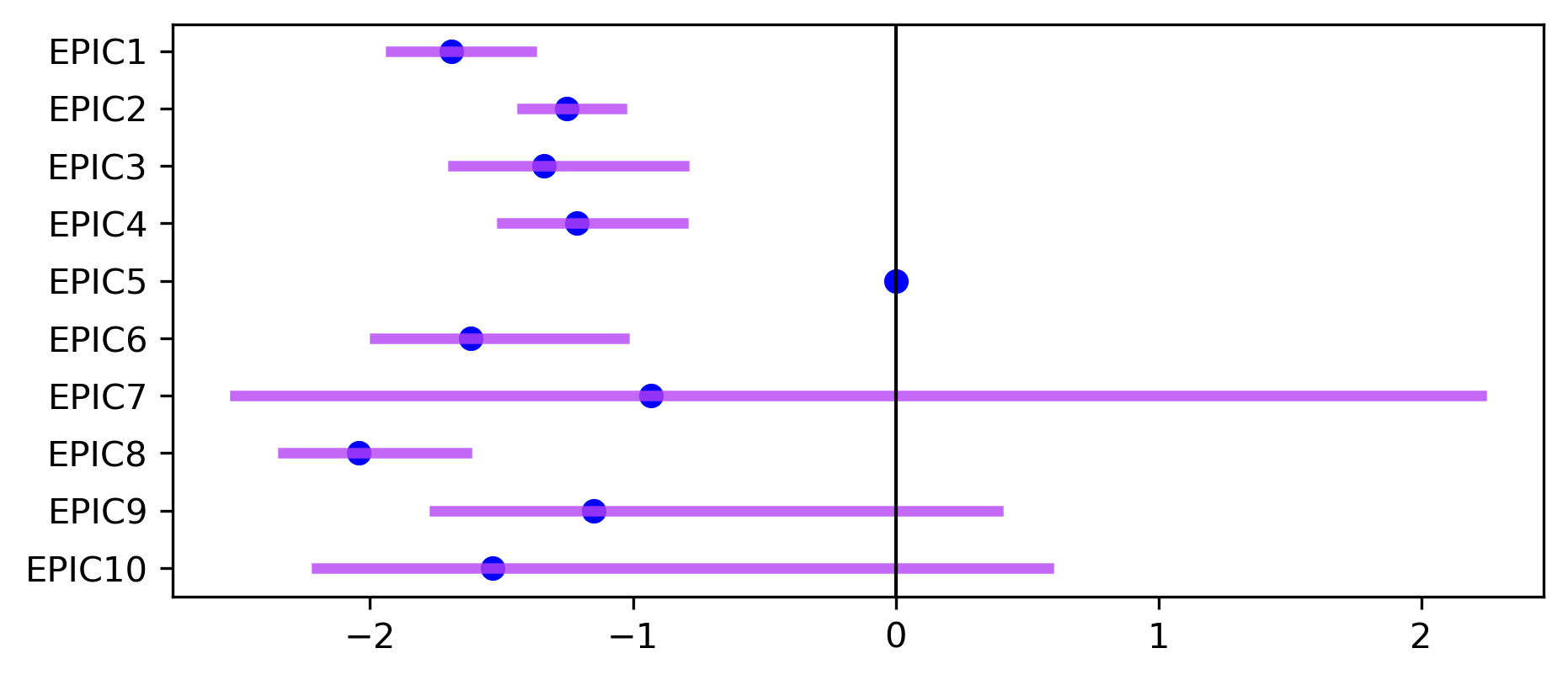}     
         \caption{History other malignancy}
    \end{subfigure}
	\caption{Estimated regression coefficients (blue dots) with {\revisionii "naive"} confidence intervals (purple lines) for four selected variables in SKCM. The X-axis indicates the value of each regression coefficient. The linear effects of age, sex, whether the patient has radiation treatment adjuvant, and whether the patient has the history of other malignancy, vary by EPICs.}
	\label{fig:skcm}
\end{figure}

We interpret MLM-EPIC via local linear models for each EPIC. Figure \ref{fig:skcm} shows the estimated regression coefficients and the associated {\revisionii``naive''} confidence intervals for four selected variables of SKCM. The EPICs are sorted in descending order according to their sizes. {\revisionii Note that the ``naive'' confidence intervals here are not  confidence intervals in a rigorous sense. 
They are computed under the assumption that the model structure of MLM is correct, which involves data-driven model selection. We provide these intervals as nonstandard quantities to help us understand MLM. 
A more rigorous study on valid confidence intervals under the MLM setting is an interesting future work. The work of~\cite{leeb2015various} provides useful insights into similar problems.}

Figure \ref{fig:skcm}(a) demonstrates that age is {\revisionii not associated} with the OS status in EPICs {1 and 8}, as the coefficients related to age are shrunk to zero by LASSO penalty. However, the irrelevance of age does not hold in other EPICs. Such EPIC-specific results enable us to gain understanding impossible with a single LR model or a DNN. In addition, EPICs $2$, $3$, $5$, $6$, and $9$ have negative coefficients for age. 
This result may motivate researchers to {\revisionii hypothesize whether the impact of age differs between different groups of patients.}
Similar analyses can be made on the other three variables. Moreover, Figure \ref{fig:skcm}(d) demonstrates that the prediction model for subjects belonging to EPIC $5$ does not include the variable, ``history of other malignancy'', suggesting a heterogeneous effect of this covariate on the OS status.

Now, we investigate how EPICs are formed. We compute explainable conditions for each EPIC to check if they provide any insight for distinguishing the EPICs. Here we interpret the largest EPIC that consists $59$ samples. $\psi$ and $\eta$ are set to 1 and 4.  Table \ref{tab:skcm_desc} shows explainable conditions for the EPIC 1. More explainable conditions for other EPICs are provided in Supplementary Material.
Specifically, around half of patients in EPIC 1 are relatively young patients who have new tumor events after the initial treatment and have less than $306$ survival months. Combining this information with the information from Figure \ref{fig:skcm}, an interesting hypothesis can be that for the patients in EPIC 1 under these conditions, age does not impact the OS status. 
Such insight is not available from MLP or other complex models which are hard to interpret.

\begin{table}[htp]
\centering
\begin{tabular}{c|p{12cm}|c}
\Xhline{2\arrayrulewidth}
	\hline\hline
	EPIC & Descriptions & Size\\
	\hline
	1& \footnotesize Age$\,<29.0$, Days to collection$\,\leq10345$, Initial pathology DX year$\,>1996$, New tumor event after initial treatment$\,=1$, Overall survival months$\,\leq 306.2$, Retrospective collection$\,=1$, AJCC staging edition$\,\neq 5$ & 11\\
	\cline{2-3}
	(59) & \footnotesize Age$\,<29.0$, $8701<\,$Days to collection$\,>10345$, Initial pathology DX year$\,>1996$, New tumor event after initial treatment$\,=1$, Overall survival months$\,\leq 306.2$, Retrospective collection$\,=1$, Submitted tumor DX days$\,>9737$, AJCC staging edition$\,\neq 5$ & 8\\
	\cline{2-3}
	& \footnotesize Age$\,<29.0$, Days to collection$\,>10345$, ICD-10$\,<5$, ICD-O-3 site$\,>23.5$, Overall survival months$\,\leq 306.2$, Retrospective collection$\,=1$, AJCC staging edition$\,\neq 5$ & 7\\
	\Xhline{2\arrayrulewidth}
\end{tabular}
	\caption{Explainable conditions for the largest EPIC for SKCM data. Numbers within the bracket indicate the size of the EPIC in training data.}
	\label{tab:skcm_desc}
\end{table}

\subsubsection{Bike Sharing Data}
\label{sec:bike}
The bike sharing dataset is often used to demonstrate the efficacy of complex machine learning models. For this dataset, RF is set with its maximum depth equal to $10$. MLP is constructed with $3$ layers with $30$ units per layer, trained by $20$ epochs. MLM-cell is computed using MLP as its co-supervising neural network. The number of cells per layer is $100$, which is chosen by CV. Finally, $1712$ cells are formed. MLM-EPIC is constructed by merging $1712$ cells into $150$ EPICs. For the estimation of the MLM-EPIC parameters, we apply pooled covariance to estimate GMM of each EPIC. 
{\revision With pooled covariance, we essentially assume that the Gaussian components in every EPIC share the same volume and shape in their covariance matrices.} {\revisionii This restriction has been stated as a possibility to regularize a mixture model~\citep{fraley1998mclust}.} In our software package, pooled covariance is provided as an option. Cross-validation on the training samples can be used to decide whether to use this option.
For other methods, we apply the default settings of the methods from \textbf{pyearth} (for MARS), \textbf{pygam} (for GAM), and \textbf{scikit-learn}  python packages.

\begin{figure}[tp]
	\centering
    \begin{subfigure}[b]{0.24\textwidth}
         \centering
         \includegraphics[width=\textwidth]{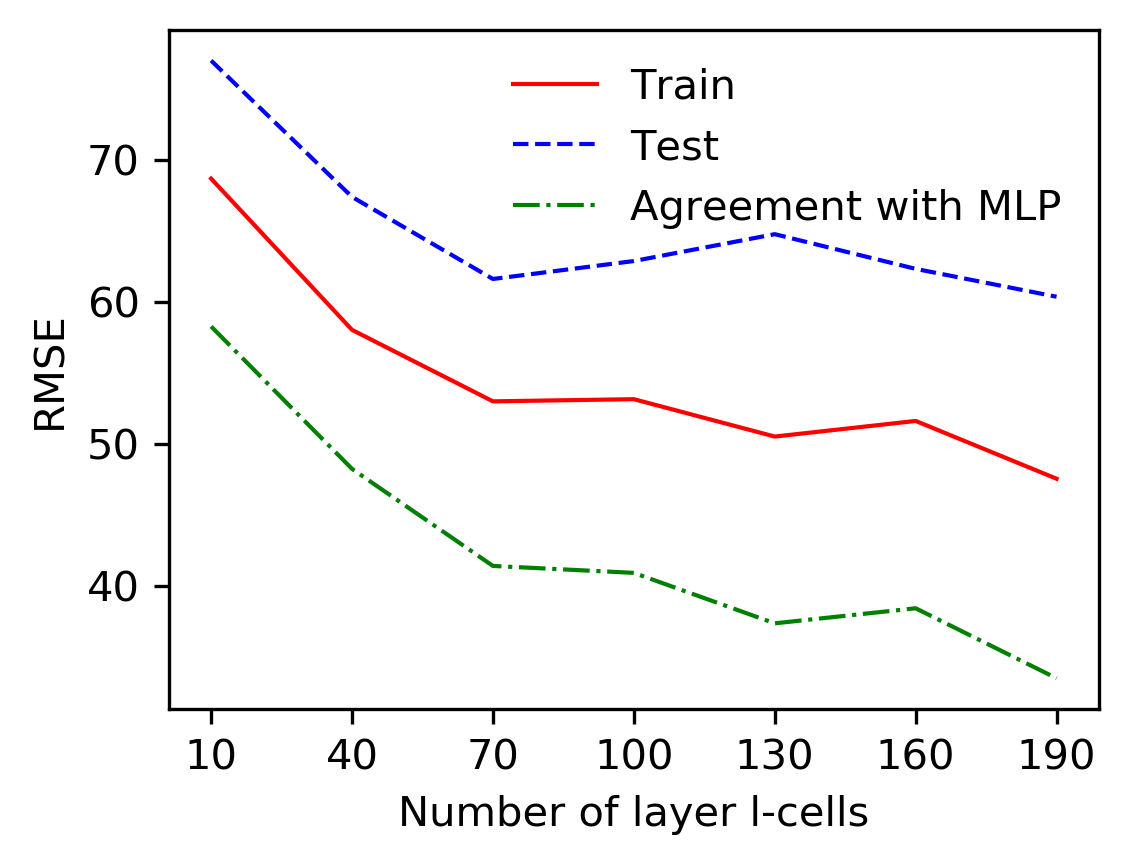}\\
         (a)
    \end{subfigure}
	\begin{subfigure}[b]{0.24\textwidth}
         \centering
         \includegraphics[width=\textwidth]{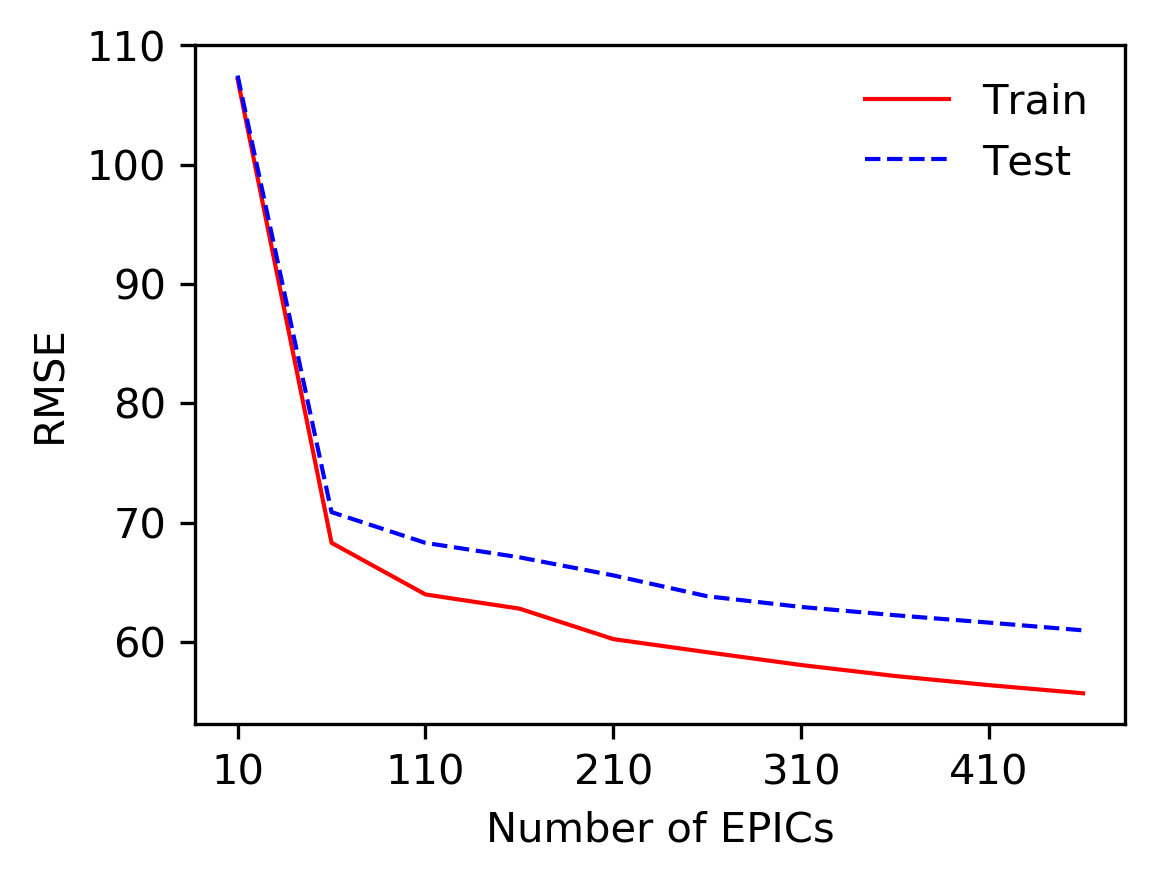}\\ (b)
    \end{subfigure}
    \begin{subfigure}[b]{0.24\textwidth}
         \centering
         \includegraphics[width=\textwidth]{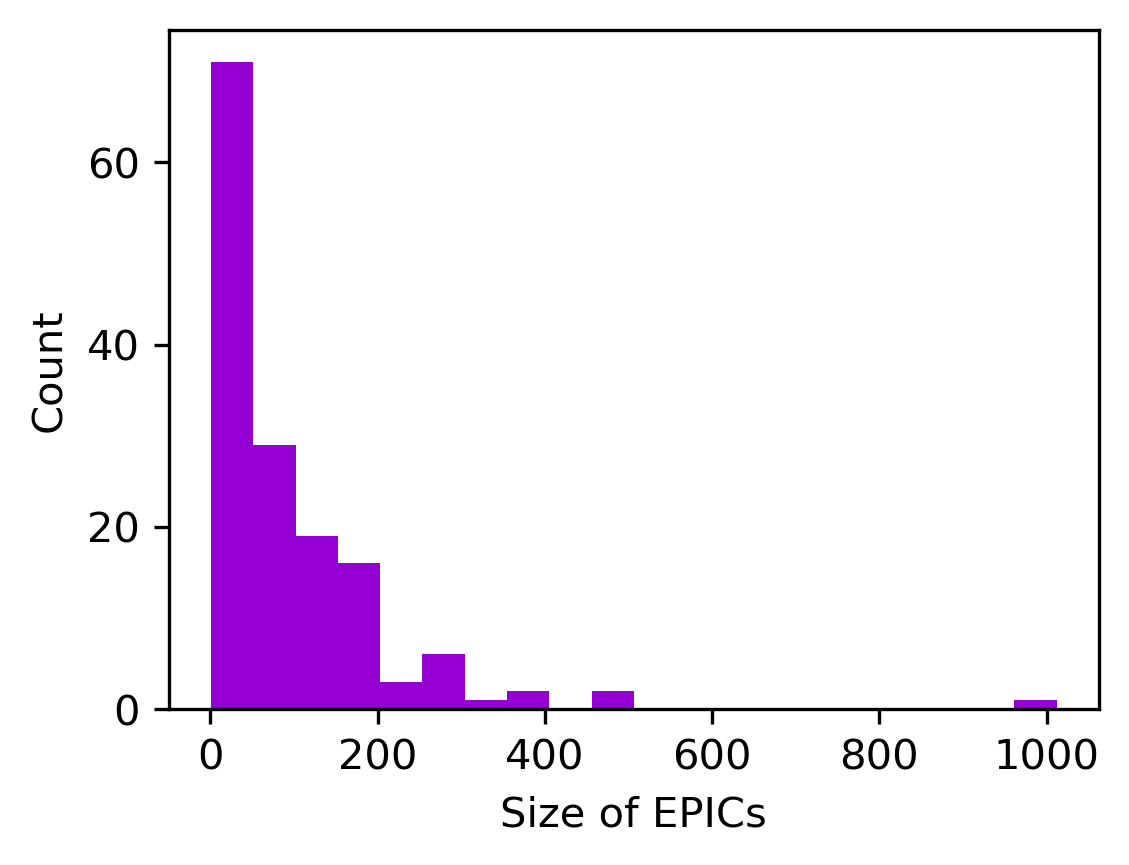}\\
         (c)
    \end{subfigure}
    \begin{subfigure}[b]{0.24\textwidth}
         \centering
         \includegraphics[width=\textwidth]{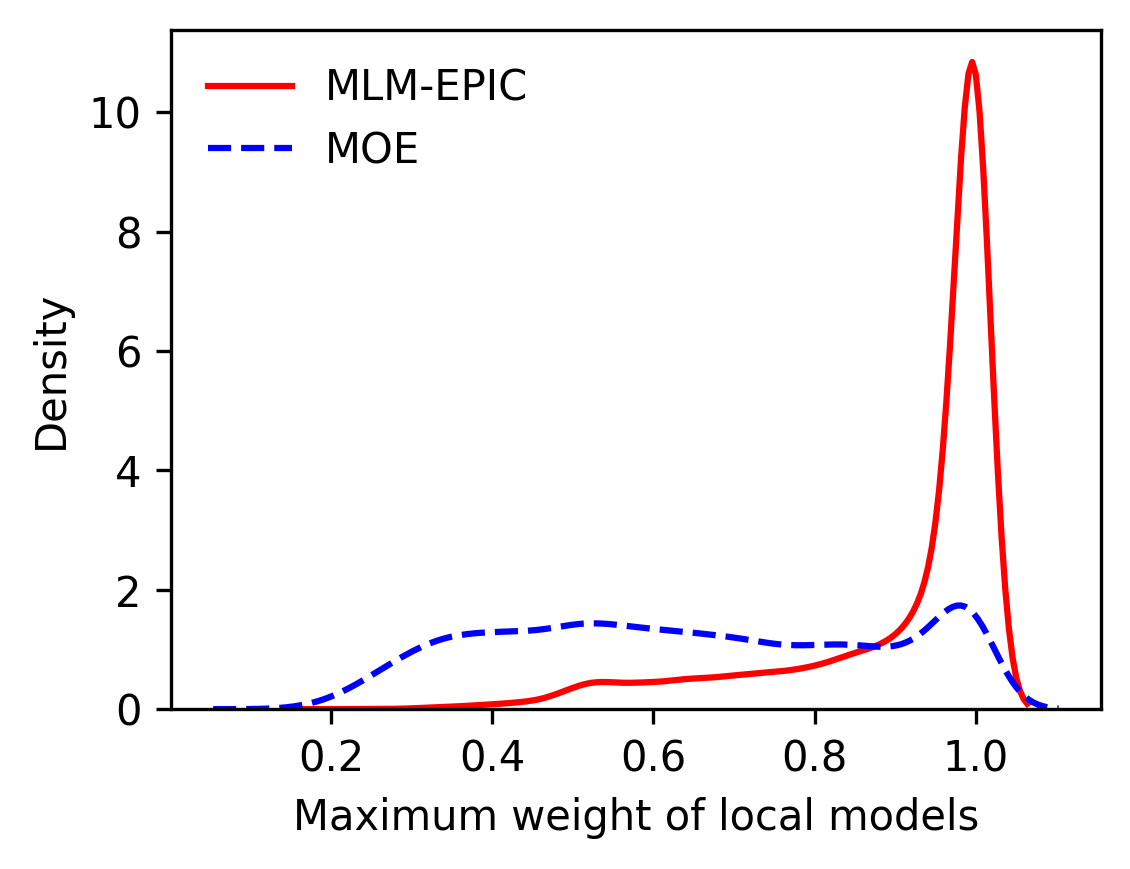}\\
         (d)
    \end{subfigure}
	\caption{Impact of hyper-parameters, $K_l$ and $\widetilde{J}$, on MLM for bike sharing data. (a) RMSE of MLM-cell at different $K_l$'s (the number of cells at layer $l$). (b) RMSE of MLM-EPIC at different $\widetilde{J}$ (the number of EPICs). 
	(c) Histogram of the sizes of the $150$ EPICs. {\revision (d) Kernel density estimate (KDE) plot of the maximum weights assigned to an MLM-EPIC or MOE local expert model, computed at all the training points.}}
	\label{fig:bike}
\end{figure}

First, we examine the prediction accuracy of the methods. Table \ref{tab:real} shows that RF and MLP achieve higher prediction accuracy in comparison with other simpler methods. The RMSEs of MLM-cell and MLM-EPIC are considerably lower than any other method except for RF and MLP. Figure \ref{fig:bike} (a) and (b) show the trade-off between the prediction accuracy and the model complexity of MLM (indicated by $K_l$ for MLM-cell in (a) and by $\widetilde{J}$ for MLM-EPIC in (b)).
As the model complexity increases, both training and testing RMSE decreases. We do not observe overfitting as the model complexity of MLM is capped by the underlying neural network.
The measure for agreement between MLP and MLM-cell is computed by the RMSE between the predicted values of MLP and MLM-cell respectively. As shown in the figure, the agreement with MLP decreases as the model complexity increases.
Figure \ref{fig:bike} (c) is the histogram of EPIC sizes obtained from the $150$ EPICs of the training data. The histogram shows that many data points belong to small EPICs. For the sample points in small EPICs, the predicted values based on MLM-EPIC are similar to using the nearest neighbor method. For example, in the extreme case, if all EPICs consist of a single point, MLM-EPIC is equivalent to using $1$-nearest neighbor method.

{\revision Among the other methods, MOE achieves higher accuracy than MLM-cell and MLM-EPIC. However, although MOE exploits local models, the prediction at any point is rarely dominated by a single local model. Specifically, the prediction of MOE is a weighted sum over the predicted values given by its local experts. The weights assigned to the local experts vary with the input points. If these weights flatly spread over the local experts, interpretation based on MOE is still difficult. This scenario is what we have observed with this dataset. For each training sample point, we compute the maximum weight assigned to the local experts of MOE and find that the average of these maximum weights (across the sample points) is $0.65$ and at half of the sample points, the maximum weight is below $0.31$. In contrast, when MLM-EPIC is used, the average maximum weight assigned to a local model is $0.91$, and $73\%$ of the sample points have a maximum weight above $0.9$. In Figure \ref{fig:bike} (d), we show the kernel density estimate (KDE) for the maximum weights across the training data for MOE and MLM-EPIC respectively. For MLM-EPIC, the KDE has a high peak close to $1.0$, while for MOE, the KDE is flat across a wide range.
}

\begin{figure}[h]
\begin{center}
	\includegraphics[width=0.9\textwidth]{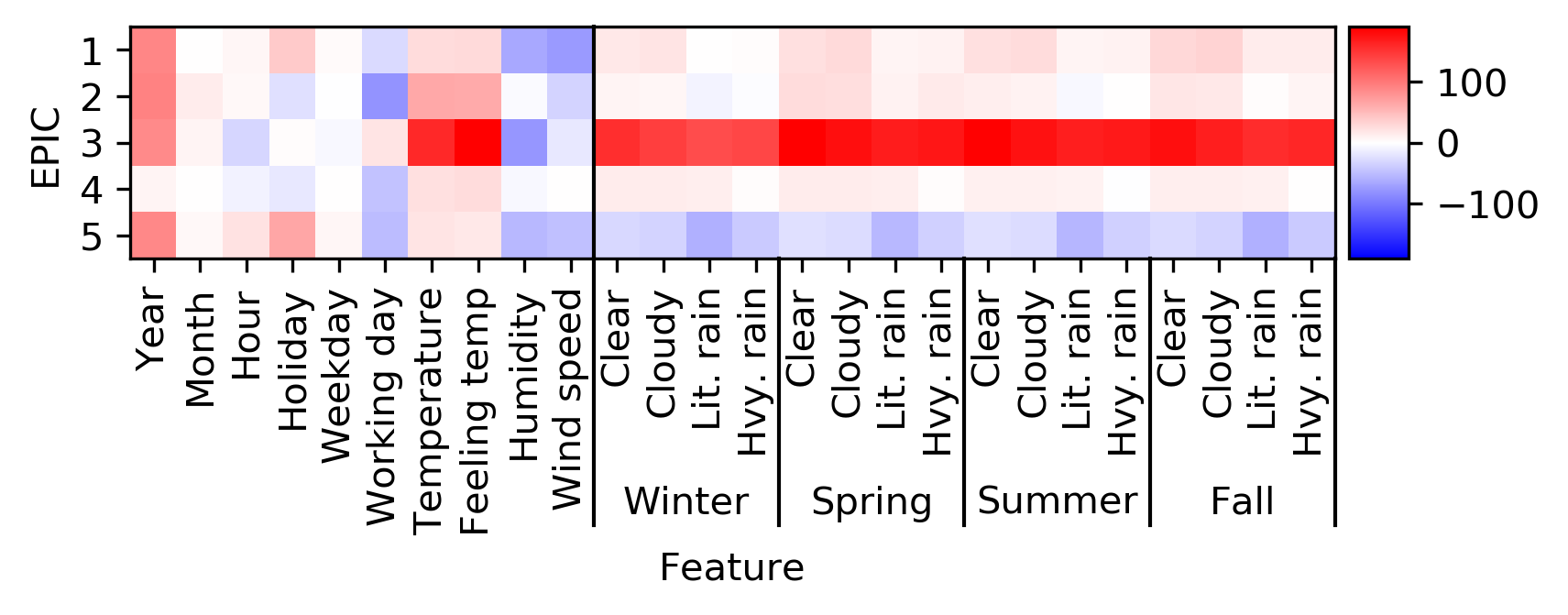}
	\caption{The linear mixture model regression coefficients $\{\mathbf{\hat{\beta}}_j|\mathbf{x}\in \mathcal{P}_j\}_{j=1}^{\widetilde{J}}$ for the top $5$ largest EPICs for bike sharing data.}
	\label{fig:bike_mosaic}
\end{center}
\end{figure}

Figure \ref{fig:bike_mosaic} shows the fitted regression coefficients of MLM-EPIC for the 5 largest EPICs. The values of the coefficients are indicated by color. Depending on which EPIC a sample point belongs to, the effect of a covariate is different when predicting the bike rental count. Specific interpretation of the fitted MLM-EPIC is provided in Supplementary Material.

\subsubsection{California Housing Prices}
\label{sec:calhousing}

California housing data was first introduced to demonstrate the efficacy of the spatial autoregressive (SAR) model \citep{pace1997sparse}, yet it is now often used to test neural network models. Previous works have shown that neural networks perform well with spatial data \citep{zhu2000mapping, ozesmi1999artificial}, however they cannot be interpreted. We hereby analyze this dataset with MLM for both prediction and interpretation.

RF is fitted with maximum depth $10$. MLP contains $3$ hidden layers with $30$ hidden units per layer, trained by $50$ epochs. 
SAR is fitted with the consideration of spatially lagged covariates except for the longitude and latitude. MLM-cell and MLM-EPICs are fitted based on the MLP model. MLM-cell is constructed with $6$ cells per layer, and $64$ final cells are formed (some combinations of cells over the $3$ layers are empty). MLM-EPIC merged the $64$ cells into $30$ EPICs. As shown by Table \ref{tab:real}, the prediction accuracy of MLM-cell and MLM-EPIC is slightly lower (i.e., slightly larger RMSE) than that of MLP or RF, but they achieve better accuracy than any other interpretable models. 

\begin{figure}[h]
	\centering
    \begin{subfigure}[b]{0.31\textwidth}
         \centering
         \includegraphics[width=\textwidth]{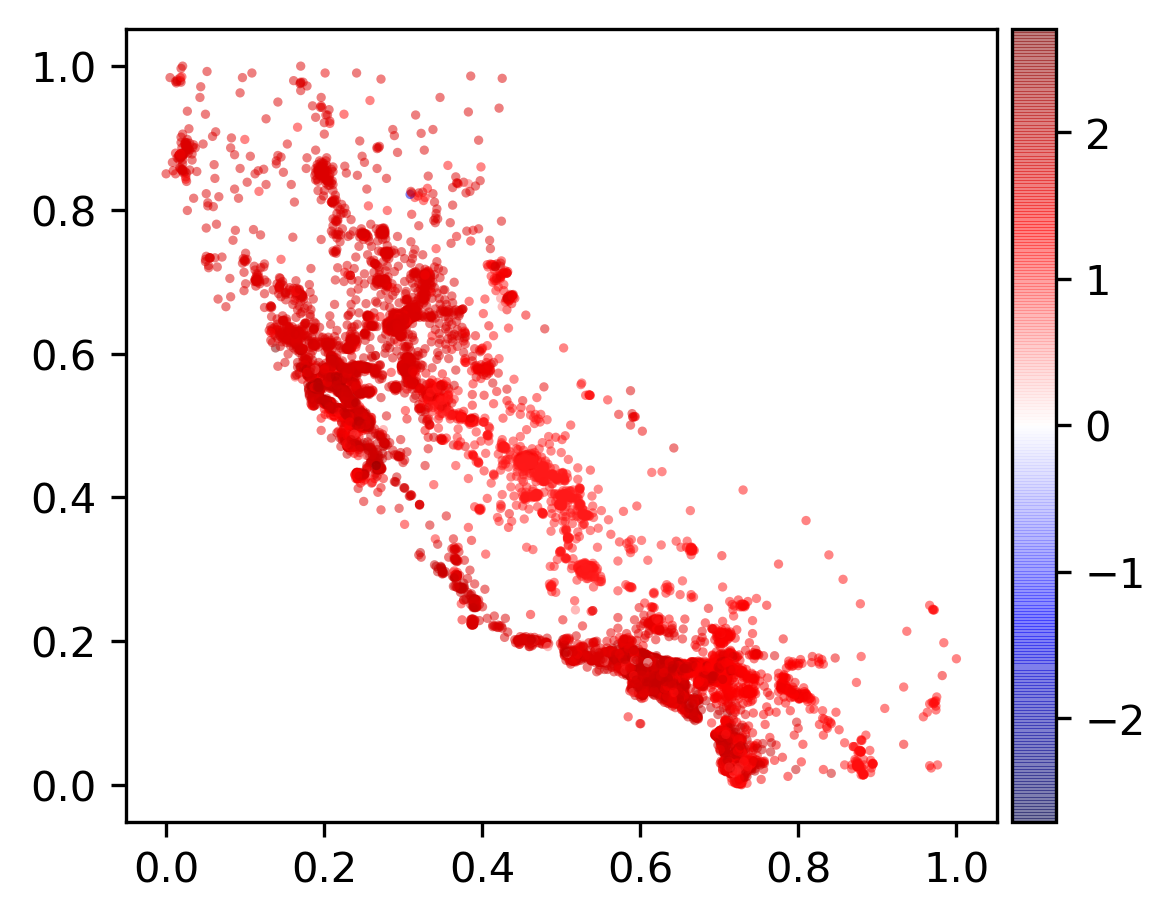}
         \caption{Median income}
    \end{subfigure}
	\begin{subfigure}[b]{0.31\textwidth}
         \centering
         \includegraphics[width=\textwidth]{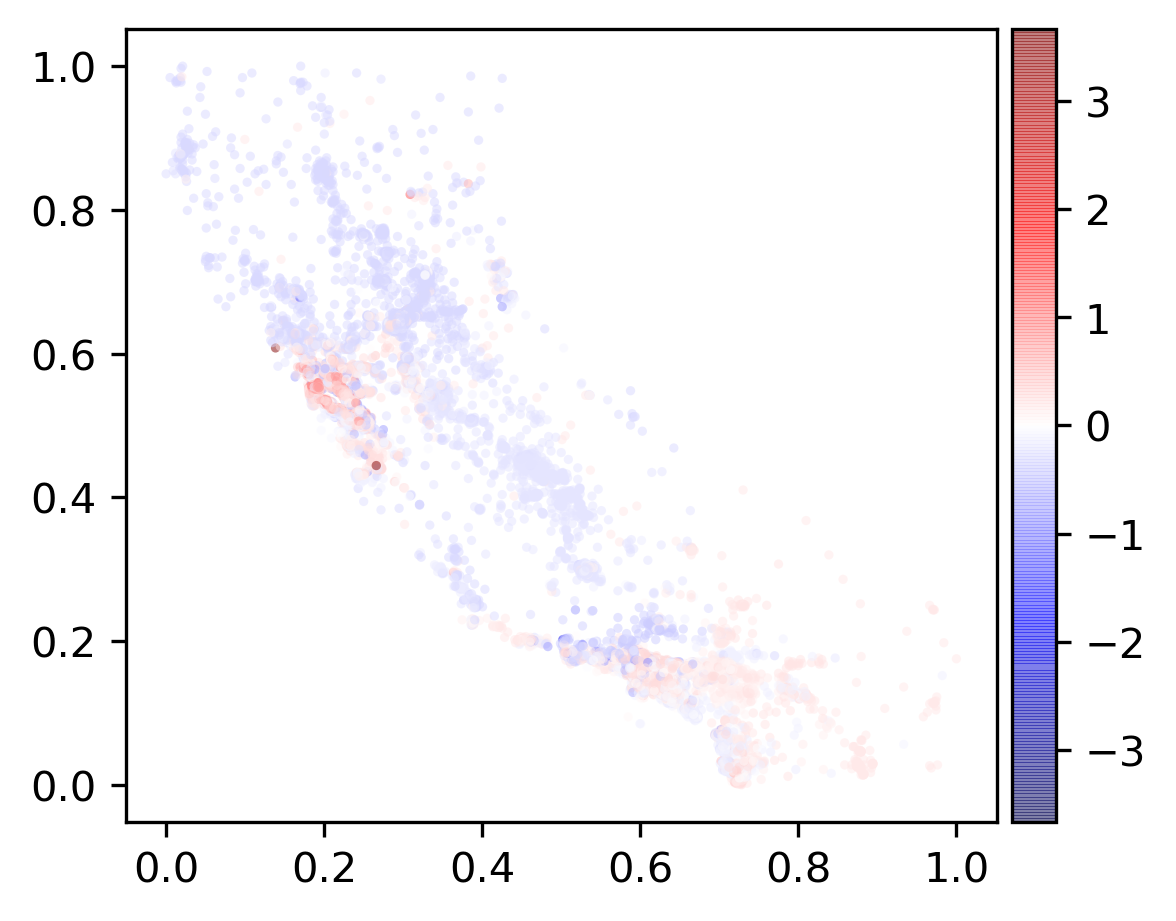}                  
         \caption{House age}
    \end{subfigure}
	\begin{subfigure}[b]{0.31\textwidth}
         \centering
         \includegraphics[width=\textwidth]{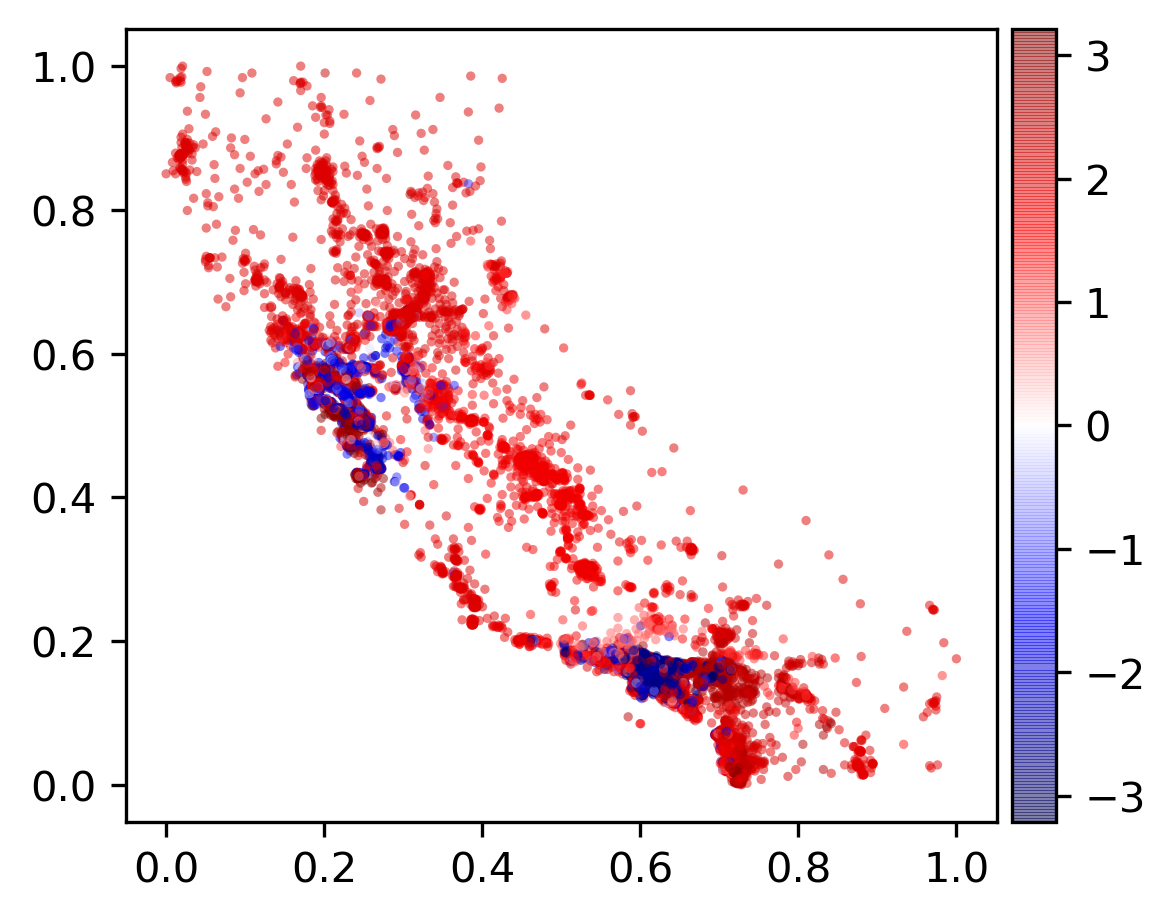}                  
         \caption{Other rooms}
    \end{subfigure}\\
	\begin{subfigure}[b]{0.31\textwidth}
         \centering
         \includegraphics[width=\textwidth]{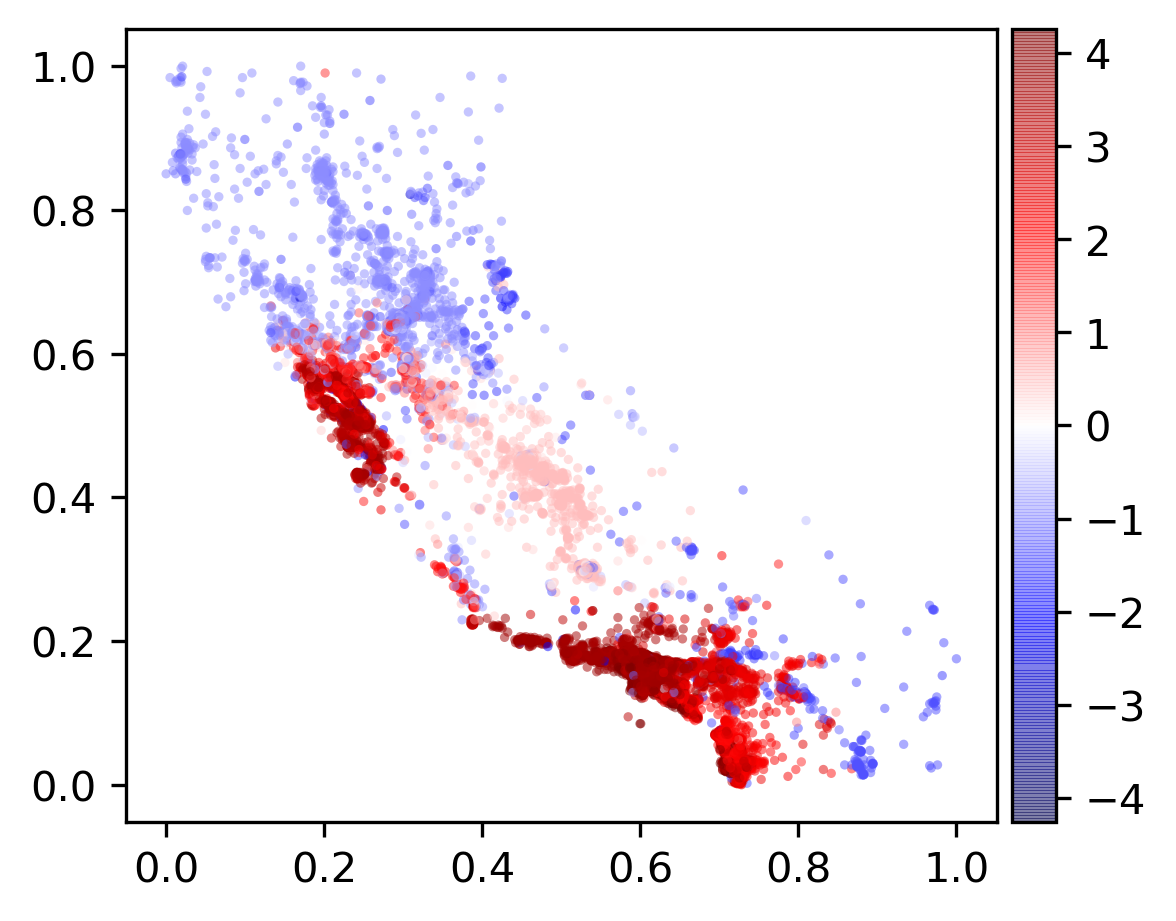}                  
         \caption{Bedrooms}
    \end{subfigure}  
    \begin{subfigure}[b]{0.31\textwidth}
         \centering
         \includegraphics[width=\textwidth]{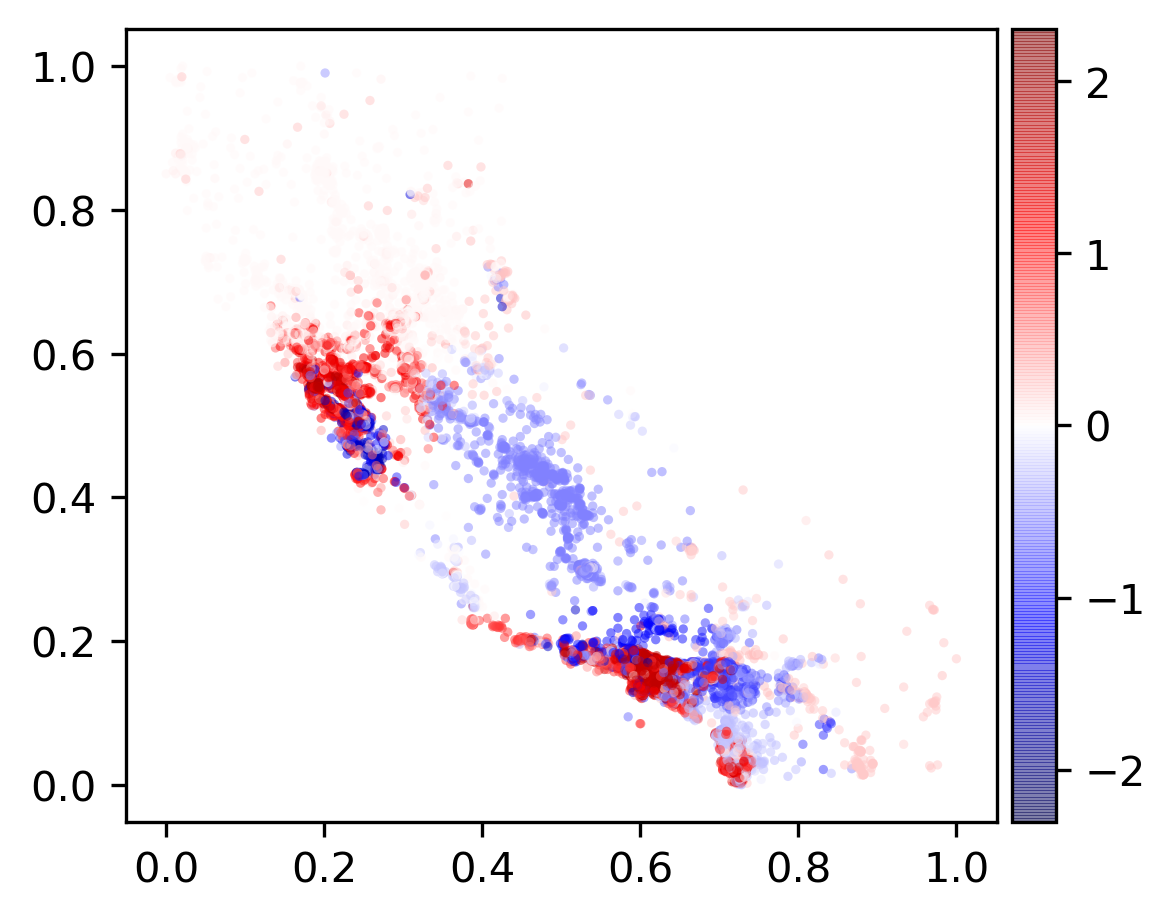}                  
         \caption{Population}
    \end{subfigure}
    \begin{subfigure}[b]{0.31\textwidth}
         \centering
         \includegraphics[width=\textwidth]{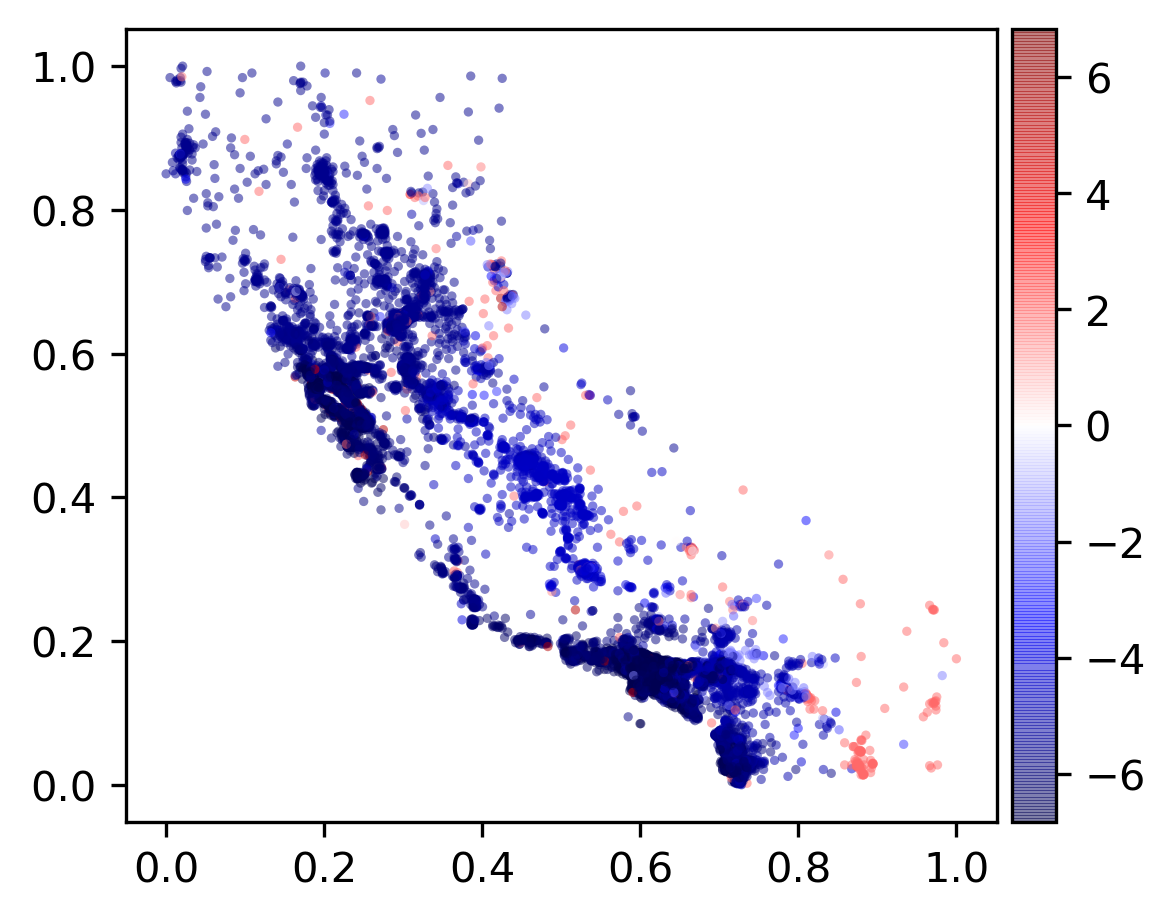}                  
         \caption{Occupancy}
    \end{subfigure}      
	\caption{Log-valued regression coefficients of local linear models plotted on longitude (horizontal axis) and latitude (vertical axis) space. {\revisionii The coefficients are transformed by $sign(x)\cdot log(|x|)$ for better visualization.}}
	\label{fig:cal_map}
\end{figure}
\begin{figure}[htp]
	\centering
    \begin{subfigure}[b]{0.31\textwidth}
         \centering
         \includegraphics[width=\textwidth]{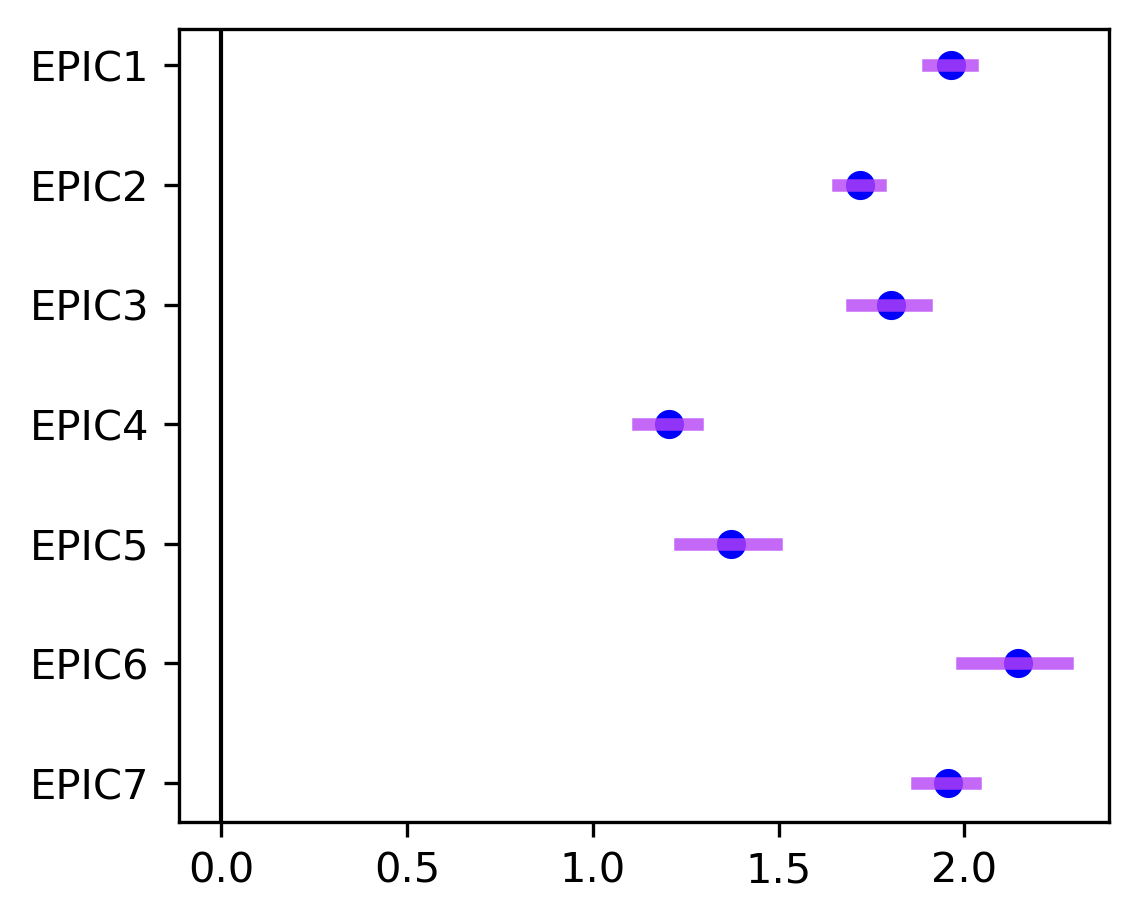}
         \caption{Median income}
    \end{subfigure}
	\begin{subfigure}[b]{0.31\textwidth}
         \centering
         \includegraphics[width=\textwidth]{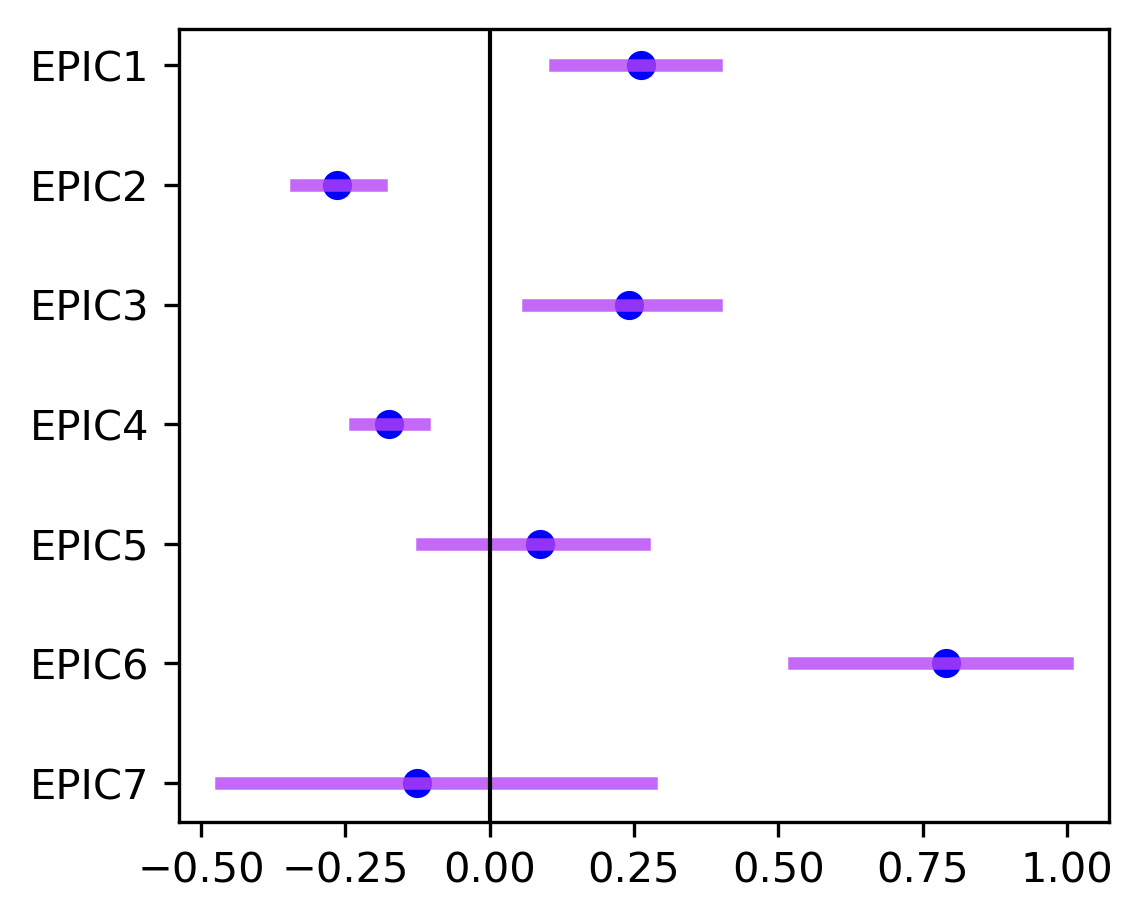}                  
         \caption{House age}
    \end{subfigure}
	\begin{subfigure}[b]{0.31\textwidth}
         \centering
         \includegraphics[width=\textwidth]{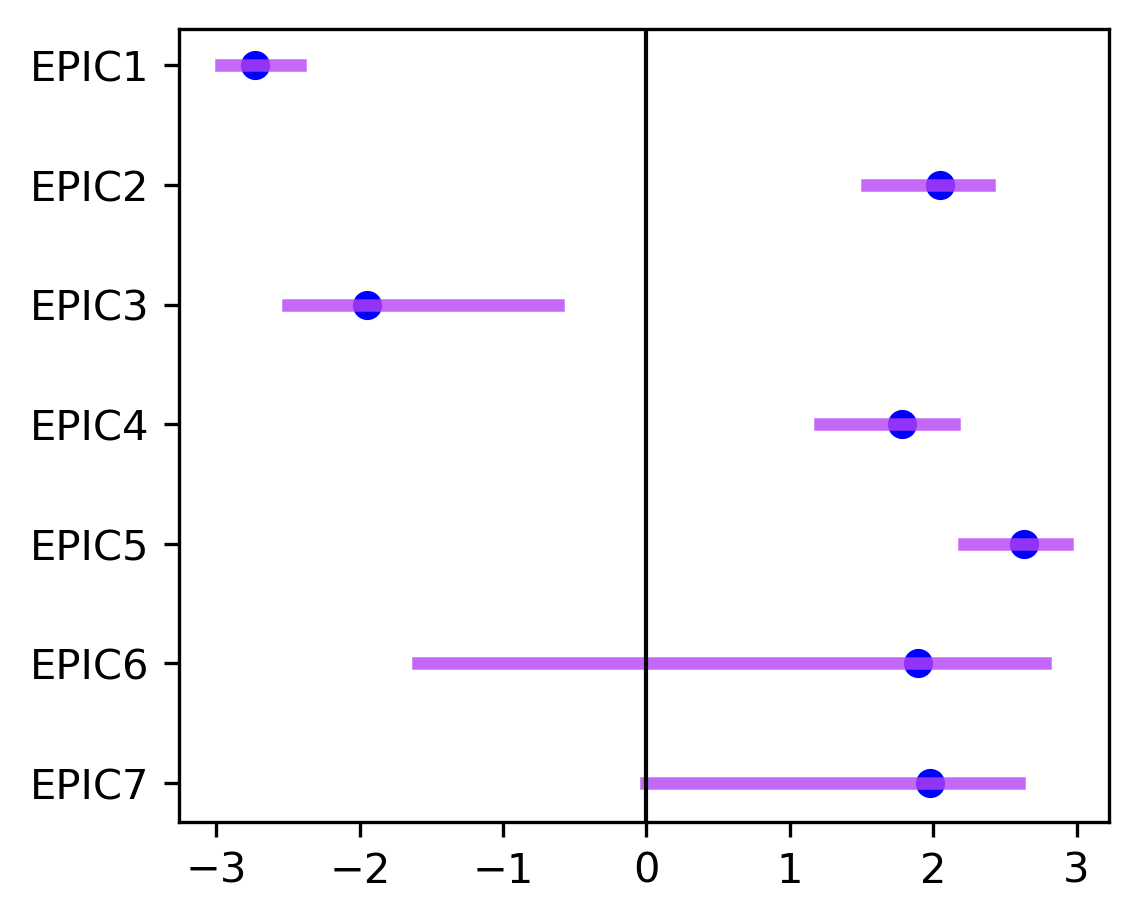}                  
         \caption{Other rooms}
    \end{subfigure}\\
	\begin{subfigure}[b]{0.31\textwidth}
         \centering
         \includegraphics[width=\textwidth]{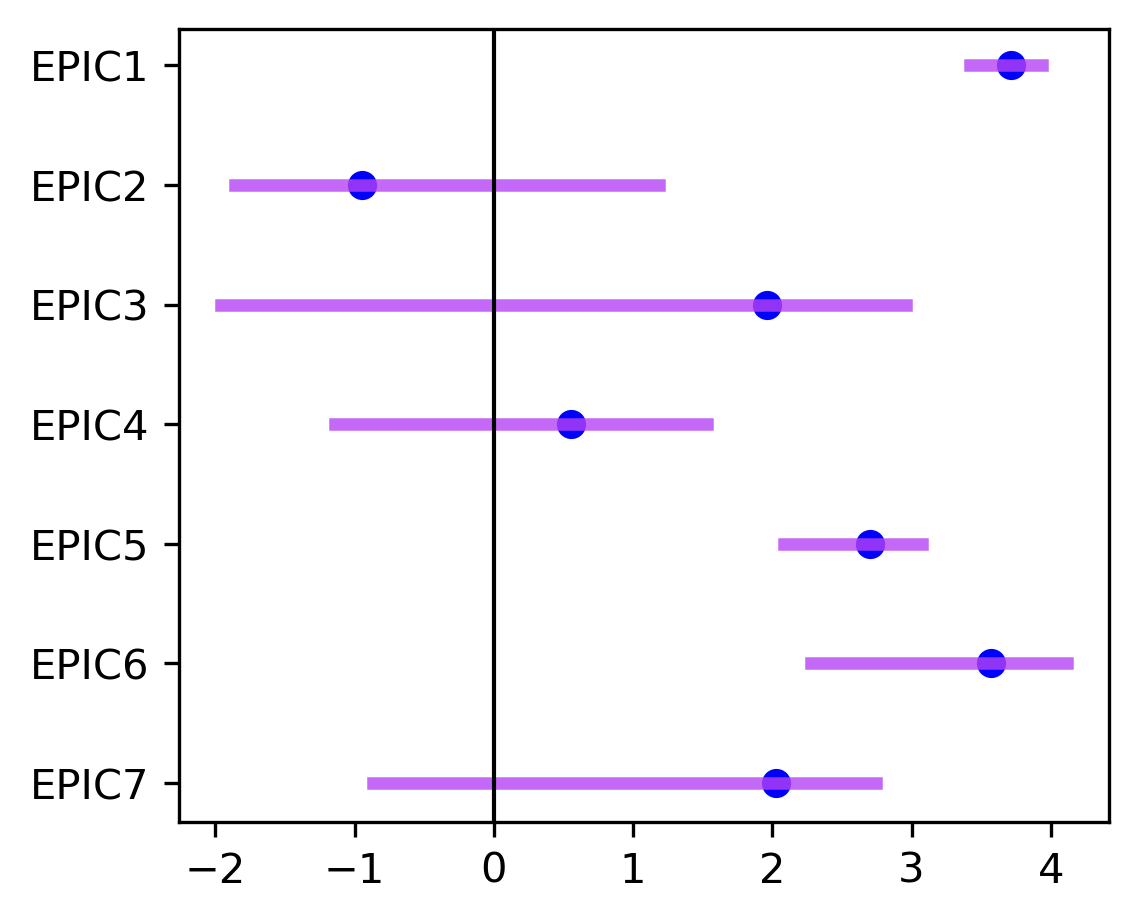}                  
         \caption{Bedrooms}
    \end{subfigure}  
    \begin{subfigure}[b]{0.31\textwidth}
         \centering
         \includegraphics[width=\textwidth]{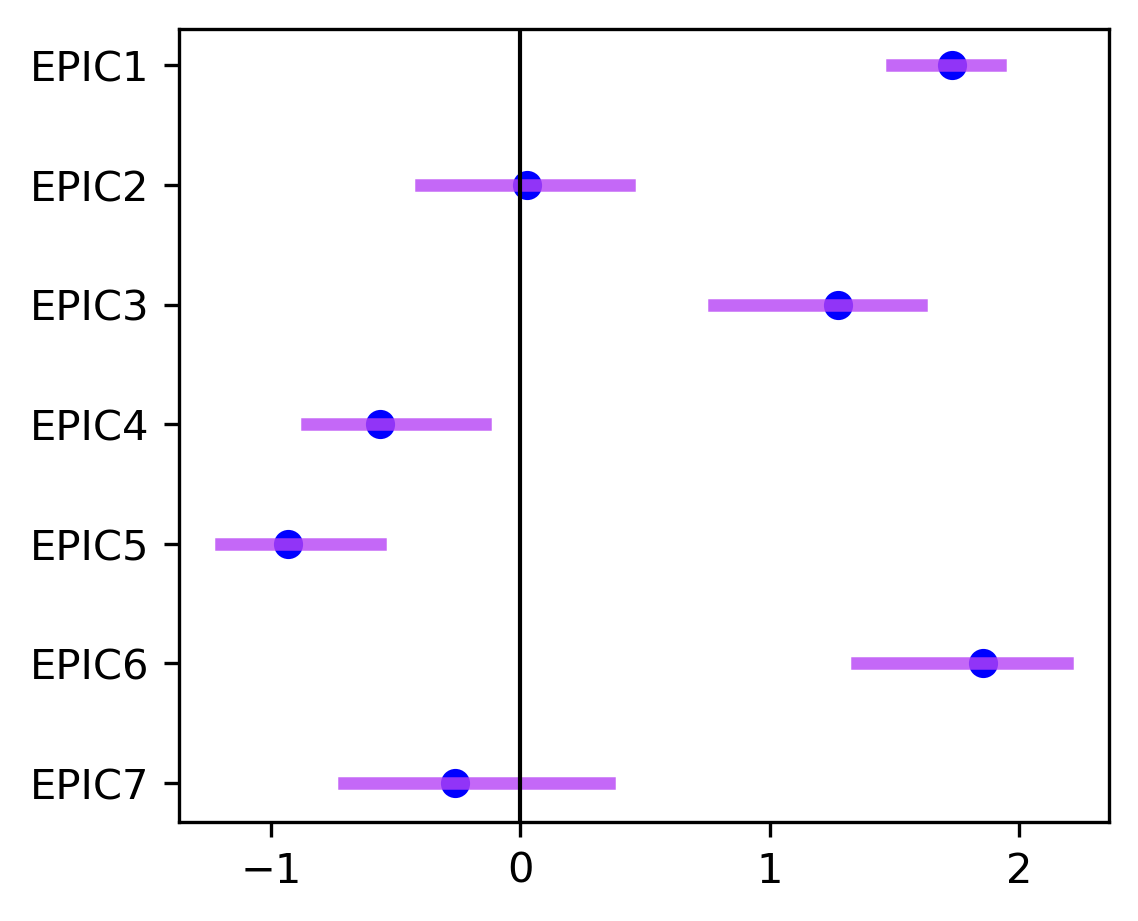}                  
         \caption{Population}
    \end{subfigure}
    \begin{subfigure}[b]{0.31\textwidth}
         \centering
         \includegraphics[width=\textwidth]{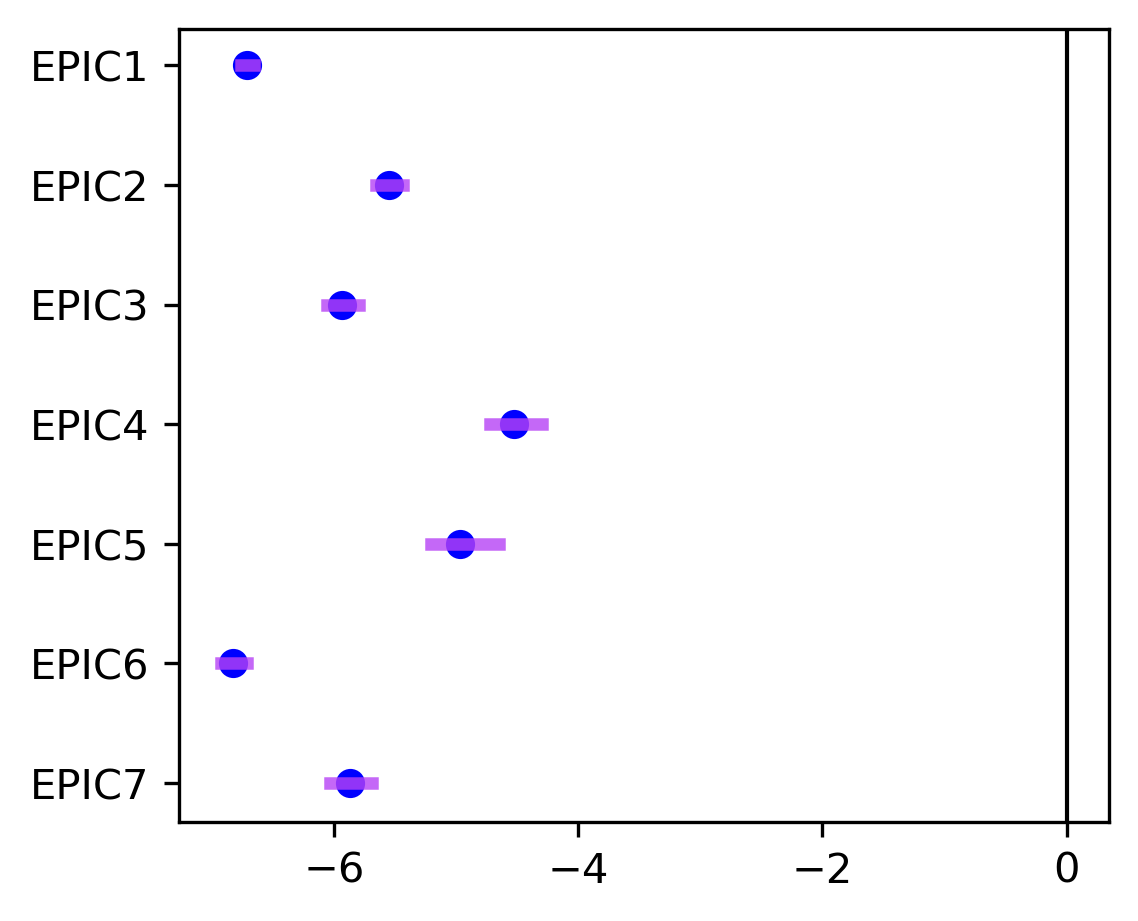}                  
         \caption{Occupancy}
    \end{subfigure}      
	\caption{Log-valued {\revisionii ``naive''} confidence intervals of each variable in the $7$ largest EPICs.}
	\label{fig:cal_ci}
\end{figure}

\begin{figure}[h]
	\centering
    \begin{subfigure}[b]{0.31\textwidth}
         \centering
         \includegraphics[width=0.95\textwidth]{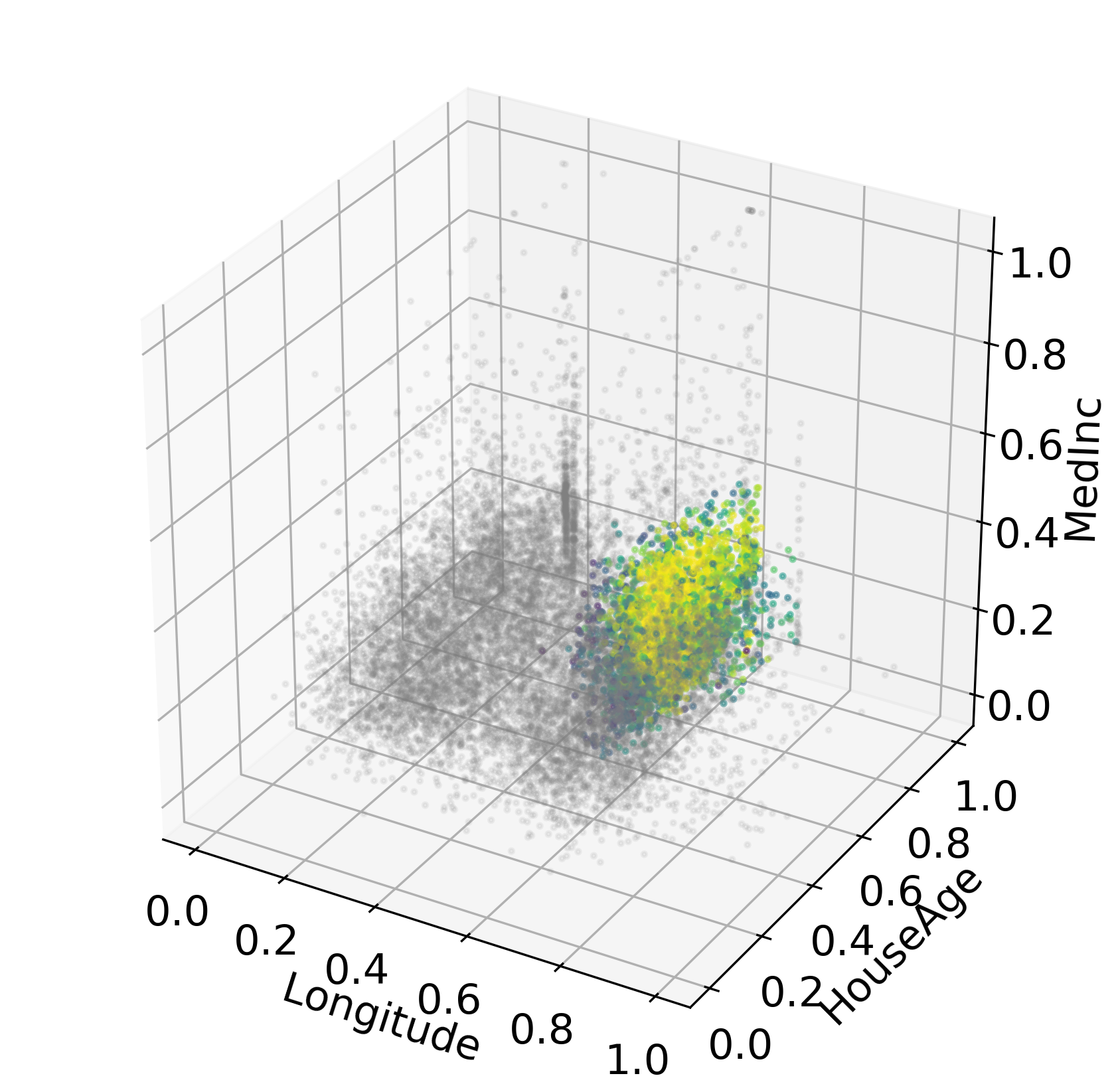}
         \caption{EPIC 1 (0.84)}
    \end{subfigure}
	\begin{subfigure}[b]{0.31\textwidth}
         \centering
         \includegraphics[width=0.95\textwidth]{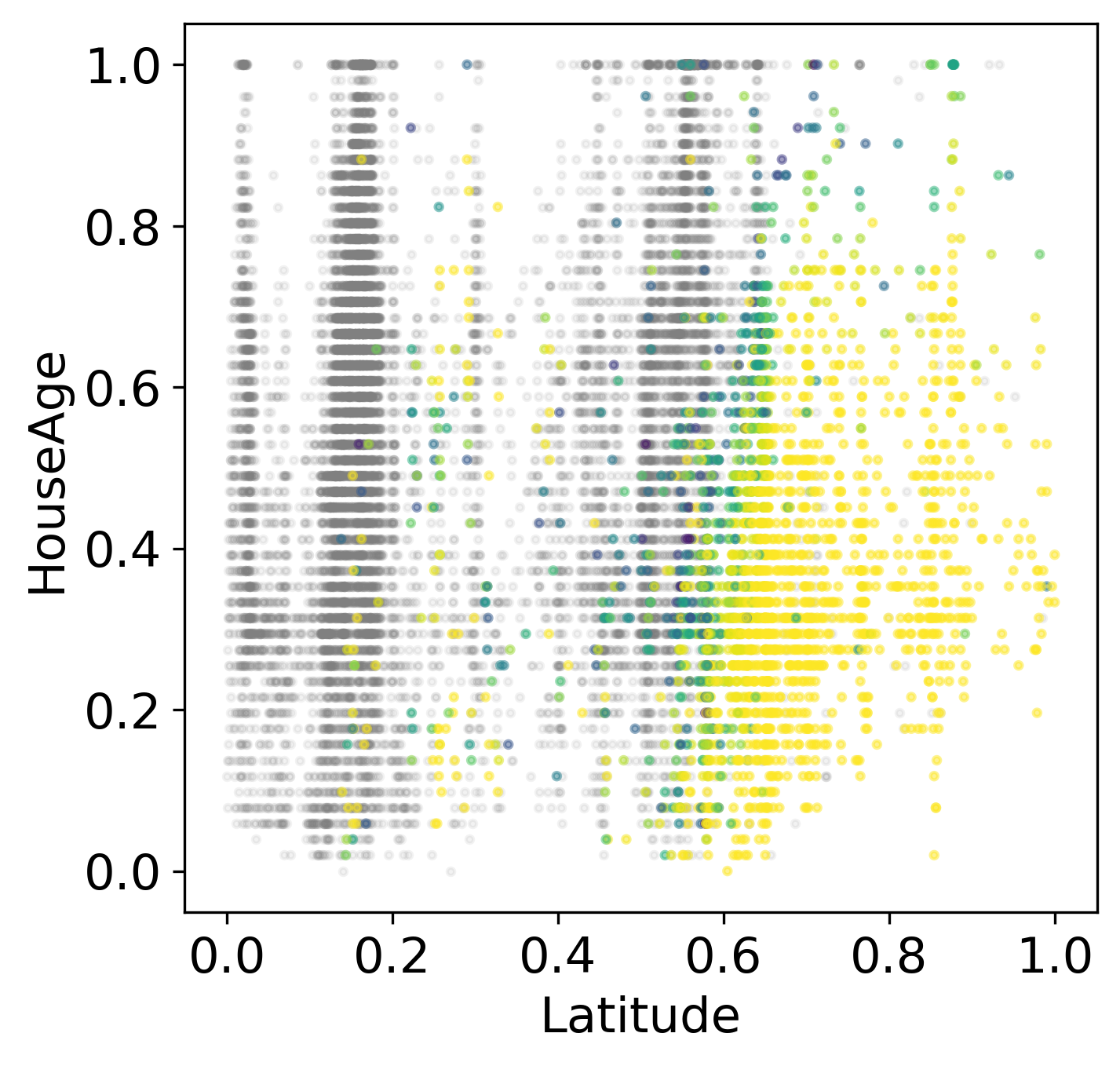}                  
         \caption{EPIC 2 (0.82)}
    \end{subfigure}
	\begin{subfigure}[b]{0.31\textwidth}
         \centering
         \includegraphics[width=0.95\textwidth]{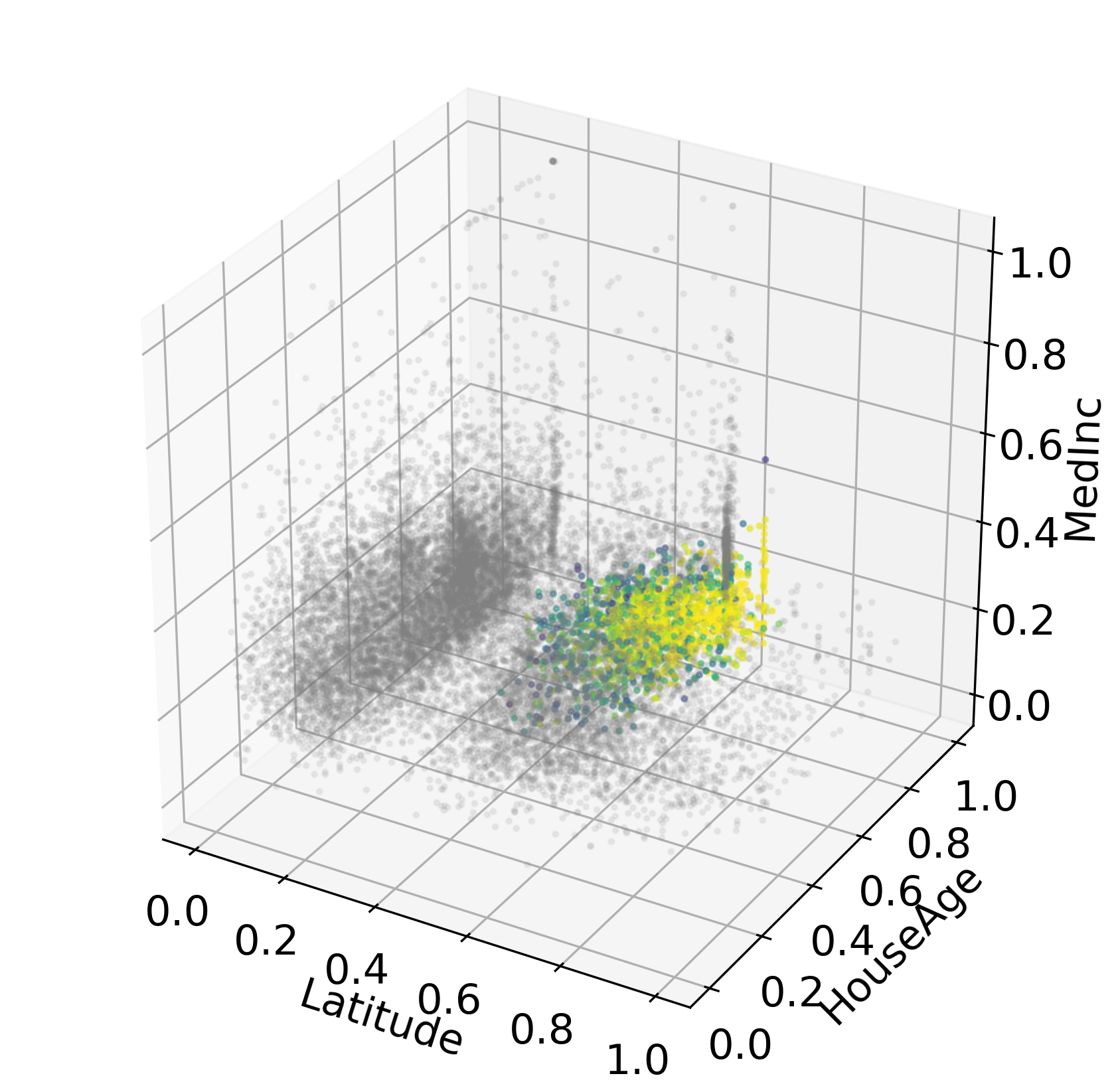}                  
         \caption{EPIC 3 (0.80)}
    \end{subfigure}\\
	\begin{subfigure}[b]{0.31\textwidth}
         \centering
         \includegraphics[width=0.95\textwidth]{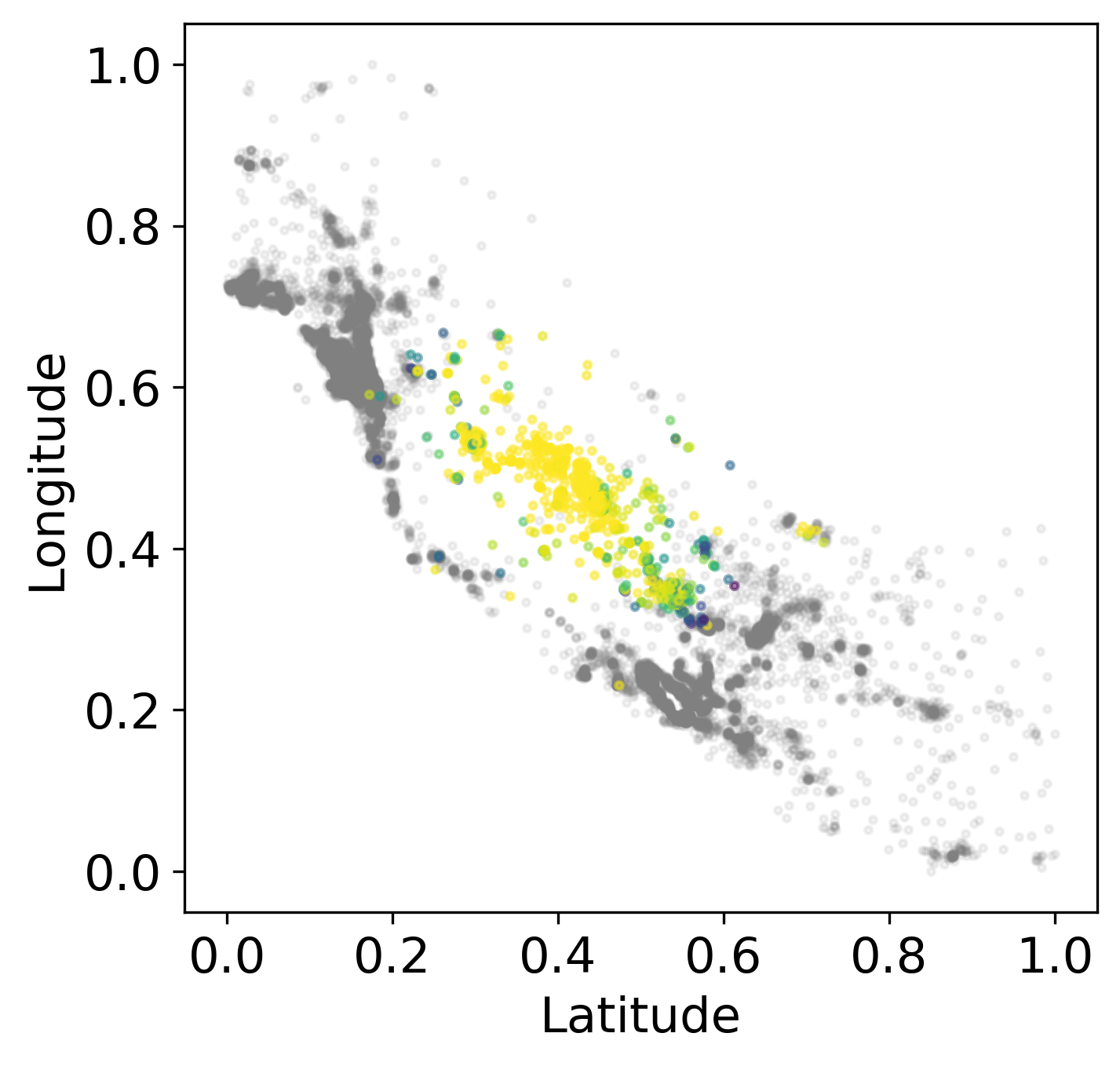}                  
         \caption{EPIC 4 (0.88)}
    \end{subfigure}  
    \begin{subfigure}[b]{0.31\textwidth}
         \centering
         \includegraphics[width=0.95\textwidth]{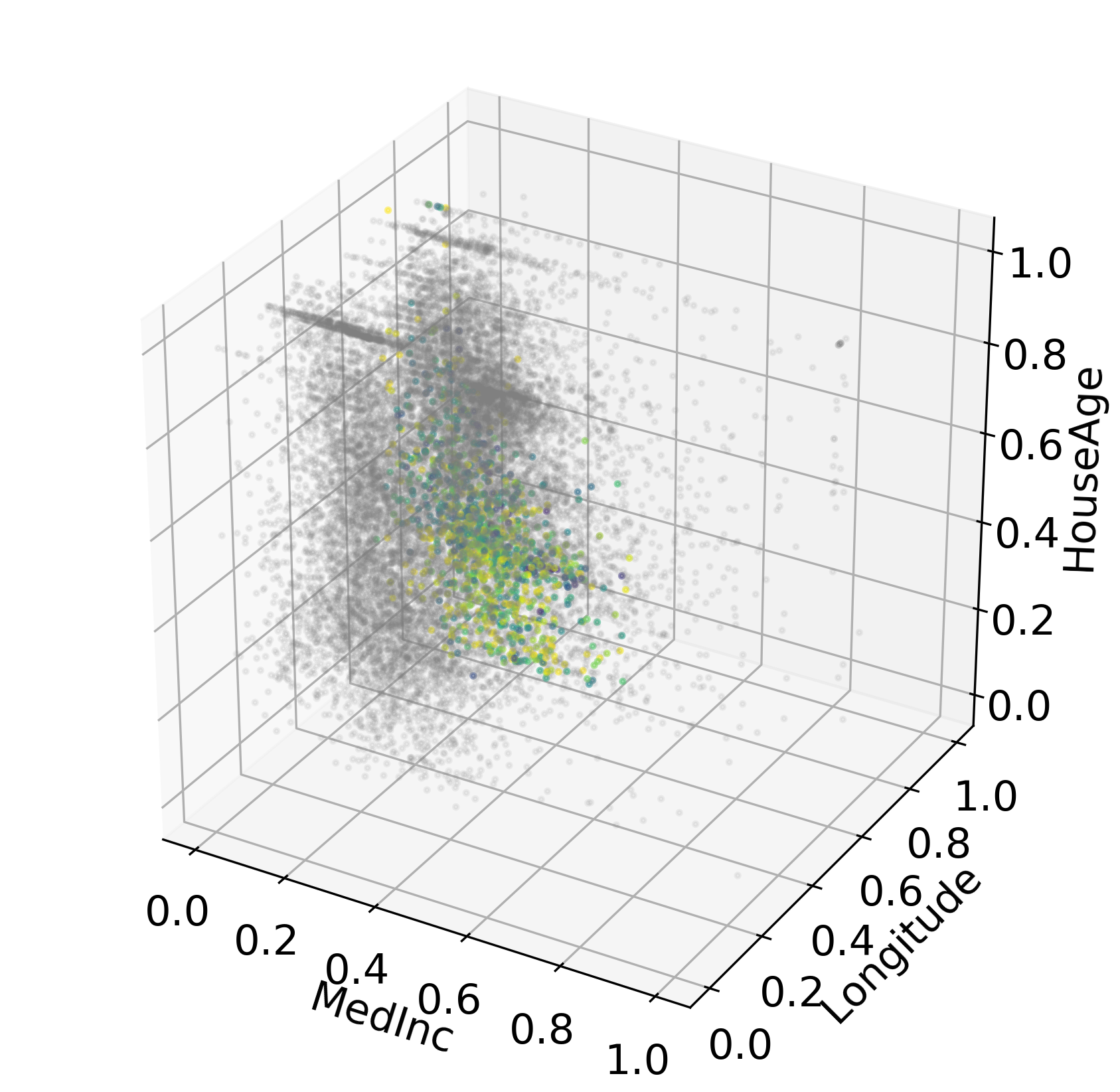}                  
         \caption{EPIC 5 (0.58)}
    \end{subfigure}
    \begin{subfigure}[b]{0.31\textwidth}
         \centering
         \includegraphics[width=0.95\textwidth]{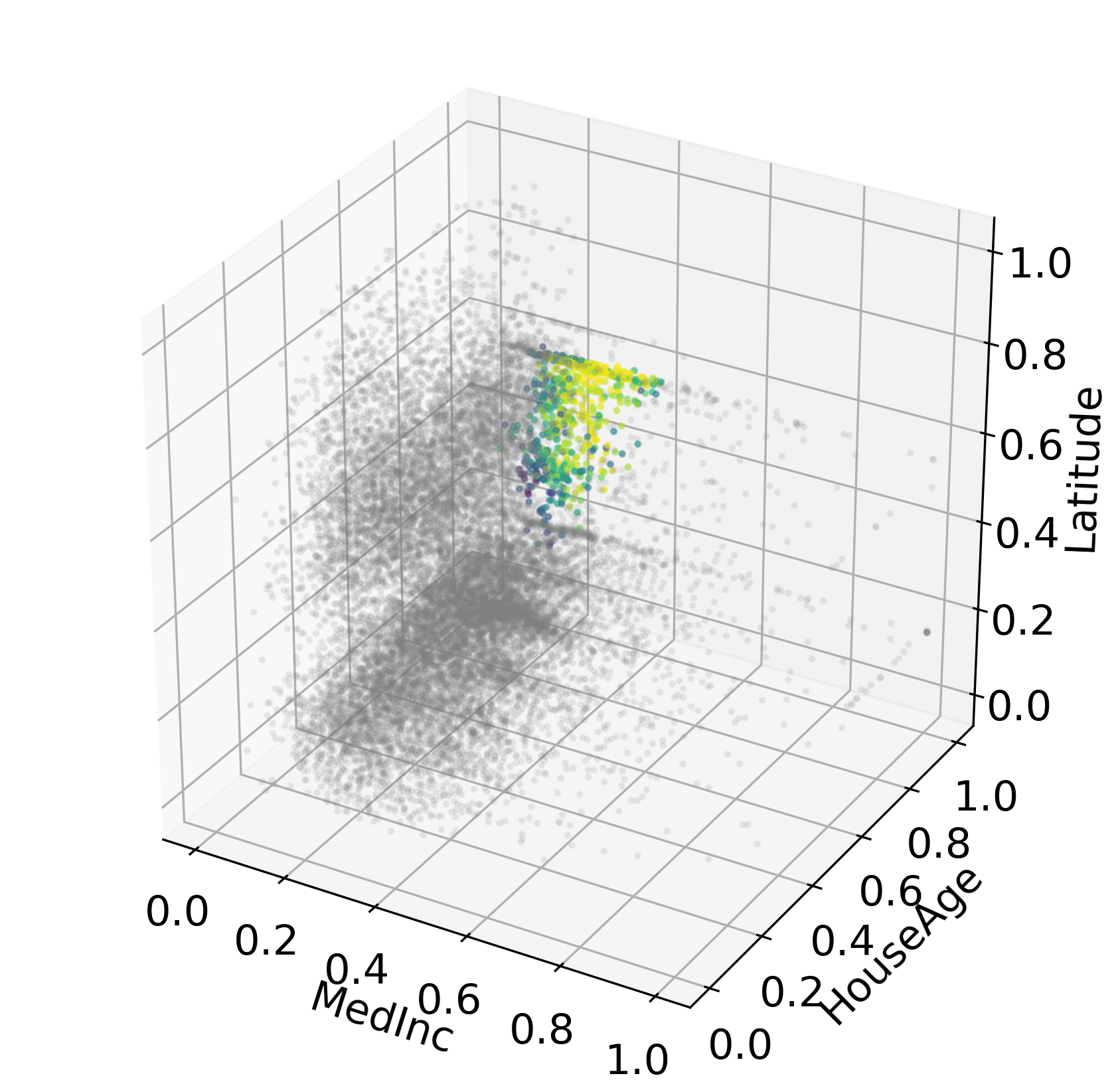}                  
         \caption{EPIC 6 (0.81)}
    \end{subfigure}      
	\caption{Explainable dimensions for the first $6$ biggest EPICs. The values in the parentheses are explainable rates of the explainable dimensions for each EPIC. }
	\label{fig:cal_dim}
\end{figure}

MLM-EPIC provides more useful interpretation than the commonly-used autoregressive models for California housing price data. 
Autoregressive models are primarily used to describe a time-varying or space-varying process, and the interpretation of the model is based on the model's representation of time or spatial variables. MLM-EPIC can be utilized to describe both time and space-varying effects as it divides samples into EPICs in different time and space. Figure \ref{fig:cal_map} shows the change of regression coefficients for each variable in the longitude and latitude space. The value of the coefficient for the variable in consideration is indicated by color. According to the fitted MLM-EPIC, the impact of the median income on the house values is relatively consistent across the California map. On the other hand, house age affects the house values differently depending on which EPIC the sample points belong to. Near the coastal area, old houses tend to be appreciated whereas in the inland area, house age has a negative impact on the house value. This observed pattern may be explained by the fact that houses in the highly populated area were built earlier and are still in high demand. Another interesting pattern in Figure \ref{fig:cal_map} is that the increase of the number of non-bedroom rooms has a negative impact on the house values in city center areas. The log-valued confidence intervals and estimates of the $6$ covariates are shown in Figure \ref{fig:cal_ci} for the $7$ biggest EPICs. These $7$ EPICs cover about $70\%$ of the training samples.

For this dataset, we cannot get concise explainable conditions for the EPICs. For example, an explainable condition for EPIC $1$ is `Median income$\,\leq86.6k$ and House age$\,\notin (3.5,27.5]$ and Bedrooms$\,\notin (33.1,33.6]$ and Population$\,\notin (32.1k, 35.7k]$ and Average occupancy$\,\notin(1239.6,1241.8]$ and Latitude$\,\notin(40.1,40.9]$ and Longitude$\,\notin(-121.2,-119.5]$'. 
Instead, we use LDS to find explainable dimensions for the EPICs. 
The explainable dimensions for the $6$ biggest EPICs are shown in {Figure \ref{fig:cal_dim}.} For example, based on $3$ features: Median income, House age, Longitude, EPIC $1$ is distinguishable from the other EPICs with accuracy $0.84$ as measured by F-1 score. EPIC $1$ mostly contains points with income lower than the median, house age older than the median, and located in the middle east part of California. In EPIC $1$, older houses tend to increase the house values slightly. Also, having more bedrooms is appreciated in the house value, while the increase of rooms other than bedrooms tends to decrease the value. EPIC $2$ contains mostly points with newer houses in the middle northern part of California (with the explainable rate equal $0.82$). For points in EPIC $2$, in contrast to EPIC $1$, the house age has a negative impact on the house value. Also, the increase in the number of rooms other than bedrooms positively affects the house value, whereas the impact of the number of bedrooms is {\revisionii less clear.}

\subsection{Parkinson's Disease Detection} 
\label{sec:pd}

Many works have studied Parkinson's disease (PD) to detect PD from vocal impairments in sustained vowel phonatations of PD patients. Due to the limited understanding of the mechanism to detect PD from the complex characteristics of patients' recorded voices, a dataset often involves a huge number of covariates that are extracted from various speech signal processing algorithms. In this subsection, we demonstrate that MLM-cell and MLM-EPIC work well with this high-dimensional dataset. At reduced complexity, the two methods achieved slightly better accuracy than the more complex models.  

\begin{table}[htp]
\centering
\begin{tabular}{c|p{12cm}|c}
\Xhline{2\arrayrulewidth}
	\hline\hline
	EPIC & Descriptions & Size\\
	\hline
	\shortstack{1 (442)}& \footnotesize std{\_}delta{\_}delta{\_}log{\_}energy$\,>0.057$, tqwt{\_}energy{\_}dec{\_}18$\,\leq 0.562$, & 381\\
	&\footnotesize tqwt{\_}TKEO{\_}std{\_}dec{\_}12$\,\leq 0.102$, tqwt{\_}medianValue{\_}dec{\_}31$\,\leq 0.454$ & \\
	\hline
	\shortstack{2 (133)}& \footnotesize mean{\_}MFCC{\_}2nd{\_}coef$\,\leq0.397$, tqwt{\_}entropy{\_}log{\_}dec{\_}27$\,\leq0.696$, & 75\\
		&\footnotesize tqwt{\_}entropy{\_}log{\_}dec{\_}35$\,>0.368$, tqwt{\_}TKEO{\_}std{\_}dec{\_}17$\,\leq0.216$ & \\
	\hline
	\shortstack{3 (29)} & \footnotesize tqwt{\_}entropy{\_}shannon{\_}dec{\_}8$\,\leq 0.007$, tqwt{\_}stdValue{\_}dec{\_}19$\,>0.373$ & 9 \\
	\hline\hline	
	\Xhline{2\arrayrulewidth}
\end{tabular}
	\caption{Explainable conditions for the $3$ EPICs for Parkinson's disease data. Numbers in the bracket indicates the size of EPIC in training data.}
	\label{tab:pd_desc}
\end{table}

For this dataset, RF is fitted with maximum depth $10$. MLP is constructed with $3$ layers with $30$ hidden units per layer. MLP is trained for $5$ epochs. Other methods follow the default parameter setting from \textbf{sci-kit learn} python package. MLM-cell is constructed based on MLP with layer-$l$ cells equal to $5$, which is chosen by CV, for $l=1,2,3$. $52$ cells are formed. We merge the cells into $3$ EPICs to build MLM-EPIC. For the local linear models of MLM-EPIC, we applied LASSO penalty with its weight parameter $\alpha =0.2$.

As shown by Table \ref{tab:real},
MLM-cell achieves the highest training and testing accuracy. In addition, although MLM-EPIC is only a mixture of $3$ local linear models, it has better testing accuracy (lower testing AUC) than MLP. Specifically, MLM-EPIC forms $3$ EPICs with sizes $442$, $133$, and $29$ for the training data. The local linear models respectively have $700$, $14$, $0$  non-zero regression coefficients. The last local linear model is a constant function which classifies all the points in EPIC 3 as class 0. In this case, the interpretation of EPIC 3 provides an explanation of a local region for class 0.

We compute the explainable dimensions for each EPIC. We set $\xi=0.8$. EPIC 1 is distinguishable with only one variable. In fact, $694$ single variables achieve an explainable rate higher than $0.8$. Among them, the variable 'tqwt{\_}TKEO{\_}std{\_}dec{\_}12' achieves the highest explainable rate of $0.87$. Whereas, EPIC 2 needs 7 variables to distinguish the EPIC with explainable rate higher than $0.8$, and EPIC 3 needs 13 variables. 

Explainable conditions provide a different perspective for interpreting the EPICs. With $\psi=0.9$ and $\eta=50$ for EPIC 1 and 2, and $\eta=5$ for EPIC 3, we find the explainable conditions listed in Table \ref{tab:pd_desc}. In the explainable conditions, 'tqwt{\_}TKEO{\_}std{\_}dec{\_}12' appears again as an important feature to distinguish EPIC 1. EPIC 2 and 3 are relatively hard to be distinguished with explainable conditions as the explainable conditions only capture 75 out of 133 points for EPIC 2, and 9 out of 29 points for EPIC 3.

\section{Conclusions}
\label{sec:discuss}
In this paper, we develop MLM under co-supervision of a trained DNN. Our goal is to estimate interpretable models without compromising performance. The experiments show that MLM achieves higher prediction accuracy than other explainable models. However, for some data sets, the gap between the performance of MLM and DNN is not negligible. Interpretation is intrinsically subjective. For different datasets, different approaches can be more suitable. We have developed a visualization method and a decision rule-based method to help understand the prediction function. In any case, these methods rely heavily on the local linear models constructed in MLM. To explore the aspect of interpretation at a greater depth, we plan to examine applications in more specific contexts. {\revision The idea of co-supervision is interesting in its own right, which can be taken as a principle for developing interpretable models. The technical components in MLM may be replaced by other approaches, for example, the regression model used within each EPIC and the method to generate EPICs. These variations can be explored in future works.} {\revisionii Also, the statistical inference of MLM coefficients is an interesting future direction. For more rigorous interpretation of MLM, we should validate the model structure using statistical inference methods.}

\section*{Acknowledgments}
Li and Lin's research is supported by the National Science Foundation under grant DMS-2013905.

\bibliographystyle{apalike}
\bibliography{biblio}

\end{document}


\def\spacingset#1{\renewcommand{\baselinestretch}%
{#1}\small\normalsize} \spacingset{1}

\newcommand{\revision}{\color{black}}
\newcommand{\revisionii}{\color{red}}
\renewcommand{\thesection}{\Alph{section}}

\newpage
\spacingset{1.5} 


\title{\textbf{Supplementary Material}}
\maketitle
\section{Additional Interpretation for SKCM Clinical Data}

Table \ref{tab:skcm_desc2} shows explainable conditions of the second and third largest EPICs for SKCM data. The explainable conditions for predicting OS status involve not only the case entity that represents clinical information about a case, such as demographic, prognosis, and diagnosis variables but also sample entity that describes procedures that samples or specimen materials are taken. Interpreting sample entity as a covariate for predictive models is not intuitive. Nonetheless, looking into those variables can aid in the development of hypotheses investigating the presence of any confounding variables. For EPIC 2, the covariates Days to collection, AJCC staging edition, and Fraction genome altered are included in the most of explainable conditions. For EPIC 3, Age, Form completion date, AJCC staging edition, and Tumor site are important variables to construct the most of explainable conditions.  

\begin{table}[ht]
\centering
\begin{tabular}{c|p{12cm}|c}
\Xhline{2\arrayrulewidth}
	\hline\hline
	EPIC & Descriptions & Size\\
	\hline
	2 & \footnotesize Days to collection$\,> 10158.5$, Overall survival months$\,\leq 195.4$, Submitted tumor DX days$\,> 10292.5$, AJCC staging edition$\,= 5$, Tumor site$\,=0$ & 12 \\
	\cline{2-3}
	(54)& \footnotesize Days to collection$\,\leq 8903$ or Days to collection$\,> 10158.5$, Fraction genome altered$\,>0.8$, AJCC staging edition$\,= 5$, Tumor site$\,\neq 0$ & 5 \\
	\cline{2-3}
	& \footnotesize Days to collection$\,> 10158.5$, Fraction genome altered$\,>0.9$, AJCC staging edition$\,= 2$ & 5 \\
	\cline{2-3}
	& \footnotesize Days to collection$\,> 10158.5$, Fraction genome altered$\,>0.9$, Overall survival months$\,>269.8$, AJCC staging edition$\,\neq 2\,$or$\,5$ & 5 \\
	\hline
	3 & Age$\,<45.5$, Form completion date$\,>45$, AJCC staging edition$\,= 5$, Tumor site$\,\neq 0\,$or$\,5$. & 11\\
	\cline{2-3}
	(46) & \footnotesize  Age$\,<45.5$, Days to collection$\,> 11850.5$, Form completion date$\,\leq 45$, Radiation treatment adjuvant$\,=0$, Sample initial weight$\,\leq 1595.0$, Sex$\,=0$, AJCC nodes pathologic PN$\,\neq 1$, AJCC staging edition$\,=5$, Tumor site$\,=5$ & 9 \\
	\cline{2-3}
	& \footnotesize Tumor site$\,=5$ & 5 \\
	\Xhline{2\arrayrulewidth}
\end{tabular}
	\caption{Explainable conditions for the second and third largest EPICs for SKCM data. Numbers within the bracket indicate the size of the EPIC in training data.}
	\label{tab:skcm_desc2}
\end{table}

\newpage
\section{Additional Interpretation for Bike Sharing Data}

To interpret MLM-EPIC fitted for bike sharing data, we use PR method in Section 3.2.1. Table \ref{tab:bike_desc} shows the explainable conditions for the first $5$ largest EPICs. The threshold $\psi$ is set to $0.8$ to reduce the number of explainable conditions. The minimum size threshold $\eta$ is set $50$. 
EPIC $1$ can be interpreted as ``Early afternoon in work days between spring to fall". Figure 4 in the main text 
shows that in EPIC $1$, humidity and wind speed affect the decrease of bike rentals more severely than in other EPICs. Whereas, EPIC $2$ is mostly characterized as ``Around noon to early afternoon in work days for winter season". For EPIC $2$, humidity does not affect the bike rental counts but high temperature has a stronger positive effect on the bike rentals than in EPIC $1$. Also, for EPIC $2$,  light rain tends to decrease the bike rentals more sharply than in other EPICs. EPIC $3$ is ``Late evening in work days between April to summer". For the EPIC $3$, temperature and feeling temperature are the most important factors for predicting the bike rentals. EPIC $4$ is ``Morning around 10 to 11 am in work days for spring and summer". For this EPIC, light rain does not decrease the bike rentals compared to clear or cloudy weather. Lastly, EPIC $5$ is more complex. It is roughly characterized by ``Early afternoon that is not working day after April or in summer". For this EPIC, humidity, wind speed, and light rain 
decrease the bike rentals more than other weather conditions.

\begin{table}[h]
\centering
\begin{tabular}{c|p{12cm}|c}
\Xhline{2\arrayrulewidth}
	\hline\hline
	EPIC & Descriptions & Size\\
	\hline
	\shortstack{1 (883)}& \footnotesize Month$\,>3.5$, $11.5<\,$Hour$\,\leq14.5$, Working day$\,=1$ & 727\\
	\hline
	\shortstack{2 (488)}& \footnotesize Month$\,\leq 3.5$, $9.5 <\,$Hour$\,\leq 14.5$, Working day$\,=1$, Season$\,\neq1$ & 377\\
	\cline{2-3}
	& \footnotesize Year$\,\leq 2011$, Month$\,\leq 2.5$, $10.5 <\,$Hour$\,\leq 14.5$, Working day$\,=1$, Season$\,\neq1$ & 51\\
	\hline
	\shortstack{3 (456)} & \footnotesize Month$\,> 3.5$, $20.5 <\,$Hour$\,\leq 21.5$, Working day$\,=1$, Season$\,\neq 3$ & 166\\
	\hline
	\shortstack{4 (370)} & \footnotesize $9.5 <\,$Hour$\,\leq 11.5$, Working day$\,=1$, Season$\,= 1$ & 182\\
	\cline{2-3}
	& \footnotesize $9.5 <\,$Hour$\,\leq 10.5$, Working day$\,=1$, Season$\,= 2$, Weather$\,\neq 2$ & 88\\
	\hline
	\shortstack{5 (307)} & \footnotesize Year$\,\leq 2011$, Month$\,>3.5$, $13.5<\,$Hour$\,\leq15.5$, Holiday$\,\neq 1$, Working day$\,\neq 1$, Temperature$\,\leq0.35$, Season$\,\neq 3$ & 66\\
	\cline{2-3}
	& \footnotesize $11.5<\,$Hour$\,\leq14.5$, Weekday$\,\leq 0.5$,  Working day$\,\neq 1$, Season$\,= 1$ & 59\\
	\hline\hline	
	\Xhline{2\arrayrulewidth}
\end{tabular}
	\caption{Explainable conditions for the top $5$ largest EPICs for bike sharing data. Numbers in the bracket are the sizes of EPICs in the training data.}
	\label{tab:bike_desc}
\end{table}

{\revision
\section{Stability Analysis for EPICs Based on California Housing Data}

{\revision Variation in explanation is a universal issue for any explanation of a complex model. In this section, we examine the stability of MLM and propose an approach to measure the uncertainty of explanation of EPICs. We use {\it OTclust}\citep{li2019optimal,zhang2020cps} which is a clustering validation tool with uncertainty analysis based on bootstrap resampling. OTclust is available as an R package from CRAN.  

To investigate the stability of EPICs, $100$ MLM models are trained on $100$ bootstrapped samples of California housing data. OTclust computes a mean partition (that is, an ``average'' clustering result) for the $100$ different EPIC partitions using optimal transport alignment. 
The largest $4$ EPICs in the mean partition have average Jaccard similarity\citep{jaccard1912distribution} of $0.67$ to the corresponding EPICs obtained based on the bootstrap samples. The score decreases to $0.53$ for the smaller EPICs except for the EPICs containing fewer than $10$ points. To summarize, on average, $53\%$ to $67\%$ of points remain in the same EPICs under bootstrap resampling. Given that $30$ clusters are formed for any bootstrap sample, this degree of agreement between the EPICs is good.


Although the majority of sample points remain in the same EPICs, small changes in an EPIC can affect considerably MLM interpretation because the local regression is sensitive to outliers in data. In California housing data, the occupancy variable has a relatively stable coefficient under bootstrap resampling. The interquantile range~(IQR) of the occupancy coefficient, according to the bootstrap samples, does not overlap for EPIC 3 and EPIC 2, the range for the former being lower. EPIC 3 contains mainly cases located in southern coastal area with lower median income, whereas EPIC 2 contains cases in the inland area of southern California. In contrast to the variable of occupancy, the median income coefficients have large bootstrap IQRs and are less stable. 

We propose to take the mean partition computed from bootstrap samples as an average result of EPIC partition. Statistics such as the bootstrap IQR can be used as an uncertainty assessment for the EPICs. This uncertainty assessment can be provided as a part of the explanation. 

}

}

\bibliographystyle{apalike}
\bibliography{biblio}